%% file: main.tex
\ifdefined\PaperTMLR
  \documentclass[10pt]{article}
  \usepackage[accepted]{tmlr}
\else
  \documentclass{article}
  \usepackage{arxiv}
\fi

\input{preamble}
\input{style}

\begin{document}

\maketitle

\input{manuscript}

\end{document}

%% file: preamble.tex
\ifdefined\PaperTMLR
  \usepackage{hyperref}
  \usepackage{url}
  \usepackage{microtype}
  \usepackage{graphicx}
  \usepackage{subcaption}
  \usepackage{booktabs}

  \usepackage{amsmath}
  \usepackage{amssymb}
  \usepackage{mathtools}
  \usepackage{amsthm}
  \usepackage{enumitem}
  \usepackage{thm-restate}
  \usepackage{bm}
  \usepackage{bbm}
  \usepackage[ruled,vlined]{algorithm2e}
  \usepackage[title]{appendix}
  \usepackage[capitalize,noabbrev]{cleveref}

  \input{macros}

  \theoremstyle{plain}
  \newtheorem{theorem}{Theorem}[section]
  
  \newtheorem{lemma}[theorem]{Lemma}
  \newtheorem{corollary}[theorem]{Corollary}
  \theoremstyle{definition}
  \newtheorem{definition}[theorem]{Definition}
  \newtheorem{assumption}[theorem]{Assumption}
  \theoremstyle{remark}
  \newtheorem{remark}[theorem]{Remark}

  \newcommand{\PaperBibliographyStyle}{tmlr}
  \newcommand{\PaperAppendixStart}{\appendix}
\else
  \usepackage{amsmath,amsfonts,bm,bbm,amsthm,amssymb}
  \usepackage{mathtools}
  \usepackage{times}
  \usepackage[ruled,vlined]{algorithm2e}

  \newtheorem{theorem}{Theorem}[section]
  \newtheorem{corollary}{Corollary}[theorem]
  \newtheorem{lemma}{Lemma}[section]
  
  \newtheorem{definition}{Definition}

\input{macros}
  \usepackage[utf8]{inputenc}
  \usepackage[T1]{fontenc}
  \usepackage{hyperref}
  \usepackage{url}
  \usepackage{booktabs}
  \usepackage{amsfonts}
  \usepackage{nicefrac}
  \usepackage{microtype}
  \usepackage{paralist,multirow,graphicx,amsmath,amsfonts,amssymb,epstopdf,color,epsfig,ragged2e,array,standalone,booktabs,subfigure}
  \usepackage{stfloats}
  \graphicspath{{./images/}}
  \usepackage[numbers]{natbib}
  \usepackage{enumitem}
  \usepackage{thm-restate}
  \usepackage[title]{appendix}
  \usepackage{todonotes}

  \newcommand{\PaperBibliographyStyle}{unsrtnat}
  \newcommand{\PaperAppendixStart}{\appendices}
\fi

%% file: macros.tex
\newcommand{\R}{\ensuremath{\mathbb{R}}}
\newcommand{\N}{\ensuremath{\mathbb{N}}}

\renewcommand{\N}{\ensuremath{\mathbb{N}}}

\newcommand{\E}{\ensuremath{\mathbb{E}}}
\renewcommand{\P}{\ensuremath{\mathbb{P}}}

\def\1{\mathbbm{1}}

\DeclareMathOperator*{\argmin}{arg\,min}

\def\vtheta{{\bm{\theta}}}
\def\vomega{{\bm{\omega}}}
\def\vbeta{{\bm{\beta}}}
\def\vsigma{{\bm{\sigma}}}
\def\vepsilon{{\bm{\epsilon}}}

\def\va{{\bm{a}}}
\def\vb{{\bm{b}}}

\def\vd{{\bm{d}}}
\def\ve{{\bm{e}}}

\def\vg{{\bm{g}}}

\def\vu{{\bm{u}}}

\def\vw{{\bm{w}}}
\def\vx{{\bm{x}}}

\def\vz{{\bm{z}}}
\def\vtheta{\mbox{\boldmath$\theta$}}

\def\mB{{\bm{B}}}

\def\mP{{\bm{P}}}
\def\mQ{{\bm{Q}}}
\def\mR{{\bm{R}}}

\def\mT{{\bm{T}}}
\def\mU{{\bm{U}}}
\def\mV{{\bm{V}}}
\def\mW{{\bm{W}}}
\def\mX{{\bm{X}}}

\def\mTheta{{\bm{\Theta}}}

\newcommand{\cA}{\ensuremath{\mathcal{A}}}

\newcommand{\cC}{\ensuremath{\mathcal{C}}}
\newcommand{\cD}{\ensuremath{\mathcal{D}}}
\newcommand{\cE}{\ensuremath{\mathcal{E}}}
\newcommand{\cF}{\ensuremath{\mathcal{F}}}
\newcommand{\cG}{\ensuremath{\mathcal{G}}}
\newcommand{\cH}{\ensuremath{\mathcal{H}}}
\newcommand{\cI}{\ensuremath{\mathcal{I}}}
\newcommand{\cJ}{\ensuremath{\mathcal{J}}}

\newcommand{\cN}{\ensuremath{\mathcal{N}}}

\renewcommand{\cN}{\ensuremath{\mathcal{N}}}
\newcommand{\cO}{\ensuremath{\mathcal{O}}}
\newcommand{\cP}{\ensuremath{\mathcal{P}}}
\newcommand{\cQ}{\ensuremath{\mathcal{Q}}}

\newcommand{\cS}{\ensuremath{\mathcal{S}}}

\newcommand{\cV}{\ensuremath{\mathcal{V}}}


%% file: style.tex
\ifdefined\PaperTMLR
  \title{Implicit Bias and Invariance: How Hopfield Networks \\Efficiently Learn Graph Orbits}

  \newcommand{\PaperAcknowledgments}{\subsubsection*{Acknowledgments}}

  \author{\name Michael Murray \email mjm253@bath.ac.uk \\
        \addr Department of Mathematical Sciences\\
        University of Bath
        \AND
        \name Tenzin Chan \\
        \addr DP Architects
        \AND
        \name Kedar Karhadkar \\
        \addr Department of Mathematics\\
        UCLA
        \AND
        \name Christopher J. Hillar \\
        \addr Algebraic
        }

\else
  \newcommand{\PaperAcknowledgments}{\section*{Acknowledgments}}
  \title{Implicit Bias and Invariance: How Hopfield \\ Networks Efficiently Learn Graph Orbits}
  \author{
    Michael Murray\textsuperscript{1} \quad
    Tenzin Chan \quad
    Kedar Karhadkar\textsuperscript{2} \quad
    Christopher J. Hillar\textsuperscript{3} \\
    \textsuperscript{1}Mathematical Sciences, University of Bath\qquad
    \textsuperscript{2}Department of Mathematics, UCLA \\
    \textsuperscript{3}Algebraic \\
    Correspondence: \texttt{mjm253@bath.ac.uk}
  }
\fi

%% file: manuscript.tex
\begin{abstract}
  Many learning problems are organized by group symmetries. While invariance is often imposed through architectures or group averaging, we ask when it can emerge from training on a finite random subset of an orbit. We study this question in classical Hopfield networks, where strict memorization can be expressed as a linear margin problem. Reparameterizing minimization of energy flow (MEF) as an exponential loss connects gradient descent to the corresponding minimum-norm hard-margin memorizer. Our main result shows that, for independent uniform samples from any finite permutation orbit, the exact sample hard-margin support vector machine (HSVM) concentrates exponentially around the invariant full-orbit HSVM. Consequently, an orbit-size-independent polynomial number of samples suffices both for approximate parameter invariance and for simultaneous memorization of every orbit element; directional convergence transfers this conclusion asymptotically to MEF gradient descent. For graph-isomorphism orbits, we characterize the invariant parameters as a three-dimensional subspace and show that every such orbit is memorizable. For cliques of fixed linear density, additional symmetry sharpens the uniform memorization bound to $O(v^4\log(1/\delta))$, exponentially smaller than the orbit size and thereby resolving Conjecture~1 of \citet{hillar2021hidden}. Together with experiments across several learning rules, these results give a finite-sample account of how optimization bias can recover symmetry from partial group-structured data.
  \ifdefined\PaperTMLR
    \footnote{\textbf{Statement on the use of LLMs.} Large language models, including OpenAI's GPT-5.6 Sol, were used extensively as research assistants for literature search, high-level ideation and planning, organizing related work, and drafting and refining exposition. All LLM-generated material was critically evaluated and revised by the authors. In particular, the authors checked every mathematical claim and proof in full, verified the cited sources, and reworked the arguments and writing where necessary. The authors take full responsibility for the correctness and presentation of the paper.}
  \fi
\end{abstract}

\section{Introduction}
Here, we analyze the emergence of invariance arising implicitly during training in Hopfield networks (HNs)~\citep{hopfield1982neural}, which are arguably the simplest example of an associative memory. Building on classical ideas~\citep{rosenblatt1958perceptron,willshaw1969non,amari1972learning,little1974existence,pastur1977exactly}, HNs are recurrent neural networks consisting of $n$ linear-threshold McCulloch--Pitts neurons~\citep{mcculloch1943logical} that can store binary patterns as distributed memories in the form of fixed-point attractors of recurrent dynamics. In the literature, HNs are usually associated with a particular Hebbian learning scheme called the outer-product rule, but for the purposes of this work we also consider other standard training methods. This setting is intentionally minimal so that we can focus on developing novel mathematical tools for understanding generalization in a classical architecture. As data symmetry is not made explicit in this model, any invariance must arise from the interplay of the group structure in the data and the implicit bias of the learning rule in question. More specifically, and inspired by~\citet{hillar2018robust,hillar2021hidden}, this paper studies whether standard learning rules and objectives, notably minimization of the energy flow (MEF)~\citep{mpf-hopf}, a tractable convex loss, can memorize unseen labelled realizations from the isomorphism class of a graph after training on a small, random subset.

The central question is whether observing a small random subset of a group orbit can force an unconstrained learner towards invariant parameters and thereby determine its behavior on the entire orbit. This is stronger than ordinary distributional generalization, which controls performance on a fresh random draw but need not guarantee correctness on every point in the support. It also differs from invariant architectures and full-orbit data augmentation, where invariance is imposed explicitly. Our main result gives such a finite-sample guarantee for minimum-norm memorization: the solution obtained from a random partial orbit approaches the invariant full-orbit solution and, with an orbit-size-independent number of samples, satisfies every constraint on the orbit simultaneously.

These results are of interest for two complementary reasons. From a learning-theoretic perspective, they provide a finite-sample route from symmetry of the data constraints to invariance in parameter space and then to uniform, support-wide generalization. From a practical perspective, they demonstrate in a concrete setting that an exponentially large augmentation orbit need not be enumerated, nor invariance imposed architecturally: indeed, for Hopfield networks a norm-biased learner trained on a small random subset can recover approximately invariant parameters and generalize to unseen transformations from a vanishingly small sample of the orbit. While Hopfield networks provide a minimal model in which this mechanism can be isolated exactly, the certificate argument we develop suggests a route for studying analogous behavior in other minimum-norm learners. We summarize our key findings as follows.

\begin{enumerate}
\item \textbf{Finite-sample emergence of invariance and uniform orbit memorization.} For uniform samples from any finite, strictly memorizable permutation orbit, we prove that the exact sample HSVM concentrates exponentially around the invariant full-orbit HSVM (Theorem~\ref{thm:exact-hsvm-concentration}). Consequently, polynomially many samples suffice both to approach the invariant parameter subspace and to memorize every orbit point simultaneously. We emphasize that these sample complexity bounds have no dependence on the orbit cardinality (Corollary~\ref{cor:uniform-orbit-memorization}), which may be exponentially or combinatorially large in comparison. For linear-density $k$-cliques with fixed density, additional symmetry yields the sharper sufficient bound $O(v^4\log(1/\delta))$ (Corollary~\ref{cor:clique-uniform-memorization}).

\item \textbf{Minimum norm as the optimization mechanism.} We reformulate strict Hopfield memorization as a linear margin problem and reparameterize MEF as an exponential loss. Gradient descent is therefore directionally biased towards the minimum-norm hard-margin memorizer (Corollary~\ref{cor:MEF-SVM}). Combined with the uniform margin supplied by our finite-sample result, this implies that with sufficiently many iterates MEF minimization also memorizes the full orbit.

\item \textbf{Graph orbits and invariant parameters.} For graphs encoded by their edge indicators, we characterize the parameters invariant under vertex relabelling as a three-dimensional subspace (Lemma~\ref{lem:inv-params-cond}). We give an explicit invariant construction showing that every graph-isomorphism orbit is strictly memorizable (Lemma~\ref{lemma:express-inv-param}), and a substantially smaller-norm construction for fixed-density cliques (Lemma~\ref{lem:gii-strict-mem-ex}). Experiments show that parameters learned from partial graph orbits approach this invariant structure as the sample size grows (Figs.~\ref{fig:weight_hists}, \ref{fig:weight_heatmaps2}, \ref{fig:HCP}, \ref{fig:other_families2}) and support our theoretical results.

\item \textbf{A general comparison bound without group structure.} Using standard Rademacher-complexity tools, we first establish a distributional baseline that applies to arbitrary sparse, strictly memorizable data. If $D\subset\{0,1\}^n$ has minimum-norm parameter $\vtheta^*$ and $\|\vx\|_0\leq m$ for all $\vx\in D$, then $N=\tilde{\Omega}\!\left(n\|\vtheta^*\|^2m\epsilon^{-2}\right)$ samples suffice for both HSVM and MEF to memorize a fresh draw with probability at least $1-\epsilon$ (Theorem~\ref{lem:gen-svm-bounds}). This result applies more generally than our orbit analysis, but controls only distributional error on a fresh sample. It provides a comparison point for Theorem~\ref{thm:exact-hsvm-concentration}: by exploiting group transitivity, we instead obtain convergence towards invariant parameters and simultaneous memorization of the entire orbit.
\end{enumerate}

The main technical obstacle is that the sample HSVM is a nonlinear function of its sampled constraints and therefore cannot be treated directly as an empirical average. We overcome this by constructing a group-invariant dual certificate for the full-orbit problem and showing that the deviation of the sample optimizer is controlled by the empirical average of its orbitwise certificate contributions.

\subsection{Related work}
\textbf{Capacity of Hopfield networks.}
The capacity of a Hopfield network depends on the learning rule and the structure of the data. For dense, uncorrelated random patterns under Hebbian learning, the statistical-mechanics analysis of \citet{amit1985storing} gives the classic linear law: reliable retrieval up to approximately $0.138$ patterns per neuron, with subsequent refinements via replica methods~\citep{gardner1988space,krauth1989storage}. Coding-theoretic analyses further show that, for Hebbian constructions and exact recovery of randomly chosen patterns, one typically cannot exceed $n/(4\ln n)$~\citep{MR901677} memories. More generally, Cover's classical separability theorem~\citep{cover1965geometrical} identifies a characteristic linear capacity scale of $2n$ for dense random classification problems. Nonetheless, superlinear capacity is achievable for certain structured datasets. For example, sparse data having few active neurons can yield an increase in capacity to nearly quadratic in $n$~\citep{tsodyks1988enhanced,amari1989characteristics}.
Additionally, robust exponential memory has been observed for particular examples of group-structured data~\citep{hillar2018robust,hillar2021hidden}, in particular for storing all $k$-clique graphs and their hypergraph analogues. Our work builds on the observations of \citet{hillar2018robust,hillar2021hidden} by proving that all graph-isomorphism classes are memorizable.

\textbf{Modern Hopfield Networks.} A line of recent investigation has sought to increase the capacity and retrieval properties of Hopfield networks by changing the energy function. Dense Associative Memories (DAMs) replace the classical quadratic energy with higher-order polynomial interactions, resulting in a capacity that scales polynomially with neuron count~\citep{krotov2016dense, horn1988capacities}. Building on this, Modern Hopfield Networks (MHNs) introduced a log-sum-exp energy function that allows the capacity to grow exponentially~\citep{ramsauer2020hopfield, demircigil2017model}. Our work provides a complementary perspective to these advancements by showing classical HNs can achieve exponential capacity by capturing the symmetry of the data in its parameters.  Preliminary experiments suggest that DAMs do not generalize well in this setting (see Fig.~\ref{fig:gen_dams}, Appendix~\ref{app:dams}).

\textbf{Generalization beyond the training set for HNs.} Theoretical study of classical HNs primarily focuses on storage limits, basins of attraction, and noise robustness around memorized patterns, rather than sample-complexity guarantees for generalization to patterns outside the training set.  Earlier analyses of concept generalization in classical HNs investigate when networks capture latent data regularities~\citep{fontanari1990generalization}. More recently, Random-Features Hopfield Models (RFHMs) exhibit learning phase transitions and even retrieval of previously unseen examples \citep{negri2023rfhm,kalaj2024rfhmgen}.  These results complement but are distinct from our own; in particular, they do not provide sample-complexity bounds or analyze the emergence of invariance induced by a symmetry present in the data. In addition, while these results primarily use techniques from statistical physics, here we leverage tools from statistical learning theory.

\textbf{Emergent invariance through data augmentation and feature averaging.}
One perspective on data augmentation is as orbit averaging over a symmetry group; in particular, empirical risk minimization on augmented data is equivalent to averaging features or predictions along group orbits. This procedure has been shown to induce approximate invariance and reduce estimator variance~\citep{chen2020gtda}. From a kernel perspective, augmentation decomposes into first-order invariant feature averaging plus a second-order variance regularizer~\citep{dao2019kernelaugment}. Enforcing invariance by feature averaging yields provable generalization benefits for invariant kernel methods~\citep{elesedy2021provably}. Beyond static estimators, recent results show that, with full-group augmentation, deep ensembles become equivariant in expectation at all training times in the infinite-width limit~\citep{gerken2024emergent}, and that the expected predictions of group-convolutional networks match those of data-augmented conventional networks throughout training~\citep{eNTK_arxiv2024}. A formal comparison between group-equivariant architectures and data augmentation, without taking the infinite-width limit, is provided by \citet{nordenfors2024optimizationdynamicsequivariantaugmented}. Finally, under certain conditions, the weights of a neural network that is invariant to a finite group recover the Fourier transform on that group, providing a signature of feature learning in group-structured data~\citep{pmlr-v247-marchetti24a}. These results primarily quantify the benefits of explicitly imposing a known symmetry, through architectural design or full-orbit averaging. We instead ask whether a standard norm-biased learner can recover approximate invariance from a small random subset of an orbit, and whether this learned invariance yields a guarantee that is uniform over the entire orbit.

\textbf{Implicit bias.} A large body of work shows that, even without explicit regularization, certain learning dynamics have a preference for particular solutions. In particular, for classification using the logistic loss, gradient descent drives the parameter norm to infinity while the parameter direction converges to the max-margin classifier \citep{soudry2018the,ji2019risk,nacson2019convergence}. We use this connection to identify the exact sampled HSVM as the limiting directional object for MEF-GD, and then characterize how this minimum-norm solution approaches invariance when the data form a permutation orbit.

\section{Preliminaries} \label{sec:prelims}

\paragraph{Notation:} We use capitalized boldface characters to denote matrices, bold lowercase characters to denote vectors and non-bold lowercase characters to denote scalar values. If $\vx \in \R^n$ is a vector then $x_i$ denotes the $i$th entry of $\vx$. If $\mX \in \R^{N \times n}$ then $\vx_i \in \R^n$ denotes the $i$th row of $\mX$ and to access individual entries of $\mX$ we use the notation $x_{ij}$ or $[\mX]_{ij}$, whichever is clearer in context. Whether a matrix, vector or scalar is deterministic or random is also inferred from context. We use $\Pi_a$ to denote the set of permutations on $[a]$ and $\cP_a$ to denote the set of $a \times a$ permutation matrices. Overloading our notation, we also refer to $\cP_a$ as the group of permutation matrices. If $\cH$ and $\cG$ are groups, we write $\cH\leq\cG$ to mean that $\cH$ is a subgroup of $\cG$. Finally, if $\cH$ is a group that acts on a set $\cA$, then the orbit of $a \in \cA$ under $\cH$ is denoted $\textnormal{Orb}(a, \cH) = \{ h a \in \cA: h \in \cH  \}$.

\paragraph{Associative Memory:} We consider a Hopfield network~\citep{hopfield1982neural} with asynchronous dynamics but do not restrict ourselves to Hebbian learning. To this end, let $\textnormal{Sym}_0^n \subset \R^{n \times n}$ denote the set of symmetric, real, $n \times n$ matrices whose diagonal entries are zero, and let $\Theta= \textnormal{Sym}_0^n \times \R^n$. Clearly $\Theta$ is a convex vector space. We introduce the energy function $E:\{ 0,1\}^n \times \mTheta \rightarrow \R$ defined as
\begin{equation} \label{eq:energy}
   E(\vx; \vtheta) = \frac{1}{2} \vx^T \mW \vx + \vb^T \vx,
\end{equation}
where $\vtheta = (\mW, \vb)\in \mTheta$. Given an input binary vector $\vx \in \{ 0,1\}^n$, the Hopfield network generates a sequence of binary vectors $(\vx(t))_{t \geq 0}$ through the following recurrence dynamics: with $\vx(0) = \vx$ then
\begin{equation}\label{eq:recurrence-dynamics}
x_j(t) = \begin{cases}
& \1(-\vw_j^T\vx(t-1)> b_j) \quad t \equiv j \pmod{n},\\
&x_j(t-1) \quad \text{otherwise}\\
\end{cases}
\end{equation}
for all $t\geq 1$ and $j\in[n]$. For any input $\vx \in \{0,1\}^n$, this sequence converges in finite time to a fixed point~\citep{bruck-hopfiel-convergence}. We define the input-output map of the Hopfield network, denoted $H: \{0,1\}^n \times \Theta \rightarrow \{0,1\}^n$, as follows: given parameters $\vtheta$ and an input $\vx$, the output $H(\vx;\vtheta)$ is the attractor or fixed point of~\eqref{eq:recurrence-dynamics} reached when initialized with $\vx(0) = \vx$. If $H(\vx;\vtheta) = \vx$, then $\vx$ is a fixed point of the recurrence dynamics; in this setting, we say that the function $H_{\vtheta}(\cdot) := H(\cdot;\vtheta)$ has \textit{memorized} $\vx$. A sufficient condition for $H_{\vtheta}$ to memorize $\vx$ is $E(\vx;\vtheta) < E(\vx';\vtheta)$ for all $\vx' \in \cN(\vx)$, where $\cN(\vx)$ denotes the set of all binary vectors at Hamming distance exactly one from $\vx$. Under this condition, we say that $H_{\vtheta}$ \textit{strictly memorizes} $\vx$. We also denote the action of a permutation matrix $\mP \in \cP_n$ on the parameters of a Hopfield network by $\mP\vtheta \coloneq (\mP^T\mW\mP,\mP^T\vb)$.

\paragraph{Training and memorization:} Let $\cS \subset \{0,1\}^n$. We say that $H_{\vtheta}$ memorizes $\cS$ if it memorizes every $\vx \in \cS$. Many methods have been proposed to train networks to memorize a set $\cS$~\citep{Hertz91,Tolmachev20}; see Appendix~\ref{app:learning-rules}. We focus on minimization of the energy flow (MEF)~\citep{mpf-hopf,hillar2018robust,hillar2021hidden} and study its implicit bias. If $\vx \in \{0,1\}^n$ and $j \in [n]$, let $\vx^{(j)} \in \{0,1\}^n$ satisfy $x_l \neq x^{(j)}_l$ if and only if $l=j$. We define the \textit{energy flow} loss as
\begin{equation}\label{eq:MEF}
L(\vtheta; \cS) = \sum_{\vx \in \cS} \sum_{j=1}^n \exp \left( E(\vx; \vtheta) - E(\vx^{(j)}; \vtheta) \right).
\end{equation}
For any given set of points $\cS$, note that $L$ is nonnegative, infinitely differentiable, and convex in $\Theta$. As a result, minimizing $L$ is a convex problem to which a wide variety of numerical techniques can be applied, including variants of gradient descent (GD) and approximate second-order methods such as L-BFGS~\citep{liu1989limited}. As long as $\cS$ can be strictly memorized, sufficiently minimizing~\eqref{eq:MEF} will result in a network that memorizes $\cS$. For further details on MEF learning for discrete recurrent neural networks, we refer the reader to~\citet{hillar2021hidden} and its inspiration from the theory of density estimation~\citep{PhysRevLett.107.220601}.

\section{Implicit bias, minimum norm memorizers and generalization} \label{sec:implicit-bias}
Before specializing to datasets drawn from graph-isomorphism classes, we connect memorization to a linear margin problem and identify the implicit bias of MEF. This enables us to provide generalization guarantees for memorization in Hopfield networks, as stated in Theorem~\ref{lem:gen-svm-bounds}. Given $\vtheta = (\mW, \vb) \in \Theta$, for any $j \in [n]$ we define the vector $\vtheta_j = [\vw_j,b_j] \in \R^{n+1}$. Overloading our notation, we also use $\vtheta = [\vtheta_1,\vtheta_2,\ldots,\vtheta_n] \in \R^{n(n+1)}$ to refer to the flattened vector of all the network parameters $(\mW,\vb)$. For any $\vx \in \{0,1\}^{n}$, let $\vz(\vx) = [\vx,1] \in \{0,1\}^{n+1}$ and $y_j(\vx) = 1-2x_j \in \{\pm 1\}$. Using this notation, the energy difference between a point and one of its neighbors is
\begin{equation}
    E(\vx^{(j)}; \vtheta ) - E(\vx; \vtheta ) = y_j(\vx) \langle \vz(\vx), \vtheta_j \rangle,
\end{equation}
see Appendix~\ref{app:props-energy-function} for further details. Consequently, parameters that strictly memorize a set $\cS\subseteq\{0,1\}^n$ must satisfy a system of linear inequalities: there must exist some $\epsilon>0$, referred to as the functional margin, such that $y_j(\vx)\langle\vz(\vx),\vtheta_j\rangle\geq\epsilon$ for all $\vx\in\cS$ and $j\in[n]$. The energy function~\eqref{eq:energy} is quadratic in $\vx$ but linear in the parameters. Moreover, $E(\vx;a\vtheta)=aE(\vx;\vtheta)$, so the energy is positively homogeneous of degree one in the parameters. Hence the set of attractors is invariant under positive rescaling of the parameters. Without loss of generality, we select a functional margin of one and define the feasible set of parameters up to positive rescaling as
\begin{equation}\label{eq:feasible}
\cF_{\theta}(\cS) = \{ \vtheta \in \Theta: y_j(\vx) \langle \vz(\vx), \vtheta_j \rangle \geq 1 \;\; \forall \vx \in \cS, \forall j \in [n] \}
\end{equation}
In addition, the inequality constraints that define $\cF(\cS)$ can be written with respect to a single vector of unique parameters, which we denote $\vomega$. To this end, let $p = \tfrac{n(n-1)}{2}$ and $q = \tfrac{n(n+1)}{2}$. There exists a matrix $\mV \in \{0,1,1/\sqrt{2}\}^{n(n+1) \times q}$ such that, for any $\vtheta \in \Theta$, there exists an $\va \in \R^p$ with $\vomega = [\sqrt{2}\va,\vb] \in \R^q$ and $\vtheta = \mV\vomega$. In short, $\mV$ copies the unique elements---the upper-triangular elements of $\mW$ and the biases $\vb$---into their appropriate locations in the flattened vector $\vtheta$. For any $j \in [n]$, let $\mV_j \in \R^{(n+1) \times q}$ denote the submatrix of rows of $\mV$ such that $\vtheta_j = \mV_j\vomega$. For any $\vx \in \{0,1\}^n$ and $j \in [n]$, let $\vu_j(\vx) = y_j(\vx)\mV_j^T\vz(\vx)$. Then each constraint can be rewritten as $y_j(\vx) \langle \vz(\vx),\vtheta_j \rangle = \langle \vu_j(\vx),\vomega \rangle$, and thus we can equivalently define the feasible set as
\begin{equation}\label{eq:feasible2}
\cF_{\omega}(\cS) = \{ \vomega\in \R^q: \langle \vu_j(\vx), \vomega \rangle \geq 1 \;\; \forall \vx \in \cS, \forall j \in [n] \}
\end{equation}
Inspecting~\eqref{eq:feasible2}, clearly strict memorization of a dataset is equivalent to solving a linear program (LP) and therefore any algorithm which successfully memorizes $\cS$ is implicitly solving an LP. Moreover, these algorithms may have an \textit{implicit bias} towards feasible points or solutions which satisfy other conditions or criteria.
A popular and well-studied example is the feasible point with the smallest norm: identifying it requires solving a quadratic program (QP), or more specifically a hard-margin support vector machine (HSVM) problem. If $\vtheta=\mV\vomega$, where $\vomega=[\sqrt{2}\va,\vb]$, then $\|\vtheta\|^2=\|\mW\|_F^2+\|\vb\|^2=\|\sqrt{2}\va\|^2+\|\vb\|^2=\|\vomega\|^2$. Thus finding the minimum-norm feasible point for a set $\cS\subset\{0,1\}^n$ is equivalent to solving
\begin{equation}\label{eq:HSVM}
    \textnormal{HSVM}(\cS) = \argmin_{\vomega \in \R^q} \| \vomega \|^2 \;\; s.t. \;\; \vomega \in \cF_{\omega} (\cS).
\end{equation}
The key takeaway of this section is that minimizing~\eqref{eq:MEF} with gradient descent (GD) is implicitly biased in direction towards the solution of~\eqref{eq:HSVM}, i.e., norm-minimization. To this end, first observe that~\eqref{eq:MEF} can be re-parameterized as
\begin{align*}
L(\vtheta; \cS) = \sum_{\vx \in \cS} \sum_{j=1}^n \exp \left( E(\vx; \vtheta) - E(\vx^{(j)}; \vtheta) \right)
= \sum_{\vx \in \cS} \sum_{j=1}^n \exp \left( -\langle \vu_j(\vx) , \vomega\rangle \right)
=: L(\vomega; \cS).
\end{align*}
Consider now updates to the parameters of the Hopfield network using GD: in particular, given initial parameters $\vomega(0)$ and step-size $\eta >0$, for all $t \geq 0$ let
\begin{equation} \label{eq:MEF-GD-update}
    \vomega(t+1) = \vomega(t) + \eta \sum_{\vx \in \cS} \sum_{j=1}^n  \exp \left( -\vu_{j}(\vx)^T \vomega(t) \right)\vu_{j}(\vx).
\end{equation}
Using~\citet[Thm.~3]{soudry2018the}, it can be proved that this sequence of GD iterates converges in direction to the solution of~\eqref{eq:HSVM}. Indeed, \citet[Thm.~3]{soudry2018the} directly implies the following in our setting.
\begin{corollary}\label{cor:MEF-SVM}\textbf{\citep[Thm.~3]{soudry2018the}}
    Assume $\cS$ can be strictly memorized, let $\vomega^* = \textnormal{HSVM}(\cS)$, $\vomega(0) \in \R^q$ be arbitrary and $\vomega(t)$ be generated for all $t \in \N_{\geq 1}$ as per~\eqref{eq:MEF-GD-update}. There exists a choice of step size $\eta$ such that $\vomega(t) = \vomega^* \log(t) + \rho(t)$ for all $t \in \N_{\geq 1}$,
    where $\rho(t)$ grows as $\| \rho(t)\| = O(\log(\log(t)))$. Moreover, $\lim_{t \rightarrow \infty} \tfrac{\vomega(t)}{\| \vomega(t)\|} = \tfrac{\vomega^*}{\| \vomega^*\|}.$
\end{corollary}
Informally, Corollary~\ref{cor:MEF-SVM} states that the solution returned by minimizing the energy flow with gradient descent (MEF-GD) after exponentially many iterations is a close approximation directionally to the solution returned by solving the HSVM problem~\eqref{eq:HSVM}. We now derive generalization bounds both for the HSVM solution and MEF with GD.

We present the following result as a general distributional benchmark: the stronger parameter-convergence and support-wide guarantees developed in Section~\ref{sec:invariance} arise only after exploiting the orbit structure.

\begin{restatable}{theorem}{GenSvmBounds}\label{lem:gen-svm-bounds}
    Let $D \subset \{0,1\}^n$ be a set which can be strictly memorized and assume $\| \vx\|_0 \leq m \in \N_{\geq 1}$ for all $\vx \in D$. Let $\mathcal{D}$ be a probability distribution on $D$, and consider a random sample $\cS_N = (\vx_i)_{i \in [N]}$, where $\vx_i \sim \cD$ are mutually i.i.d. Let $\hat{\vomega} = \textnormal{HSVM}(\cS_N)$, $\vomega^* = \textnormal{HSVM}(D)$, $\vomega(0) \in \R^q$ be arbitrary and $\vomega(t)$ be generated for all $t \in \N_{\geq 1}$ as per~\eqref{eq:MEF-GD-update}, $\delta, \epsilon \in (0,1)$ and assume $\vx \sim \cD$ is sampled independent of $\cS_N$. If $N \gtrsim  \epsilon^{-2}n\|\vomega^* \|^2 m \log(1/\delta)$ then both
    \begin{align*}
    \P(H(\vx; \mV \hat{\vomega}) \neq \vx) \leq \epsilon \quad \text{and} \quad
    \P(H(\vx; \mV\vomega(t)) \neq \vx)  = \tilde{O}\left( \frac{ \sqrt{m} \| \vomega^* \|}{\log(t)}\right) +  \epsilon
    \end{align*}
    hold with probability at least $1-\delta$ over the sample $\cS_N$.
\end{restatable}
The $\tilde{O}(\cdot)$ notation hides a factor of $\log(\log(t))$ arising in~\citep[Thm.~5]{soudry2018the}. To prove Theorem~\ref{lem:gen-svm-bounds}, we combine a vector contraction inequality~\citep[Corollary~1]{vector-contrac} with Rademacher bounds; see, e.g.,~\citep[Theorem~3.3]{FoML}. Appendix~\ref{app:lem:gen-svm-bounds} gives the full proof. This theorem applies to an arbitrary distribution on any sparse, strictly memorizable set: it bounds the probability of failing to memorize one fresh independent sample rather than guaranteeing simultaneous memorization of the support. In Section~\ref{sec:invariance}, additional orbit structure and uniform sampling yield the latter, stronger conclusion for the exact HSVM. The MEF bound inherited from \citet{soudry2018the} may require exponentially many iterations to give a prescribed directional accuracy. This is a tail phenomenon: once the weights are approximately aligned, points with a significant margin have exponentially small loss contributions.

\section{Storing isomorphism classes of graphs} \label{sec:iso-class-storage}

\subsection{Graph encoding} \label{sec:encoding-graph-data}
Let $\cG_v$ denote the set of all simple, undirected graphs on $v \in \N$ vertices. Recall two graphs $G = (V, E)$, $G' = (V', E')$ are isomorphic, which we denote $G \cong G'$, if there exists a bijection $\phi: V \rightarrow V'$ such that $(\phi(\nu_1), \phi(\nu_2)) \in E'$ if and only if $(\nu_1, \nu_2) \in E$. We refer to such a $\phi$ as an isomorphism between $G$ and $G'$. Note when $V = V'$, as is the case here since $V = V' =[v]$, then $\phi$ is a permutation. The \textit{isomorphism class} of a graph $G \in \cG_v$ is defined as $\cI(G) \coloneq \{G' \in \cG_v: G' \cong G \}$. Let $\cV_2$ denote the set of unordered pairs of $[v]$, $n \coloneqq |\cV_2| = \binom{v}{2}$ and $\text{Ind}: \cV_2 \rightarrow [n]$ be a bijection which indexes the elements of $\cV_2$.

\begin{definition}[Edge representation of a graph]
    Let $\cE_{rep}: \cG_v \rightarrow \{0,1 \}^n$ be defined as follows: if $G = ([v], E) \in \cG_v$ and $\vx = \cE_{rep}(G)$ then, for all $j \in [n]$, $x_j \coloneqq \1(\text{Ind}^{-1}(j) \in E)$. We refer to $\vx$ as the edge representation of $G$.
\end{definition}

To be clear, $\cE_{rep}$ is a bijection which assigns each graph to a binary vector of dimension $n$ whose support defines the vertex pairs present in the edge set of the graph in question. Any vertex permutation induces an edge permutation.

\begin{definition}[Edge permutation induced by a vertex permutation] \label{def:vertex-induced-edge-perm}
    Let $\phi: [v] \rightarrow [v]$ be a permutation. The edge permutation $\pi_{\phi}: [n] \rightarrow [n]$ induced by $\phi$ is defined as follows: if $\textnormal{Ind}^{-1}(j) = (\nu_1, \nu_2)$ then  $\pi_{\phi}(j) = \textnormal{Ind}((\phi(\nu_1), \phi(\nu_2)))$. We denote the subset of these edge permutations as $\Pi_n^{\Phi}$ and the corresponding subset of permutation matrices as $\Phi_n$.
\end{definition}

We now claim the following: first $\Phi_n$ is a subgroup of $\cP_n$, second if two graphs $G,G' \in \cG_v$ are isomorphic then there is a vertex induced edge permutation which maps between their edge representations, and third, for any $G \in \cG_v$ we have $\cE_{rep}(\cI(G)) = \textnormal{Orb}(\cE_{rep}(G), \Phi_n)$. For proofs of these claims we refer the reader to Appendix~\ref{app:sec:encoding-graph-data}. Two edges are said to be \textit{adjacent} if they share a vertex in common: more specifically, if $j,l \in [n]$ then $j$ and $l$ are \textit{adjacent}, which we denote $j \sim l$, if and only if $|\text{Ind}^{-1}(j) \cap \text{Ind}^{-1}(l)| = 1$, otherwise we say $j$ and $l$ are not adjacent, which we denote $j \nsim l$. We now identify a subset of edge permutations which are characterized by preserving edge adjacency.

\begin{definition}[Edge adjacency preserving permutation]
    A permutation $\pi: [n] \rightarrow [n]$ \textit{preserves edge adjacency} if $\pi(j) \sim \pi(l)$ if and only if $j\sim l$. We denote such permutations as $\Pi_n^{\cQ}$ and the corresponding set of permutation matrices as $\cQ_n$.
\end{definition}

Similar to $\Phi_n$, this subset forms a subgroup of $\cP_n$. Moreover $\Phi_n$ is a subgroup of $\cQ_n$ and as a result $\cE_{rep}(\cI(G)) \subset \textnormal{Orb}(\cE_{rep}(G), \cQ_n)$. Again we refer the reader to Appendix~\ref{app:sec:encoding-graph-data} for further details. As a result, the edge representations of the isomorphism class of a graph are a subset of the orbit of the edge representation of the graph in question under edge adjacency preserving permutations.

\subsection{Experiments on graph data}\label{sec:experiments-graphs}
To assess storage across isomorphism classes experimentally, we study clique, bipartite, quadratic-residue, and Johnson graphs. Clique graphs, or more specifically $k$-cliques, contain a fully connected subset of $k$ vertices while the remaining $v-k$ vertices are isolated. Bipartite graphs split the $v$ vertices into two equally sized groups, with all possible inter-group edges present and no intra-group edges. The experimental code constructs modular quadratic-residue graphs using the NetworkX routine~\citep{networkx}; cf.~\citet{bollobás2001randomgraphs}. The original plots label these graphs ``Paley.'' We retain that label inside the figures but refer to the family as quadratic-residue graphs in the text: for orders outside the standard Paley regime, the plotted label is an abuse of notation. This naming issue is immaterial to the experiment, and no theoretical claim relies on a Paley-specific property. For integers $n,r$, the Johnson graph $J(n,r)$ has as vertices the $r$-element subsets of $[n]$, with two vertices adjacent when the corresponding subsets differ in exactly one element.\footnote{Equivalently, their intersection has size $r-1$.} These families provide structurally distinct, easily sampled test cases with substantial symmetry. Extensive experiments for cliques are already provided in~\citet{hillar2018robust,hillar2021hidden}; we include them here for comparison and completeness. We selected these classes because they are easy to sample and enumerate, and regard them as representative rather than special. We observe similar behavior for other graph-isomorphism classes, including random graphs. The reproducibility statement provides a link to our code. Further experiments and preliminary results on the Hidden Clique Problem (Fig.~\ref{fig:HCP}), comparison with Dense Associative Memories (DAMs) (Fig.~\ref{fig:gen_dams}), orbit generalization for other graph families (Figs.~\ref{fig:other_families} and~\ref{fig:other_families2}), and double-descent phenomena (Fig.~\ref{fig:ddescent}) appear in Appendix~\ref{app:add-exp}.

Fig.~\ref{fig:accuracy_combined} shows test accuracy versus training sample size, with the mean and minimum--maximum range over $10$ trials, for Hopfield networks trained by MEF, Perceptron, and Delta (the latter two are used only as baselines; see Appendix~\ref{app:learning-rules}). For small graphs ($v=8$), we enumerate the full isomorphism class and report the fraction memorized. For larger graphs ($v=20$), accuracy is estimated on an independent random sample of $1{,}000$ graphs. The figures appear to support two observations: (i) MEF reaches higher test accuracy with fewer samples than the other methods in these experiments, despite all methods memorizing their training sets; and (ii) for MEF and Delta, the number of samples required is small relative to the class size. The optimization runs were capped at $1{,}000$ iterations, so these experiments also suggest that finite-iteration behavior can be considerably better than the generic asymptotic comparison predicts. We do not interpret these finite-range observations as asymptotic rates. Appendix~\ref{app:add-exp} reports analogous experiments for other graph types.

\begin{figure}[!htb]
    \centering
    \setlength{\tabcolsep}{3pt}
    \begin{tabular}{c}
        \includegraphics[width=.49\textwidth]{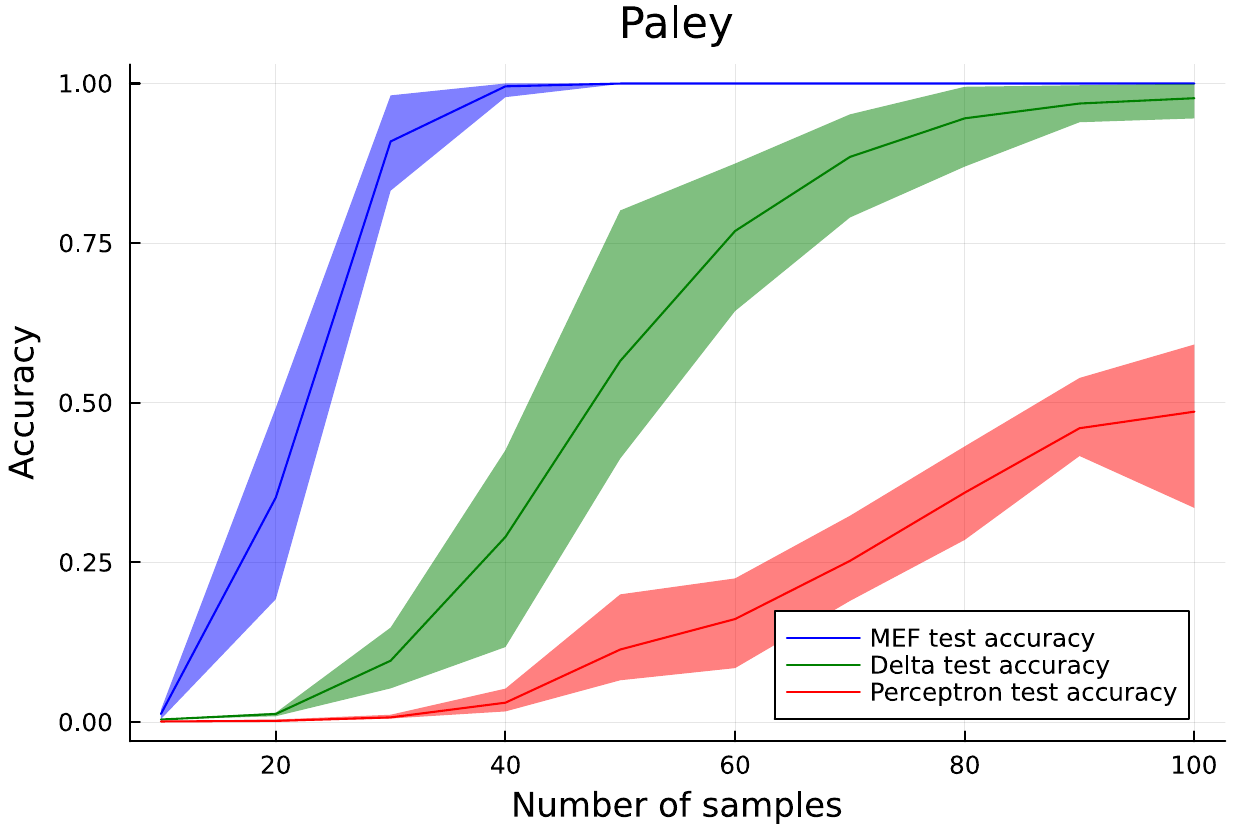}
        \includegraphics[width=.49\textwidth]{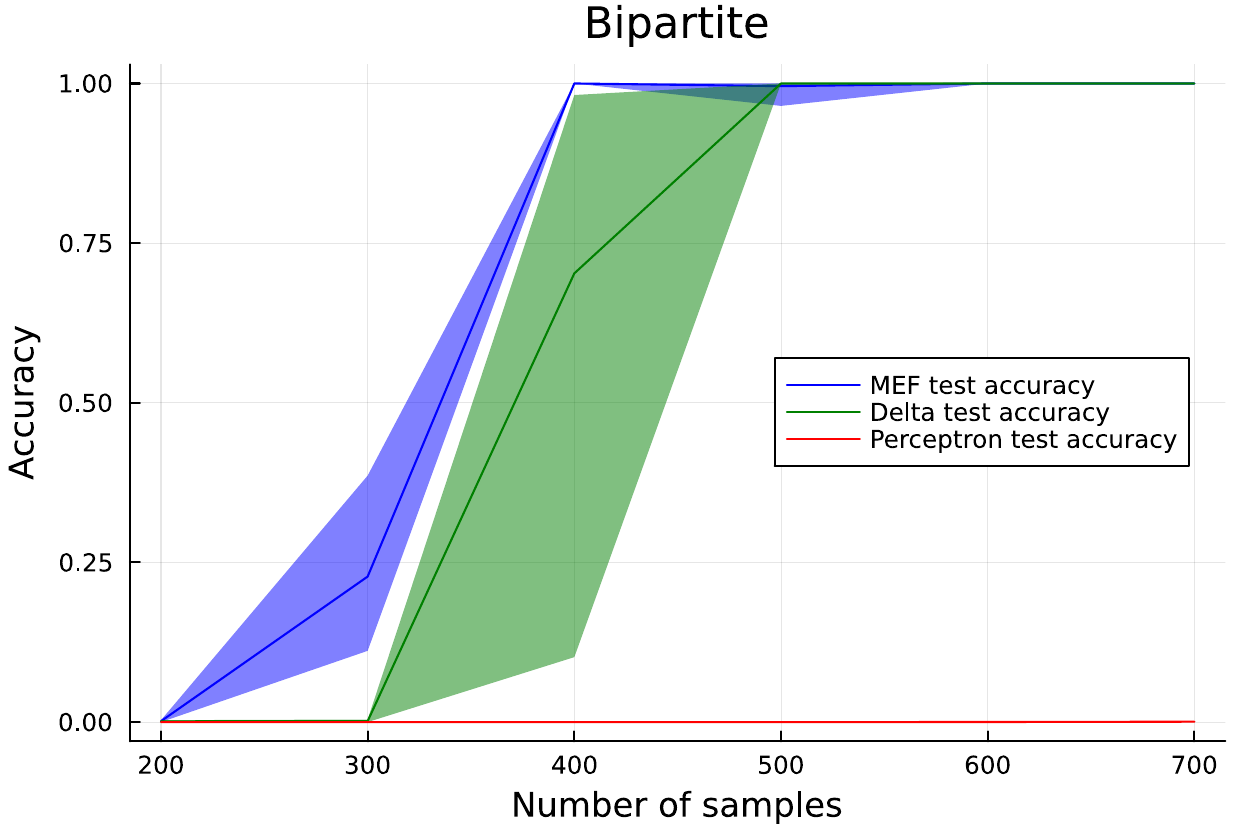} \\
    \end{tabular}
    \caption{\textbf{Test accuracy vs.\ training sample size for isomorphism classes at two scales.} Left: $v=8$; the quadratic-residue class (labelled ``Paley'' in the plot) has size $2{,}520$. Right: $v=20$; the bipartite class has size $92{,}378$. Curves show the mean and minimum--maximum range over $10$ trials. Networks are trained with the Perceptron, Delta (MSE), and MEF learning rules.}
    \label{fig:accuracy_combined}
\end{figure}

\begin{figure}[!htb]
    \centering
    \includegraphics[scale=0.38]{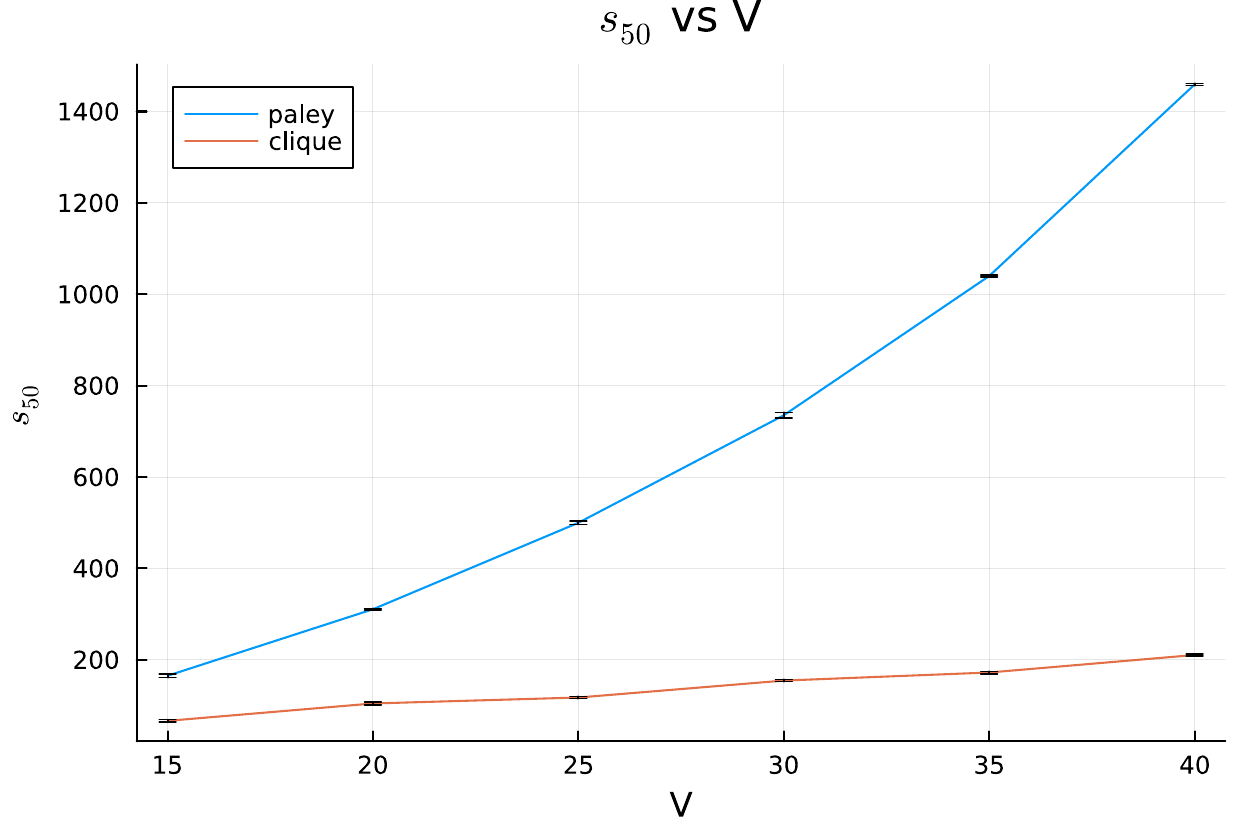}
    \includegraphics[scale=0.38]{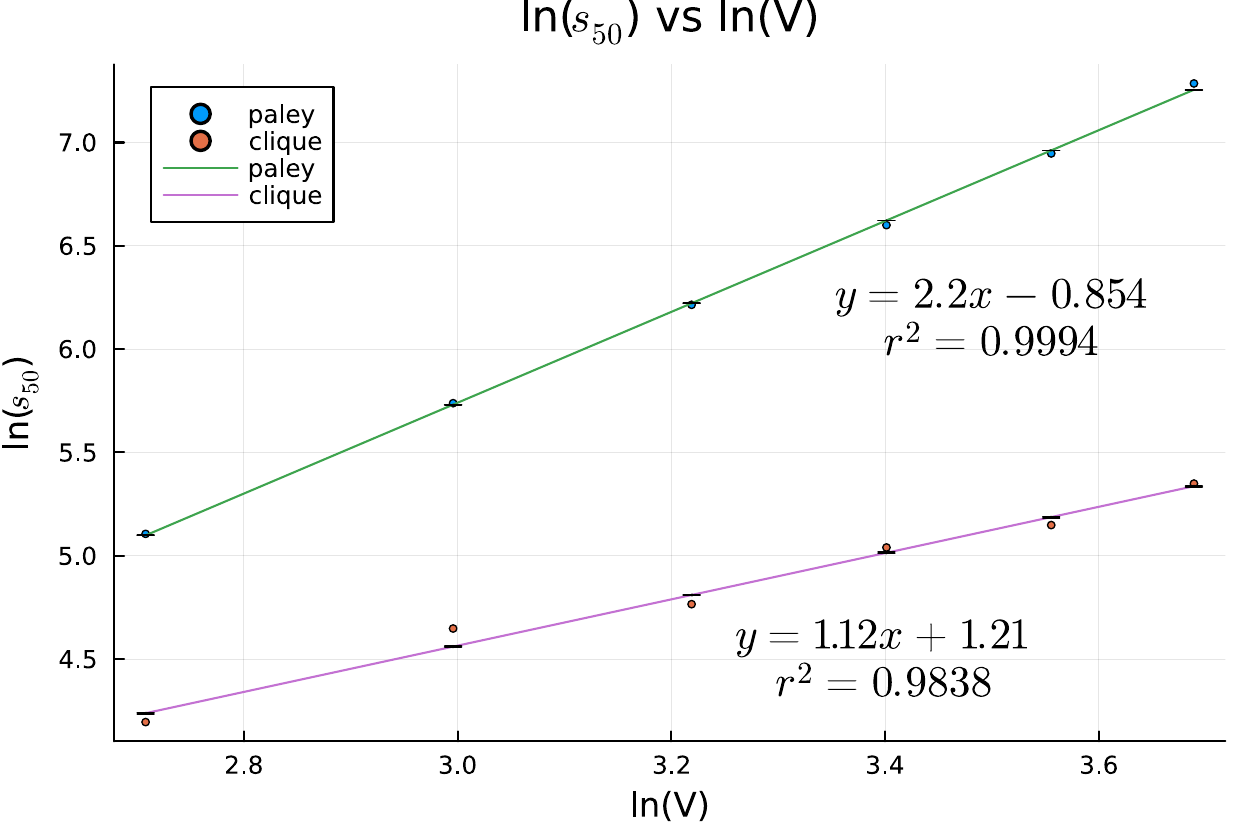}
    \caption{\textbf{Sample-complexity scaling.} Number of training samples $s_{50}$ required for a Hopfield network trained by MEF with accelerated gradient descent to memorize $50\%$ of an independent sample of $1{,}000$ graphs drawn from clique and quadratic-residue isomorphism classes on $v$ vertices. The quadratic-residue family is labelled ``Paley'' in the plots. Right: $\ln(s_{50})$ versus $\ln(v)$ with least-squares fits.
    }
    \label{fig:scaling}
\end{figure}

Fig.~\ref{fig:scaling} empirically compares the scaling of the sample count required to learn $k$-cliques and quadratic-residue graphs (labelled ``Paley'' in the plots). This comparison illustrates that the difficulty of learning an isomorphism class may depend on its connectivity structure. For each graph size $v$, we record
$s_{50}$, the smallest training sample size for which MEF attains at least $50\%$ average test accuracy on test samples of size $1{,}000$, averaged over $10$ trials. The left subplot shows $s_{50}$ versus $v$, and the right subplot shows a log--log fit. Assuming $s_{50}=Cv^p$ for some constant $C\in\R_{>0}$, we estimate $p$ by linear regression. The fitted slopes are approximately $2.2$ for the quadratic-residue family and $1.12$ for cliques. Over the range studied, these figures appear to support a roughly quadratic dependence for the former and a roughly linear dependence for the latter. These are empirical finite-range fits rather than asymptotic bounds.
Thus, although Hopfield networks can memorize every class considered here (Lemma~\ref{lemma:express-inv-param}), the observed sample requirement varies with graph connectivity.

\begin{figure}[!htb]
    \centering
    \includegraphics[scale=0.49]{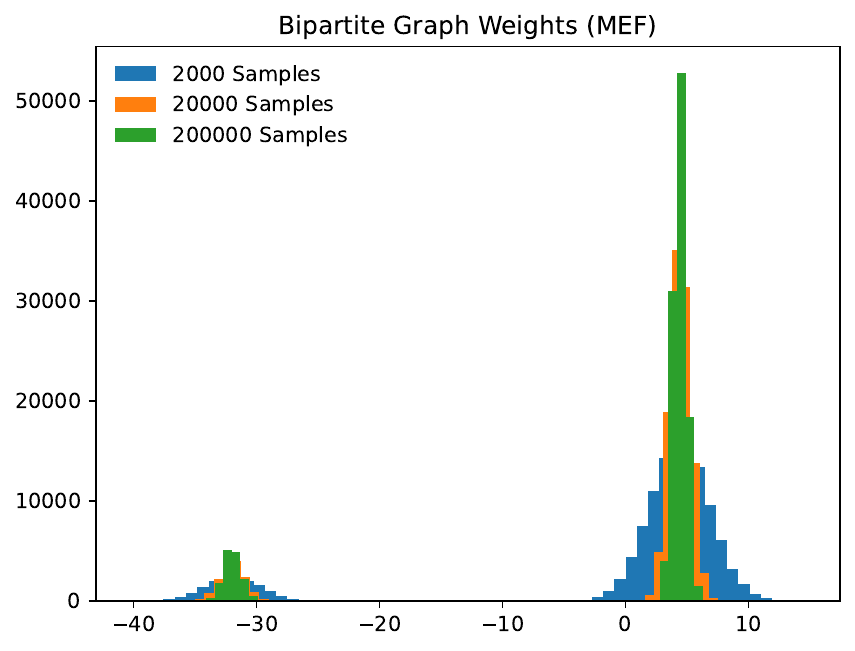}
    \includegraphics[scale=0.49]{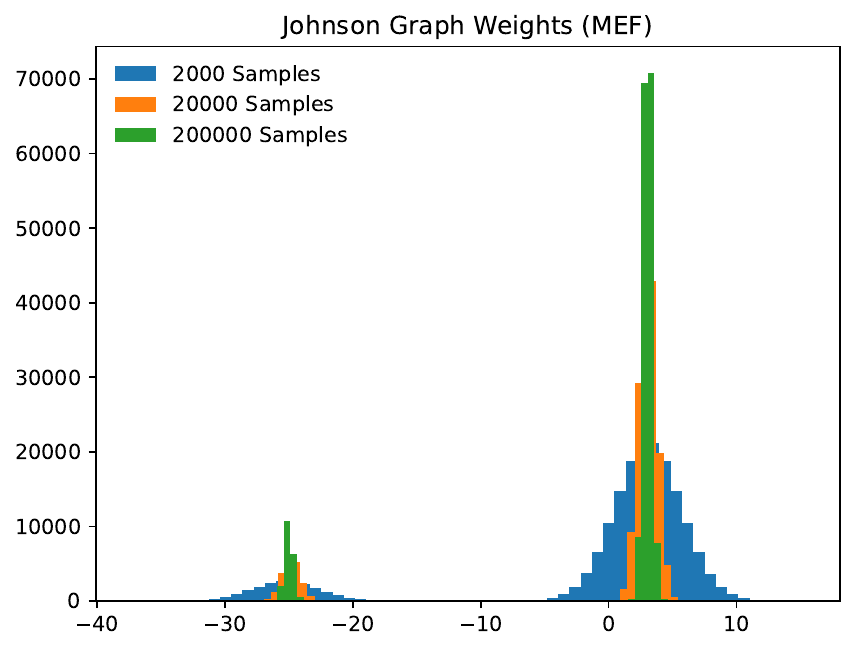}
    \caption{\textbf{Learned weights approach invariant subspace as sample size increases.} Histograms of MEF-learned weights over 3 sample counts. Left: Bipartite graph with $v=32$, $k=16$ (496-bit). Right: Johnson graph $J(7,3)$ on $v=35$ vertices (595-bit). Parameters are normalized so that thresholds (not shown) have mean absolute value $1$.}
    \label{fig:weight_hists}
\end{figure}

Fig.~\ref{fig:weight_hists} shows histograms of the elements of weight matrices of Hopfield networks trained on isomorphism classes of graphs, namely Bipartite and Johnson graphs, for three different sample sizes. Indeed, it is apparent that as the sample size grows the parameters returned by the optimizer converge onto a distinct subspace: we identify this subspace as the parameters invariant to the underlying group action of the data in Lemma~\ref{lem:inv-params-cond} below. Heatmaps further supporting this conclusion can be observed in Figure~\ref{fig:weight_heatmaps2} in Appendix~\ref{app:add-exp}.

\subsection{Invariant parameters}\label{sec:invariance}
In what follows let $\Gamma_n$ denote an arbitrary subgroup of $\cP_n$. For any $\mQ \in \cP_n$ and $\vtheta = (\mW, \vb) \in \mTheta$ recall that we define the action of $\mQ$ on the parameter $\vtheta$ as $\mQ \vtheta \coloneqq (\mQ^T \mW \mQ, \mQ^T \vb)$. Let $\mQ \in \cP_n$, $\vtheta = (\mW, \vb) \in \Theta$ and $\vtheta' = \mQ \vtheta = (\mW', \vb')$. Note $\mW'^T = (\mQ^T \mW \mQ)^T = \mQ^T \mW \mQ = \mW'$ and for all $j \in [n]$ we have $W_{jj}' = \ve_{\pi(j)}^T\mW \ve_{\pi(j)} = W_{\pi(j) \pi(j)} = 0$. As a result $\mW' \in \textnormal{Sym}_0^n$, in addition trivially $\vb' \in \R^n$ and therefore $\vtheta' \in \Theta$. As a result, the action of $\cP_n$, or any subgroup $\Gamma_n$ of $\cP_n$, on $\Theta$ is well defined.

\begin{definition}[Parameter invariance to the action of a subgroup] \label{def:parmas-inv-graph-iso}
A parameter $\vtheta\in\Theta$ is invariant with respect to $\Gamma_n$ if $\mQ\vtheta=\vtheta$ for every $\mQ\in\Gamma_n$. We denote the set of these parameters by $\Psi(\Gamma_n)$.
\end{definition}
Recall $E(\mQ \vx; \vtheta)= \frac{1}{2} \vx^T (\mQ^T \mW \mQ) \vx + (\mQ^T \vb)^T \vx$. If $\vtheta \in \Psi(\Gamma_n)$, then for any $\vx \in \{0,1 \}^n$ and $\mQ \in \Gamma_n$ we have
\begin{equation}\label{eq:intertwining}
     E(\mQ \vx; \vtheta)
         = E(\vx; \mQ \vtheta)
         = E(\vx; \vtheta).
\end{equation}
We refer to~\eqref{eq:intertwining} as the intertwining property of the energy function. Using this property, the following lemma extends energy difference bounds between neighbors from a point to an orbit.

\begin{restatable}{lemma}{LemLB}\label{lemma:gii-lb}
    Let $\vx_0 \in \{0,1\}^n$ and $\vtheta \in \Psi(\Gamma_n)$. For $\delta \in \R$, if $ E(\vx_0^{(j)}; \vtheta) - E(\vx_0; \vtheta)  \geq 1 - \delta$ for all $j \in [n]$, then for all $\vx \in \textnormal{Orb}(\vx_0, \Gamma_n)$ it follows that $ E(\vx^{(j)}; \vtheta) - E(\vx; \vtheta) \geq 1 - \delta$ for all $j \in [n]$.
\end{restatable}
Appendix~\ref{app:invariance} proves this lemma and the other results in this section. A key implication of Lemma~\ref{lemma:gii-lb} is that, if $\vtheta\in\Psi(\Gamma_n)$ strictly memorizes $\vx_0\in\{0,1\}^n$, then it strictly memorizes $\textnormal{Orb}(\vx_0,\Gamma_n)$. We next show that invariance under edge-adjacency-preserving permutations, which include vertex-induced graph-isomorphism permutations, corresponds to a particular rank-three parameter subspace.
\begin{restatable}{lemma}{LemInvParamsCond}\label{lem:inv-params-cond}
    Assume $v\geq 4$ and $n=\binom{v}{2}$. Let $F: \R^3 \rightarrow \mTheta$ denote the linear map defined as follows: if $(\mW, \vb ) = F(\vbeta)$ then for all $i,j \in [n]$ we have $w_{ij}=0$ if $i=j$, $w_{ij}=\beta_1$ if $i\sim j$, $w_{ij}=\beta_2$ if $i\neq j$ and $i \nsim j$, and $b_j = \beta_3$. Then
    \[
        \Psi(\Phi_n)=\Psi(\cQ_n)=F(\R^3).
    \]
\end{restatable}

Here $\Phi_n$ is the vertex-induced edge action that generates graph-isomorphism orbits. The possibly larger group $\cQ_n$ need not have the same group elements or data orbits; Lemma~\ref{lem:inv-params-cond} asserts only that the two actions have the same fixed parameter space. By inspection, the parameter patterns observed in Figs.~\ref{fig:weight_hists}, \ref{fig:weight_heatmaps2}, \ref{fig:HCP}b, \ref{fig:other_families2}bd for MEF appear to approximately lie in this space. We next show that invariant parameters can memorize any graph orbit.

\begin{restatable}{lemma}{ExpressInvParam}\label{lemma:express-inv-param}
    For $m \in [0,n]$, let $\vbeta = [2,2,1-2m] \in \R^3$ and $\vtheta = F(\vbeta)$. Then $E(\vx^{(j)}; \vtheta) - E(\vx; \vtheta) \geq 1$ for all $j \in [n]$ and for all $\vx \in \{0,1 \}^n$ satisfying $ \| \vx\|_{0} = m$.
\end{restatable}

Combining Lemmas~\ref{lemma:express-inv-param} and~\ref{lemma:gii-lb}, we conclude that any graph-isomorphism class can be strictly memorized by a Hopfield network. The only statistic used by the construction in Lemma~\ref{lemma:express-inv-param} is the sparsity of the representation: in fact, this construction memorizes $\vx \in \{0,1\}^n$ if and only if $\|\vx\|_{0}=m$. Consequently, if the goal is to memorize an isomorphism class while avoiding spurious memories, this is a poor parameter candidate. Moreover, when $m=\Theta(n)$, the norm of this construction grows as $\Theta(n^{3/2})=\Theta(v^3)$. For specific graph-isomorphism classes, solutions with much smaller norm exist. As an example, we consider $k$-cliques: for typographical ease, we denote the set of binary representations of $k$-cliques on $v$ vertices by $\cC_{v,k}$.

\begin{restatable}{lemma}{NormKClique}\label{lem:gii-strict-mem-ex}
If $\vbeta = [-10/k, 38/k^2, 0] \in \R^3$, $\vtheta = F(\vbeta)$, and $k \geq 5$, then the following hold.
\begin{enumerate}
\item $E(\vx^{(j)};\vtheta)-E(\vx;\vtheta) \geq 1$ for all $\vx \in \cC_{v,k}$ and all $j \in [n]$.
\item If $k=cv$ for some constant $c \in (0,1]$, then there exists a constant $C>0$ such that $\|\vtheta\|^2 \leq Cv$.
\end{enumerate}
\end{restatable}

Lemma~\ref{lem:gii-strict-mem-ex} shows that a parameter exists that strictly memorizes $\cC_{v,k}$ with norm only $O(\sqrt{v})$, rather than $\Theta(v^3)$ for the general construction when $m=\Theta(n)$. The construction in Lemma~\ref{lem:gii-strict-mem-ex} also does not memorize all $m$-sparse binary vectors. For example, fix $j \in [n]$ and suppose $\vx$ satisfies $\|\vx\|_0=m\leq 2(v-2)$ and $l\sim j$ for every $l\in\operatorname{supp}(\vx)$. Then $\vu_j(\vx)^T\vomega=-10m/k$, so $\vx$ is not strictly memorized. We speculate that the parameter norm may be related to the number of spurious memories, but leave a proper investigation to future work.

Our experiments suggest that memorization of a graph isomorphism class occurs when the minimum-norm solution is forced close to the invariant set. The following lemma identifies the invariant population target.

\begin{restatable}{lemma}{OrbitHSVMInv}\label{lemma:orbit-hsvm-inv}
    Let $\vx_0 \in \{0,1\}^n$ and $\Gamma_n$ denote a subgroup of $\cP_n$ and assume $\textnormal{Orb}(\vx_0, \Gamma_n)$ can be strictly memorized. If $\vomega^*=\textnormal{HSVM}(\textnormal{Orb}(\vx_0, \Gamma_n))$ and $\vtheta^*=\mV\vomega^*$, then $\vtheta^* \in \Psi(\Gamma_n)$.
\end{restatable}

\subsection{Finite-sample convergence and orbit memorization}
\label{sec:finite-sample-orbit}

Lemma~\ref{lemma:orbit-hsvm-inv} identifies an invariant population target, but does not by itself control the solution obtained from a finite sample. To make this distinction precise, fix a finite, strictly memorizable orbit $\cO$, let $M=|\cO|$, and write
\[
    \vomega^*=\operatorname{HSVM}(\cO),
    \qquad
    R=\max_{\vx\in\cO,\,j\in[n]}\|\vu_j(\vx)\|.
\]
For an i.i.d. sample $\cS_N=(\vx_i)_{i\in[N]}$ from $U(\cO)$, let $\widehat{\vomega}_N=\operatorname{HSVM}(\cS_N)$. Since the population solution has margin at least one on every population constraint,
\[
    \langle \vu_j(\vx),\widehat{\vomega}_N\rangle
    \geq 1-R\|\widehat{\vomega}_N-\vomega^*\|
    \qquad (\vx\in\cO,\ j\in[n]).
\]
In particular, $\|\widehat{\vomega}_N-\vomega^*\| \leq 1/(2R)$ guarantees a margin of at least $1/2$ on every orbit element. Note proximity to the invariant space alone is not sufficient to ensure all constraints are satisfied. The main obstacle to bounding $\|\widehat{\vomega}_N-\vomega^*\|$ is that $\widehat{\vomega}_N$ is not an empirical average. Indeed, it depends nonlinearly on the sample through the active constraints of the hard-margin problem, and these constraints can change when the sample changes. Consequently, a concentration inequality cannot be applied directly to the sample optimizer. For orbit data, however, the population KKT conditions provide a convenient route around this difficulty. Lemma~\ref{lemma:invariant-dual-certificate} shows that the population problem has an optimal dual certificate $(\bar\alpha_{\vx j})$ which respects the orbit symmetry: every $\vx\in\cO$ has the same collection of $n$ coefficients, up to the coordinate relabelling induced by the group action. The population solution can therefore be decomposed into pointwise contributions by defining
\[
    \vg(\vx)
    =M\sum_{j=1}^n\bar\alpha_{\vx j}\vu_j(\vx).
\]
This is a contribution to the population KKT representation, rather than a constraint-satisfaction score or a dual solution fitted to the sample. Its scaling is chosen so that
\[
    \E_{\vx\sim U(\cO)}[\vg(\vx)]
    =\frac1M\sum_{\vx\in\cO}\vg(\vx)
    =\vomega^*.
\]

Now form the empirical average $\vg_N=N^{-1}\sum_{i=1}^N\vg(\vx_i)$ using these fixed population coefficients. Although $\vg_N$ need not equal the sample HSVM solution, population optimality, sample feasibility, and complementary slackness give the deterministic upper bound
\[
    \|\widehat{\vomega}_N-\vomega^*\|
    \leq 2\|\vg_N-\vomega^*\|.
\]
Lemma~\ref{lemma:empirical-certificate-comparison} proves this inequality, stated there as~\eqref{eq:empirical-certificate-comparison}, in detail. This is the key step: it substitutes the nonlinear random sample solution by an empirical average of independent, mean-zero random vectors $\vg(\vx_i)-\vomega^*$. Standard vector concentration can then be applied. A generic bound on $\|\vg(\vx)-\vomega^*\|$ gives the theorem below; for cliques, the additional structure of the population certificate yields the sharper bound in Lemma~\ref{lemma:clique-certificate-deviation} and hence improved polynomial rates. In both cases the resulting sample complexity is independent of the orbit cardinality, which may itself be combinatorial or exponential.

\begin{restatable}{theorem}{ExactHSVMConcentration}\label{thm:exact-hsvm-concentration}
    Let $\cO=\textnormal{Orb}(\vx_0,\Gamma_n)$ be strictly memorizable and let $\cS_N=(\vx_i)_{i\in[N]}$, where the $\vx_i\sim U(\cO)$ are mutually i.i.d. If $\widehat{\vomega}_N=\textnormal{HSVM}(\cS_N)$, then for every $t>0$,
    \[
        \P\!\left(\|\widehat{\vomega}_N-\vomega^*\|>t\right)
        \leq 2\exp\!\left(-\frac{Nt^2}{32\|\vomega^*\|^4R^2}\right).
    \]
\end{restatable}

Theorem~\ref{thm:exact-hsvm-concentration} has the following implication.

\begin{restatable}{corollary}{UniformOrbitMemorization}\label{cor:uniform-orbit-memorization}
    Under the assumptions of Theorem~\ref{thm:exact-hsvm-concentration}, let $\widehat{\vtheta}_N=\mV\widehat{\vomega}_N$. For any $\delta\in(0,1)$, if
    \[
        N\gtrsim \|\vomega^*\|^4R^4\log(2/\delta),
    \]
    then, with probability at least $1-\delta$, every $\vx\in\cO$ is memorized by $\widehat{\vtheta}_N$ with energy gap at least $1/2$.
\end{restatable}

Corollary~\ref{cor:uniform-orbit-memorization} is qualitatively different from Theorem~\ref{lem:gen-svm-bounds}. Theorem~\ref{lem:gen-svm-bounds} applies to an arbitrary distribution on an arbitrary sparse, strictly memorizable set, but controls memorization only of a new independent draw. Corollary~\ref{cor:uniform-orbit-memorization} assumes uniform sampling from an orbit and proves simultaneous memorization of every orbit element with high probability. The orbit assumption is essential to the cardinality-independent rate, though not to every step of the certificate argument. The deterministic comparison in~\eqref{eq:empirical-certificate-comparison} relies only on a population KKT decomposition. Transitivity of the orbit is what allows Lemma~\ref{lemma:invariant-dual-certificate} to distribute the total dual mass equally across population points, thereby controlling each $\vg(\vx)$ without a factor depending on $|\cO|$. Without this symmetry, a small number of population points could carry most of the dual mass and the same proof would not give an orbit-cardinality-independent bound.

This conclusion is not specific to graph-isomorphism orbits: it applies to every strictly memorizable orbit under a subgroup of coordinate permutations. Together with the fixed-Hamming-weight construction of Lemma~\ref{lemma:express-inv-param}, it gives polynomial sample complexity in the ambient problem size even when $|\cO|$ is combinatorial or exponential. In particular, set $k=v/2$ in the clique problem considered by Conjecture~1 of \citet{hillar2021hidden}. Since $|\cC_{v,v/2}|=\binom{v}{v/2}$ grows exponentially whereas our sufficient sample size is polynomial in $v$, for fixed success probability the sufficient training fraction satisfies
\[
    \frac{N}{|\cC_{v,v/2}|}
    =\frac{\operatorname{poly}(v)}{\binom{v}{v/2}}
    =e^{-\Omega(v)}.
\]
Thus Corollary~\ref{cor:uniform-orbit-memorization} verifies the exponential upper bound predicted by that conjecture for the exact HSVM, and hence for the asymptotic max-margin direction selected by MEF.

For a specific orbit family, tighter polynomial rates can arise through two separate mechanisms. Throughout the clique-specific bounds below, $O_c(\cdot)$ and $\Theta_c(\cdot)$ allow hidden constants to depend on the fixed clique density $c$, but not on $v$, $N$, $\epsilon$, or $\delta$. First, a family-specific feasible construction can give a smaller upper bound on the norm of the population solution. For dense cliques, Lemma~\ref{lem:gii-strict-mem-ex} gives $\|\vomega^*\|^2=O_c(v)$, while $R^2=O_c(v^2)$. Substitution into Corollary~\ref{cor:uniform-orbit-memorization} gives a sufficient sample size of order $v^6$. Second, one can improve the generic deviation bound on the population contributions themselves. Lemma~\ref{lemma:clique-certificate-deviation} shows that clique symmetry spreads the dual coefficients across three large constraint classes and proves
\[
    \max_{\vx\in\cC_{v,k}}\|\vg(\vx)-\vomega^*\|=O_c(v),
\]
whereas the generic estimate in Lemma~\ref{lemma:empirical-certificate-comparison} gives only $O_c(v^2)$. Combining this sharper deviation estimate with~\eqref{eq:empirical-certificate-comparison} improves the sufficient orbit-memorization rate from $v^6$ to $v^4$. Once Lemma~\ref{lemma:clique-certificate-deviation} is established, the remaining conclusions follow directly from the empirical-certificate comparison and the uniform margin argument; we therefore state them as a corollary.

\begin{restatable}{corollary}{CliqueUniformMemorization}\label{cor:clique-uniform-memorization}
    Assume $k=cv\geq5$ for a constant $c\in(0,1)$, and let $\cS_N=(\vx_i)_{i\in[N]}$ with mutually i.i.d. $\vx_i\sim U(\cC_{v,k})$. Define
    \[
        \widehat{\vomega}_N=\operatorname{HSVM}(\cS_N),
        \quad
        \vomega^*=\operatorname{HSVM}(\cC_{v,k}),
        \quad
        \widehat{\vtheta}_N=\mV\widehat{\vomega}_N,
        \quad
        \vtheta^*=\mV\vomega^*.
    \]
    Then $\vtheta^*\in\Psi(\Phi_n)$. Moreover, there are constants $C_1,C_2>0$, depending only on $c$, such that, for any $\epsilon,\delta\in(0,1)$,
    \begin{enumerate}
        \item if $N\geq C_1v^2\epsilon^{-2}\log(2/\delta)$, then with probability at least $1-\delta$, $\|\widehat{\vtheta}_N-\vtheta^*\|\leq\epsilon$;
        \item if $N\geq C_2v^4\log(2/\delta)$, then with probability at least $1-\delta$, $\widehat{\vtheta}_N$ memorizes every $k$-clique with energy gap at least $1/2$.
    \end{enumerate}
\end{restatable}

There remains room to improve both clique rates. Lemma~\ref{lemma:clique-certificate-deviation} uses only the three constraint-class sizes, nonnegativity of the dual coefficients, and their total mass; it does not use the actual mass assigned to each class or determine which classes are active. The concentration step then uses the worst-case radius $\max_{\vx}\|\vg(\vx)-\vomega^*\|$, rather than the covariance or typical size of these vectors. Finally, uniform memorization is obtained from the generic bound
\(
    |\langle\vu_j(\vx),\widehat{\vomega}_N-\vomega^*\rangle|
    \leq R\|\widehat{\vomega}_N-\vomega^*\|
\),
which ignores the alignment between the parameter error and the individual clique constraints. An exact analysis of the reduced clique dual, a variance-sensitive concentration inequality, or direct control of the constraint margins could therefore yield smaller powers of $v$.

\section{Conclusions, Limitations and Future Work}
\label{sec:conclusion}

In summary, this work shows that classical Hopfield networks can exploit symmetry in graph-structured data. By reformulating strict memorization as a linear feasibility problem, we identify norm efficiency as a mechanism linking MEF training, max-margin solutions, and sample-efficient orbit generalization. For a uniform distribution on a finite orbit, the exact sample HSVM concentrates around its invariant population solution; with enough samples it therefore memorizes the entire orbit with high probability. For dense cliques, a symmetric certificate gives polynomial sample bounds both for proximity to the invariant subspace and for uniform orbit memorization. These results show that even in minimal associative-memory models, implicit bias can recover invariance without imposing it architecturally.

The clean finite-sample result here concerns the exact HSVM, which is the limiting direction selected by MEF. A corresponding finite-iteration rate for the MEF iterates themselves remains open: the implicit-bias result of \citet{soudry2018the} is asymptotic and does not by itself yield the quantitative directional control needed to transfer our HSVM orbit bound at a prescribed iteration. Other natural directions are to explain why some isomorphism classes are easier than others, quantify spurious fixed points and basin robustness, treat other subgroups and unions of orbits, handle noisy or non-uniform group data, extend the analysis to hypergraphs, and study continuous and modern HNs. Appendix~\ref{app:add-exp} records preliminary experiments relevant to these questions, including invariant-subspace convergence (Fig.~\ref{fig:weight_heatmaps2}), the Hidden Clique Problem (Fig.~\ref{fig:HCP}), comparison with DAMs (Fig.~\ref{fig:gen_dams}), other graph families (Figs.~\ref{fig:other_families}, \ref{fig:other_families2}), double descent (Fig.~\ref{fig:ddescent}), and isomorphic graph checking (Algorithm~\ref{alg:hngic}).

\textbf{Reproducibility Statement:} Code able to reproduce our experimental results can be found in the following repositories: \url{https://github.com/hopnetorbit/HopfieldNetworksIsomorphism} and \url{https://github.com/qualiaphile/mefgraphs}.

\textbf{Ethics Statement:} This work uses only synthetic, non-sensitive data, involves no human or animal subjects, and carries minimal dual-use or environmental risks (modest compute).

\ifdefined\PaperTMLR
  \textbf{Broader Impact Statement:} This theoretical work poses minimal immediate societal risk. Its analysis of symmetry, generalization, and representation may inform more reliable and interpretable learning systems.
\else
  \textbf{Broader Impact Statement:} This work contributes to the mathematical and conceptual foundations of machine learning and intelligent systems. Although the results are primarily theoretical, a better understanding of the underlying principles governing learning, generalization, and representation can inform the long-term development of more reliable, interpretable, and robust algorithms.
\fi

\ifdefined\PaperTMLR
\else
  \textbf{Statement on the use of LLMs.} Large language models, including OpenAI's GPT-5.6 Sol, were used extensively as research assistants for literature search, high-level ideation and planning, organizing related work, and drafting and refining exposition. All LLM-generated material was critically evaluated and revised by the authors. In particular, the authors checked every mathematical claim and proof in full, verified the cited sources, and reworked the arguments and writing where necessary. The authors take full responsibility for the correctness and presentation of the paper.
\fi

\PaperAcknowledgments
This work was initiated during the Mathematics of Intelligences long program at the Institute for Pure and Applied Mathematics (IPAM), UCLA, held from September to December 2024. The authors thank IPAM for its support and hospitality, and the program participants for creating a stimulating research environment.

\clearpage
\bibliography{refs}
\bibliographystyle{\PaperBibliographyStyle}

\newpage

\PaperAppendixStart

\section{Background}

\subsection{Energy gap for binary vectors a Hamming distance one apart}\label{app:props-energy-function}
As discussed in Section~\ref{sec:prelims}, strict memorization of a point is equivalent to a positive energy gap between it and each of its Hamming-distance-one neighbors. Recall that $\vx^{(j)}\in\{0,1\}^n$ denotes the vector that differs from $\vx\in\{0,1\}^n$ only at coordinate $j$, and $\vz(\vx)=[\vx,1]\in\R^{n+1}$.

\begin{restatable}{lemma}{LemOneBitDiff}\label{lem:one-bit-diff}
$E(\vx^{(j)}; \vtheta ) - E(\vx; \vtheta ) = y_j(\vx) \langle \vz(\vx), \vtheta_j \rangle$.
\end{restatable}
\begin{proof}
By definition, $x_l^{(j)}=x_l$ for $l\neq j$ and $x_j^{(j)}-x_j=1-2x_j$. Thus $x_l^{(j)}x_r^{(j)}-x_lx_r=0$ whenever neither index equals $j$. Furthermore, recall $\mW$ is symmetric and $W_{jj}=0$ for all $j \in [n]$. As a result,
\begin{align*}
    E(\vx^{(j)}; \vtheta ) - E(\vx; \vtheta ) &= \frac{1}{2} \sum_{l,r \in [n]} W_{rl}(x_r^{(j)}x_l^{(j)} - x_rx_l) + \sum_{l=1}^n b_l (x_l^{(j)} - x_l)\\
    &= \frac{1}{2} \sum_{r \in [n], r\neq j} W_{rj}(x_r^{(j)}x_j^{(j)} - x_rx_j) + \frac{1}{2} \sum_{l \in [n], l \neq j} W_{jl}(x_j^{(j)}x_l^{(j)} - x_jx_l) + b_j(1 - 2x_j)\\
    &= \sum_{l \in [n], l \neq j} W_{jl}(x_j^{(j)}x_l^{(j)} - x_jx_l) + b_j(1 - 2x_j)\\
    &= \sum_{l \in [n], l \neq j} W_{jl}(x_j^{(j)} - x_j)x_l + b_j(1 - 2x_j)\\
    &=(1 - 2x_j) \left(\sum_{l \in [n]} W_{jl}x_l + b_j\right)\\
    &=y_j(\vx) \langle \vz(\vx), \vtheta_j \rangle.
\end{align*}
as claimed.
\end{proof}

\subsection{Learning algorithms for Hopfield Networks}\label{app:learning-rules}
We briefly describe several classical learning rules \citep{Hertz91} that can be applied to find parameters in Hopfield networks. These methods typically trade off between biological plausibility and performance. We remark that this list is far from exhaustive; see \citet{Tolmachev20} for a recent summary.
\begin{itemize}
    \item \textbf{Outer-Product Rule} \citep{hebb1949organization, Nakano72, amari1972learning, hopfield1982neural}. In the attractor neural network case \citep{hopfield1982neural}, this Hebbian rule constructs weights as the normalized sum of training pattern outer products. This rule is simple, biologically motivated, and local in nature, but it is often observed to suffer from spurious attractors, shallow basins of attraction, and overall limited capacity.
    \item \textbf{Perceptron Rule} \citep{rosenblatt1958perceptron}.
    The difference between the desired response -- that the network dynamics should fix a training sample -- and the actual linear-threshold response of a neuron gives a learning signal to update parameters.
    \item \textbf{Delta} \citep{hoff1960adaptive, rescorla1972theory}.
    The delta rule, also called the Mean Squared Error (MSE) or Least Mean Square (LMS) rule, considers a relaxation and follows a gradient to minimize the squared error between the linear output activations of neurons and the training pattern to memorize.
    \item \textbf{Projection Rule} \citep{personnaz1985storage, personnaz1986collective}.
    The weight matrix is obtained by projecting onto the span of the training data and then zeroing the diagonal entries.
    \item \textbf{Storkey Rule }\citep{storkey1997increasing, storkey1999basins}.
    A modification of the Hebbian update that reduces interference between patterns by accounting for previously stored ones.
    This rule achieves higher storage capacity than Hebbian learning and reduces spurious minima.
\end{itemize}

\subsection{Encoding simple, undirected graphs as binary vectors} \label{app:sec:encoding-graph-data}
First, we recap some notation. Let $\cG_v$ denote the set of all simple, undirected graphs on $v\in\N$ vertices. Two graphs $G=(V,E)$ and $G'=(V',E')$ are \textit{isomorphic}, denoted $G\cong G'$, if there exists a bijection $\phi:V\rightarrow V'$ such that $(\phi(\nu_1),\phi(\nu_2))\in E'$ if and only if $(\nu_1,\nu_2)\in E$. We call such a map $\phi$ an isomorphism between $G$ and $G'$. When $V=V'=[v]$, the map $\phi$ is a permutation, which is the setting considered below. The \textit{isomorphism class} of $G\in\cG_v$ is $\cI(G)\coloneq\{G'\in\cG_v:G'\cong G\}$. An automorphism is an isomorphism from a graph to itself. Recall that $\Phi_n\subset\cP_n$ denotes the set of edge-permutation matrices induced by vertex permutations; see Definition~\ref{def:vertex-induced-edge-perm}.

\begin{lemma}
    $\Phi_n$ is a subgroup of $\cP_n$.
\end{lemma}

\begin{proof}
    Clearly this is equivalent to showing that $\Pi_n^{\Phi}$ is a subgroup of $\Pi_n$. It is easy to check that $I \in \Pi_n^{\Phi}$, $\pi_{\phi} \in \Pi_n^{\Phi}$ implies $\pi_{\phi}^{-1} \in \Pi_n^{\Phi}$ and $\pi_{\phi}, \pi_{\phi'} \in \Pi_n^{\Phi}$ implies $\pi_{\phi} \circ \pi_{\phi'} \in\Pi_n^{\Phi} $, therefore $\Pi_n^{\Phi}$ is a subgroup of $\Pi_n$.
\end{proof}

The following lemmas establish a straightforward equivalence between isomorphism classes of graphs and certain orbits of binary vectors. First, Lemma~\ref{lem:graph-iso-vertex-induced} shows that if two graphs $G,G' \in \cG_v$ are isomorphic then there is a vertex induced edge permutation which maps between their edge representations.

\begin{lemma} \label{lem:graph-iso-vertex-induced}
    Suppose $G = ([v], E), G' = ([v], E') \in \cG_v$ and $\vx = \cE_{rep}(G)$, $\vx' = \cE_{rep}(G')$. Then $G \cong G'$ iff there exists a $\mP \in \Phi_n$ such that $\mP \vx = \vx'$.
\end{lemma}

\begin{proof}
    Assume $G\cong G'$. Then there exists a permutation $\phi:[v]\rightarrow[v]$ such that $(\nu_1,\nu_2)\in E$ if and only if $(\phi(\nu_1),\phi(\nu_2))\in E'$. Let $\pi_{\phi}:[n]\rightarrow[n]$ be the edge permutation induced by $\phi$, and let $\mP\in\Phi_n$ be its permutation matrix. By construction, $x_j=x'_{\pi_{\phi}(j)}$ for all $j\in[n]$, equivalently $\mP\vx=\vx'$.

    Conversely, suppose that $\mP\vx=\vx'$ for some $\mP\in\Phi_n$. Then $\mP$ is induced by a vertex permutation $\phi:[v]\rightarrow[v]$. If $j=\textnormal{Ind}((\nu_1,\nu_2))$, then $\pi_{\phi}(j)=\textnormal{Ind}((\phi(\nu_1),\phi(\nu_2)))$, and $x_j=x'_{\pi_{\phi}(j)}$ implies
    \[
        (\nu_1,\nu_2)\in E
        \quad\Longleftrightarrow\quad
        (\phi(\nu_1),\phi(\nu_2))\in E'.
    \]
    Therefore $\phi$ is an isomorphism and $G\cong G'$.
\end{proof}

Building on Lemma \ref{lem:graph-iso-vertex-induced}, the following lemma characterizes the isomorphism class of a graph in terms of the orbit of its edge representation under vertex induced edge permutations.

\begin{lemma}
    For any $G \in \cG_v$ we have
    $\cE_{rep}(\cI(G)) = \textnormal{Orb}(\cE_{rep}(G), \Phi_n)$.
\end{lemma}

\begin{proof}
    Let $\vx = \cE_{rep}(G)$, then
    \[
    \textnormal{Orb}(\cE_{rep}(G), \Phi_n) = \{ \mP \vx : \mP \in \Phi_n\}.
    \]
    Suppose $\cE_{rep}(\cI(G)) \not \subset \textnormal{Orb}(\cE_{rep}(G), \Phi_n)$. Then there exists a $G' \in \cI(G)$ such that $\vx' \coloneq \cE_{rep}(G') \not \in \textnormal{Orb}(\cE_{rep}(G), \Phi_n)$. Therefore there does not exist a $\mP \in \Phi_n$ such that $\mP \vx = \vx'$. However, $G \cong G'$ which implies a contradiction by Lemma \ref{lem:graph-iso-vertex-induced}, therefore $\cE_{rep}(\cI(G)) \subset \textnormal{Orb}(\cE_{rep}(G), \Phi_n)$. Now suppose $\textnormal{Orb}(\cE_{rep}(G), \Phi_n) \not \subset \cE_{rep}(\cI(G))$, then there exists a $\vx' \in  \textnormal{Orb}(\cE_{rep}(G), \Phi_n)$ such that $G'= \cE_{rep}^{-1}(\vx') \not \in \cI(G)$. However, as $\vx' \in  \textnormal{Orb}(\cE_{rep}(G), \Phi_n)$ then there exists a $\mP \in \Phi_n$ such that $\mP \vx = \vx'$, but using Lemma \ref{lem:graph-iso-vertex-induced} this implies $G \cong G'$ which is a contradiction. Therefore $\textnormal{Orb}(\cE_{rep}(G), \Phi_n) \subset \cE_{rep}(\cI(G))$. Combining these two observations we conclude that $\cE_{rep}(\cI(G)) = \textnormal{Orb}(\cE_{rep}(G), \Phi_n)$.
\end{proof}

\begin{lemma}
    $\cQ_n$ is a subgroup of $\cP_n$.
\end{lemma}

\begin{proof}
    Trivially it suffices to show that $\Pi_n^{\cQ}$ is a subgroup of $\Pi_n$. It is easy to check that $I \in \Pi_n^{\cQ}$, $\pi \in \Pi_n^{\cQ}$ implies $\pi^{-1} \in \Pi_n^{\cQ}$ and $\pi, \pi' \in \Pi_n^{\cQ}$ implies $\pi \circ \pi' \in\Pi_n^{\cQ}$. Therefore $\Pi_n^{\cQ}$ is a subgroup of $\Pi_n$.
\end{proof}

The following lemma states that the vertex induced edge permutations form a subgroup of the edge adjacency preserving subgroup of permutations. As a result, the edge representations of the isomorphism class of a graph are a subset of the orbit of the edge representation of the graph in question under edge adjacency preserving permutations.

\begin{lemma}
   $\Phi_n$ is a subgroup of $\cQ_n$ and $\cE_{rep}(\cI(G)) \subset \textnormal{Orb}(\cE_{rep}(G), \cQ_n)$.
\end{lemma}

\begin{proof}
    It suffices to show that every vertex-induced edge permutation preserves edge adjacency. Let $i,j\in[n]$. If $i\sim j$, their corresponding vertex pairs share exactly one vertex. Because a vertex permutation $\phi$ is bijective, the image pairs also share exactly one vertex, so $\pi_{\phi}(i)\sim\pi_{\phi}(j)$. If $i\nsim j$, then either $i=j$, in which case their images are equal and hence nonadjacent by definition, or the two vertex pairs are disjoint, in which case their images remain disjoint. Thus
    \[
        i\sim j \quad\Longleftrightarrow\quad \pi_{\phi}(i)\sim\pi_{\phi}(j).
    \]
    Hence $\Pi_n^{\Phi}\subseteq\Pi_n^{\cQ}$ and therefore $\Phi_n\leq\cQ_n$. The orbit inclusion follows immediately:
    \[
        \textnormal{Orb}(\cE_{rep}(G),\Phi_n)
        \subseteq \textnormal{Orb}(\cE_{rep}(G),\cQ_n).
    \]
\end{proof}

\subsection{Bounded representations}
Recall that $\vtheta=\mV\vomega$, $\vtheta_j=\mV_j\vomega$, and $\vu_j(\vx)=y_j(\vx)\mV_j^T\vz(\vx)$. The following lemma gives the exact radius of these constraint vectors.

\begin{lemma}\label{lemma:bounded-representations}
    For any $\vx \in \{0,1 \}^n$,
    \[
    \|\vu_j(\vx) \|^2 = \frac{1}{2}\left( \| \vx \|_0 - x_j \right) + 1
    \]
    for all $j \in [n]$.
\end{lemma}

\begin{proof}
    Recall that $p=n(n-1)/2$ and that $\va\in\R^p$ lists the strictly upper-triangular entries of $\mW$ in a fixed order. By symmetry, these entries determine all off-diagonal weights. Let $\delta_n(j)\in\{0,1\}^n$ be the one-hot vector with support $\{j\}$. Define $\mB_j\in\{0,1\}^{n\times p}$ as follows: for each $l\neq j$, row $l$ selects the entry of $\va$ corresponding to the unordered pair $\{j,l\}$, while row $j$ is zero. Thus $\vw_j=\mB_j\va$, and we can write
    \[
    \vtheta_j =
    \begin{bmatrix}
        \vw_j\\
        b_j
    \end{bmatrix}
    = \begin{bmatrix}
        \tfrac{1}{\sqrt{2}} \mB_j & \textbf{0}_{n\times n} \\
        \textbf{0}_{1\times p} & \delta_n(j)^T
    \end{bmatrix}
    \begin{bmatrix}
    \sqrt{2} \va\\
    \vb
    \end{bmatrix}
    = \mV_j \vomega.
    \]
    By definition
    \begin{align*}
        \vu_j(\vx) = y_j(\vx)\mV_j^T \vz(\vx) = y_j(\vx)\begin{bmatrix}
            \tfrac{1}{\sqrt{2}} \mB_j^T & \textbf{0}_{p\times 1} \\
        \textbf{0}_{n\times n} & \delta_n(j)
        \end{bmatrix}
        \begin{bmatrix}
            \vx \\
            1
        \end{bmatrix}
        = y_j(\vx)\begin{bmatrix}
            \frac{1}{\sqrt{2}} \mB_j^T\vx \\
            \delta_n(j)
        \end{bmatrix},
    \end{align*}
    therefore
    \begin{align*}
        \| \vu_j(\vx) \|^2 = \frac{1}{2} \vx^T \mB_j \mB_j^T \vx + 1.
    \end{align*}
    Each nonzero row of $\mB_j$ selects a distinct unordered pair $\{j,l\}$. Hence these rows are distinct standard basis vectors, while row $j$ is zero. As a result,
    \[
    \mB_j \mB_j^T = \textbf{I}_n - \ve_j\ve_j^T.
    \]
    This implies
    \[
    \vx^T \mB_j \mB_j^T \vx = \vx^T \left(\textbf{I}_n - \ve_j\ve_j^T \right) \vx = \| \vx \|_0 - x_j
    \]
    for all $j \in [n]$. Since $y_j(\vx)^2=1$, this proves the claim.
\end{proof}

\subsection{Hoeffding's inequality in Hilbert space}
The exact-HSVM result uses the following concentration bound for sums of independent, mean-zero, bounded random vectors. It is a specialization of more general results for martingales in 2-smooth Banach spaces.

\begin{lemma} \label{lemma:vectorHoeffding}
    \textbf{[Specialization of~\citep[Thm.~3.5]{pinelis94}.]} For $i\in [N]$ and $K \in \R_{>0}$, let $\vx_i \in \R^q$ be independent, mean-zero random vectors satisfying $\| \vx_i\| \leq K$ almost surely. Let $S_N = \sum_{i=1}^N \vx_i$. Then, for $t \in \R_{\geq 0}$,
    \[
    \P(\| S_N \| \geq t) \leq 2\exp\left( - \frac{t^2}{2N K^2}\right).
    \]
\end{lemma}

\section{Proofs of results}

\subsection{Proof of Theorem~\ref{lem:gen-svm-bounds}}\label{app:lem:gen-svm-bounds}

\GenSvmBounds*

\begin{proof}
    For any $t \in \R$ define the margin loss $\phi: \R \rightarrow \R$ as
    \[
    \phi(t) = \begin{cases}
        0, & 1 \leq t,\\
        1 - t,& 0 \leq t \leq 1,\\
        1, & t \leq 0,
    \end{cases}
    \]
    and note trivially for all $t \in \R$ that $\1(t \leq 0) \leq \phi(t)$. We note on occasion we overload this notation and apply $\phi$ to vectors by applying it elementwise. Observe for any $\vx \in \R^n$ and $\vomega \in \R^q$ with $\vtheta = \mV \vomega$, that
    \begin{align*}
    \1(H(\vx; \vtheta) \neq \vx)
    &\leq \1 \left(\exists j \in [n]: \vu_j(\vx)^T \vomega \leq 0 \right)\\
    &= \1 \left( \min_{j \in [n]} \vu_j(\vx)^T \vomega \leq 0 \right)\\
    &\leq \phi\!\left( \min_{j \in [n]} \vu_j(\vx)^T \vomega\right)
     = \max_{j \in [n]} \phi\!\left(\vu_j(\vx)^T \vomega\right).
    \end{align*}
    For any $\vz \in \R^n$, let $\ell(\vz) = \max_{j \in [n]} \phi(z_j)$. Using the fact that $\phi$ is 1-Lipschitz, for any $\vz, \vz' \in \R^n$ we have
    \begin{align*}
        |\ell(\vz) - \ell(\vz')|
        \leq \max_{j \in [n]} |\phi(z_j) - \phi(z_j')|
         = \| \phi(\vz) - \phi(\vz')\|_{\infty}
         \leq \|\vz - \vz' \|_{\infty}
        \leq \| \vz - \vz'  \|_2.
    \end{align*}
    Therefore $\ell$ is 1-Lipschitz with respect to the Euclidean norm. Let $\mU(\vx) \in \R^{n \times q}$ denote the matrix whose $j$th row is $\vu_j(\vx)^T \in \R^{1 \times q}$ for all $j \in [n]$. Furthermore, for some $\Lambda \in \R_{>0}$, define
    \[
    \cH_{\Lambda} = \{ \vx \mapsto \mU(\vx) \vomega \; : \vomega \in \R^{q}, \; \|\vomega \| \leq \Lambda \}
    \]
    and let
    \[
    \cG_{\Lambda} = \{ \vx \mapsto (\ell \circ h )(\vx): h \in \cH_{\Lambda}\}.
    \]
    By construction, $g \in \cG_{\Lambda}$ implies $g: \R^n \rightarrow [0,1]$. We now compute the empirical Rademacher complexity of $\cG_{\Lambda}$ on a sample $\cS_N = (\vx_i)_{i \in [N]}$. Let $\vsigma \in \{\pm 1\}^N$ and $\vepsilon \in \{\pm 1\}^{N \times n}$ be a random vector and matrix, respectively, whose entries are mutually i.i.d. and uniform on $\{\pm 1\}$. Since $\ell$ is 1-Lipschitz in the $\ell_2$ norm, a vector contraction inequality~\citep[Corollary 1]{vector-contrac} gives
    \begin{align*}
    \tilde{\mathfrak{R}}_{\cS}(\cG_{\Lambda})
        &= \E_{\vsigma} \left[\sup_{h \in \cH_{\Lambda}} \frac{1}{N}\sum_{i=1}^N\sigma_i (\ell \circ h)(\vx_i) \right] \\
        &\leq \sqrt{2} \E_{\vepsilon} \left[\sup_{\| \vomega \| \leq \Lambda} \frac{1}{N}\sum_{i=1}^N \sum_{j=1}^n \epsilon_{ij} \vu_j(\vx_i)^T \vomega \right]\\
        & \leq \sqrt{2} \E_{\vepsilon} \left[\sup_{\| \vomega \| \leq \Lambda} \langle \vomega,  \frac{1}{N}\sum_{i=1}^N \sum_{j=1}^n \epsilon_{ij} \vu_j(\vx_i) \rangle \right]\\
        & \leq \frac{\sqrt{2} \Lambda}{N} \E_{\vepsilon} \left[ \left\|  \sum_{i=1}^N \sum_{j=1}^n \epsilon_{ij} \vu_j(\vx_i) \right\| \right].
    \end{align*}
    Let $Z = \sum_{i=1}^N \sum_{j=1}^n \epsilon_{ij} \vu_j(\vx_i)$, as $t \mapsto \sqrt{t}$ is concave then $\E_{Z} [\sqrt{\| Z\|^2} ] \leq \sqrt{ \E_{Z} \left[ \|Z\|^2 \right] }$ by Jensen's inequality. As a result
    \begin{align*}
    \E_{\vepsilon} \left[ \left\|  \sum_{i=1}^N \sum_{j=1}^n \epsilon_{ij} \vu_j(\vx_i) \right\| \right]
    &\leq \sqrt{ \E_{\vepsilon} \left[ \left\langle \sum_{i=1}^N \sum_{j=1}^n \epsilon_{ij} \vu_j(\vx_i),\sum_{l=1}^N \sum_{k=1}^n \epsilon_{lk} \vu_k(\vx_l)\right\rangle \right] } \\
    &= \sqrt{ \sum_{i=1}^N \sum_{j=1}^n\sum_{l=1}^N \sum_{k=1}^n \E_{\vepsilon} [\epsilon_{ij} \epsilon_{lk}] \vu_j(\vx_i)^T \vu_k(\vx_l) }.
    \end{align*}
    The Rademacher random variables are mutually i.i.d., therefore
    \[
    \E_{\vepsilon} [\epsilon_{ij} \epsilon_{lk}] =
    \begin{cases}
        & 1, \;\; (i = l) \wedge (j=k),\\
        & 0, \;\; \text{otherwise.}
    \end{cases}
    \]
    Recall also from Lemma~\ref{lemma:bounded-representations} that for any $i \in [N]$ and for all $j \in [n]$
    \[
    \|\vu_j(\vx_i) \|^2 = \frac{1}{2}\left( \| \vx_i \|_0 - x_{ij} \right) + 1.
    \]
    Under the assumption $\| \vx_i\|_0 \leq m$ for all $i \in [N]$, then
    \begin{align*}              \tilde{\mathfrak{R}}_{\cS}(\cG_{\Lambda})
    & \leq\frac{\sqrt{2} \Lambda}{N} \E_{\vepsilon} \left[ \left\|  \sum_{i=1}^N \sum_{j=1}^n \epsilon_{ij} \vu_j(\vx_i) \right\| \right]\\
    & \leq \frac{\sqrt{2}\Lambda}{N} \sqrt{ \sum_{i=1}^N \sum_{j=1}^n \| \vu_j(\vx_i)\|^2}\\
    & \leq \frac{\Lambda}{N} \sqrt{\sum_{i=1}^N \sum_{j=1}^n \left( \| \vx_i \|_0 - x_{ij}  + 2\right) }\\
    & \leq  \Lambda \sqrt{\frac{n(m + 2 )}{N}}.
    \end{align*}
    Let $\delta \in \R_{>0}$. Applying a Rademacher complexity bound, e.g.,~\citep[Thm 3.3]{FoML},  then with probability at least $1-\delta$ over the random sample $\cS_N$
    \begin{align*}
        \E[g(\vx)]
        &\leq \frac{1}{N} \sum_{i=1}^N g(\vx_i) + 2\tilde{\mathfrak{R}}_{\cS}(\cG_{\Lambda}) + 3 \sqrt{\frac{\log(2/ \delta)}{2N}} \\
        &\leq \frac{1}{N} \sum_{i=1}^N g(\vx_i) + 2 \Lambda \sqrt{\frac{n(m + 2 )}{N}} + 3 \sqrt{\frac{\log(2/ \delta)}{2N}}
    \end{align*}
    for all $g \in \cG_{\Lambda}$. In what follows let $\Lambda = \| \vomega^*\|$. Then, with probability at least $1-\delta$ over the random sample $\cS_N$, for any $\vomega \in \R^q$ such that $\| \vomega \| \leq \| \vomega^*\|$ we have
    \begin{equation} \label{eq:rademacher}
        \P(H(\vx; \mV\vomega) \neq \vx) \leq \frac{1}{N} \sum_{i=1}^N \phi(\min_{j \in [n]} \vu_j(\vx_i)^T \vomega) + 2 \sqrt{\frac{ \| \vomega^*\|^2 n(m + 2 )}{N}} + 3 \sqrt{\frac{\log(2/ \delta)}{2N}}
    \end{equation}
    First we consider $\hat{\vomega}$: for any sample $\cS_N$ trivially $\text{set}(\cS_N) \subseteq D$, therefore $\vomega^* \in \cF_{\vomega}(\cS_N)$. As a result, with probability one $\| \hat{\vomega} \| \leq \| \vomega^* \| $. Conditioning on this event, observe also that $\min_{j \in [n]} \vu_j(\vx_i)^T \hat{\vomega} \geq 1$ for $i \in [N]$.
    As a result, with probability at least $1 - \delta$ over $\cS_N$ we have
    \[
    \P(H(\vx; \mV\hat{\vomega}) \neq \vx)  \leq    \sqrt{\frac{4 \| \vomega^* \|^2n(m + 2 )}{N}} +  \sqrt{\frac{9\log(2/ \delta)}{2N}}.
    \]
    As $(a+b)^2 \leq 2(a^2 + b^2)$ for $a,b \in \R$ and assuming $m \geq 1$, then for any $\epsilon \in \R_{>0}$, if $N \gtrsim  \epsilon^{-2}n\|\vomega^* \|^2 m \log(1/\delta)$
    \[
    \P_{\vx}(H(\vx; \mV\hat{\vomega}) \neq \vx) \leq \epsilon
    \]
    with probability at least $1-\delta$ over the sample $\cS_N$.

    We now turn our attention to $\vomega(t)$: recall for any $a \in \R_{>0}$ as $E(\vx; a\vtheta) = a E(\vx; \vtheta)$ then
    \[
    a(E(\vx^{(j)};\vtheta) - E(\vx; \vtheta)) \geq  0 \iff E(\vx^{(j)};a\vtheta) - E(\vx; a\vtheta) \geq 0.
    \]
    Indeed, this implies that the set of memories in a Hopfield network is invariant under positive rescalings of the parameters. Consequently, for any distribution $\cD$ on $\{0,1\}^n$, $a \in \R_{>0}$, and $\vtheta \in \Theta$, if $\vx \sim \cD$ we have
    \[
    \P(H(\vx; \vtheta) \neq \vx) = \P(H(\vx; a \vtheta) \neq \vx).
    \]
    Therefore, if we define
    \[
    \bar{\vomega}(t) =  \tfrac{\| \hat{\vomega} \|}{\| \vomega(t)\| } \vomega(t)
    \]
    it follows that
    \[
    \P(H(\vx; \mV \vomega(t)) \neq \vx) = \P(H(\vx; \mV \bar{\vomega}(t) ) \neq \vx).
    \]
    From \citet[Theorem~5]{soudry2018the},
    \[
    \left \| \frac{\bar{\vomega}(t)}{\| \hat\vomega \| } - \frac{\hat\vomega}{\| \hat\vomega\| } \right\| = \left \| \frac{\vomega(t)}{\| \vomega(t) \| } - \frac{\hat\vomega}{\| \hat \vomega\| } \right\| = O \left(\frac{\log(\log(t))}{\log(t)} \right).
    \]
    Recalling $\| \hat \vomega\| \leq \| \vomega^*\|$ this implies
    \[
    \| \bar{\vomega}(t) - \hat \vomega \| = O \left(\frac{ \| \vomega^* \| \log(\log(t))}{\log(t)} \right).
    \]
    As a result, for all $i \in [N]$ we have
    \begin{align*}
        \min_{j \in [n]}\vu_j(\vx_i)^T \bar{\vomega}(t) &= \min_{j \in [n]} \left(\vu_j(\vx_i)^T \hat \vomega +  \vu_j(\vx_i)^T(\bar{\vomega}(t) - \hat \vomega)\right)\\
        &\geq 1 - \max_{j \in [n]}\| \vu_j(\vx_i) \| \|\bar{\vomega}(t) - \hat \vomega \| \\
        &\geq 1 - O \left(\frac{ \sqrt{m} \| \vomega^* \| \log(\log(t))}{\log(t)} \right),
        \end{align*}
    where the final inequality follows from Lemma~\ref{lemma:bounded-representations} and the assumption $m\geq 1$. By the definition of $\phi$ it follows that
    \[
    \phi(\min_{j \in [n]}\vu_j(\vx_i)^T \bar{\vomega}(t)) = O\left( \frac{ \sqrt{m} \| \vomega^* \| \log(\log(t))}{\log(t)} \right)
    \]
    for all $i \in [N]$. Using~\eqref{eq:rademacher}, this implies with probability at least $1 - \delta$ over $\cS_N$
    \[
    \P(H(\vx; \mV\vomega(t)) \neq \vx)  \leq O\left( \frac{ \sqrt{m} \| \vomega^* \| \log(\log(t))}{\log(t)}\right) +    \sqrt{\frac{4 \| \vomega^* \|^2n m}{N}} +  \sqrt{\frac{9\log(2/ \delta)}{N}}.
    \]
    As a result, if $N \gtrsim  \epsilon^{-2}n\|\vomega^* \|^2 m \log(1/\delta)$ then
    \[
    \P(H(\vx; \mV\vomega(t)) \neq \vx)  = \tilde{O}\left( \frac{ \sqrt{m} \| \vomega^* \|}{\log(t)}\right) +  \epsilon.
    \]
    with probability at least $1 - \delta$ over the sample $\cS_N$.
\end{proof}

\subsection{Properties of invariant parameters} \label{app:invariance}

\subsubsection{Energy bounds for invariant parameters across orbits}

The following result states, in the context of invariant parameters, that energy bound differences for a point in an orbit extend to the entire orbit.

\LemLB*

\begin{proof}
    As $\vtheta \in \Psi(\Gamma_n)$, recall the intertwining relation~\eqref{eq:intertwining}, for $\vx \in \{ 0,1\}^n$ and $\mQ \in \Gamma_n$ then
    \begin{align*}
        E(\mQ \vx; \vtheta)
         = E(\vx; \mQ \vtheta)
         = E(\vx; \vtheta).
    \end{align*}
    For any $\vx \in \textnormal{Orb}(\vx_0, \Gamma_n)$, there exists a $\mQ \in \Gamma_n$ such that $\vx_0 = \mQ\vx$. Let $\pi$ be its coordinate permutation, with $\mQ\ve_l=\ve_{\pi(l)}$. Then $(\mQ\vx)^{(j)}=\mQ\vx^{(\pi^{-1}(j))}$. By invariance,
    \begin{align*}
        1-\delta
        &\leq E(\vx_0^{(j)};\vtheta)-E(\vx_0;\vtheta)\\
        &=E(\mQ\vx^{(\pi^{-1}(j))};\vtheta)-E(\mQ\vx;\vtheta)\\
        &=E(\vx^{(\pi^{-1}(j))};\vtheta)-E(\vx;\vtheta).
    \end{align*}
    Since $\pi$ is a bijection, this implies for every $j\in[n]$ that
    \[
    E(\vx^{(j)}; \vtheta) - E(\vx; \vtheta) \geq 1-\delta
    \]
    as claimed.
\end{proof}

\subsubsection{Characterizing the invariant set for edge adjacency preserving permutations}

The following lemma identifies the parameters invariant under edge-adjacency-preserving permutations as a particular rank-three subspace.

\LemInvParamsCond*

\begin{proof}
    We first show that $F(\R^3)\subseteq\Psi(\cQ_n)$. Let $\vtheta=(\mW,\vb)=F(\vbeta)$ and let $\mQ\in\cQ_n$ correspond to the edge permutation $\pi$. Since $b_i=\beta_3$ for every $i\in[n]$, we have $b_i=b_{\pi(i)}$. Since $\pi$ is bijective, $i=j$ if and only if $\pi(i)=\pi(j)$, and $i\neq j$ if and only if $\pi(i)\neq\pi(j)$. Moreover, $\pi$ preserves edge adjacency by the definition of $\cQ_n$. It therefore also preserves the remaining case of distinct, non-adjacent edges. Consequently,
    \[
        W_{ij}=W_{\pi(i)\pi(j)}
        \qquad\text{for every }i,j\in[n].
    \]
    Thus $\mQ^T\mW\mQ=\mW$ and $\mQ^T\vb=\vb$, so $\vtheta\in\Psi(\cQ_n)$. Since $\Phi_n$ is a subgroup of $\cQ_n$, this also gives
    \begin{equation}\label{eq:fixed-space-first-inclusions}
        F(\R^3)\subseteq\Psi(\cQ_n)\subseteq\Psi(\Phi_n).
    \end{equation}

    It remains to show that every parameter fixed by $\Phi_n$ belongs to $F(\R^3)$. Let $\vtheta=(\mW,\vb)\in\Psi(\Phi_n)$. We use the fact that any bijection between two subsets of $[v]$ of the same cardinality can be extended to a permutation of $[v]$.

    First consider the biases. Let $i,a\in[n]$ be arbitrary, write
    \[
        \operatorname{Ind}^{-1}(i)=\{\nu_1,\nu_2\},
        \qquad
        \operatorname{Ind}^{-1}(a)=\{\mu_1,\mu_2\},
    \]
    and choose a bijection sending $\nu_1$ to $\mu_1$ and $\nu_2$ to $\mu_2$. Extend it to a vertex permutation $\phi\in\Pi_v$, and let $\pi_\phi\in\Pi_n^\Phi$ be its induced edge permutation. Then $\pi_\phi(i)=a$. Invariance of $\vtheta$ gives
    \[
        b_i=b_{\pi_\phi(i)}=b_a.
    \]
    Since $i$ and $a$ were arbitrary, all biases have a common value, which we denote by $\beta_3$.

    We next consider weights indexed by two distinct incident edges. Let $i\sim j$ and $a\sim b$. There are distinct vertices $\nu_0,\nu_1,\nu_2$ and $\mu_0,\mu_1,\mu_2$ such that
    \[
        \operatorname{Ind}^{-1}(i)=\{\nu_0,\nu_1\},
        \quad
        \operatorname{Ind}^{-1}(j)=\{\nu_0,\nu_2\},
    \]
    and
    \[
        \operatorname{Ind}^{-1}(a)=\{\mu_0,\mu_1\},
        \quad
        \operatorname{Ind}^{-1}(b)=\{\mu_0,\mu_2\}.
    \]
    The bijection $\nu_r\mapsto\mu_r$ for $r\in\{0,1,2\}$ extends to a vertex permutation $\phi$. Its induced edge permutation satisfies $\pi_\phi(i)=a$ and $\pi_\phi(j)=b$. Hence
    \[
        W_{ij}=W_{\pi_\phi(i)\pi_\phi(j)}=W_{ab}.
    \]
    Therefore all weights indexed by distinct incident edge pairs share a common value, denoted $\beta_1$.

    Finally consider two ordered pairs of disjoint edges: $i\nsim j$ with $i\neq j$, and $a\nsim b$ with $a\neq b$. Write
    \[
        \operatorname{Ind}^{-1}(i)=\{\nu_1,\nu_2\},
        \quad
        \operatorname{Ind}^{-1}(j)=\{\nu_3,\nu_4\},
    \]
    and
    \[
        \operatorname{Ind}^{-1}(a)=\{\mu_1,\mu_2\},
        \quad
        \operatorname{Ind}^{-1}(b)=\{\mu_3,\mu_4\},
    \]
    where the four $\nu$ vertices are distinct and the four $\mu$ vertices are distinct. This is the step for which $v\geq4$ is needed. Extending the bijection $\nu_r\mapsto\mu_r$, $r\in\{1,2,3,4\}$, to a vertex permutation gives an induced edge permutation satisfying $\pi_\phi(i)=a$ and $\pi_\phi(j)=b$. Invariance again yields
    \[
        W_{ij}=W_{ab}.
    \]
    Thus all weights indexed by disjoint edge pairs share a common value, denoted $\beta_2$.

    The diagonal entries of $\mW$ are zero by the definition of $\Theta$. We have therefore shown that every $\vtheta\in\Psi(\Phi_n)$ has the form
    \[
        b_i=\beta_3,
        \qquad
        W_{ij}=
        \begin{cases}
            0, & i=j,\\
            \beta_1, & i\sim j,\\
            \beta_2, & i\neq j\text{ and }i\nsim j,
        \end{cases}
    \]
    and hence $\vtheta=F(\vbeta)$. This proves $\Psi(\Phi_n)\subseteq F(\R^3)$. Combining this with~\eqref{eq:fixed-space-first-inclusions} gives
    \[
        \Psi(\Phi_n)=\Psi(\cQ_n)=F(\R^3),
    \]
    as claimed.
\end{proof}

The next lemma states that parameters invariant to edge adjacency preserving permutations can memorize any binary vector. In combination with Lemma~\ref{lemma:gii-lb} this implies any graph isomorphism class is strictly memorizable.

\ExpressInvParam*

\begin{proof}
Let $\vx \in \{ 0,1\}^n$ satisfy $|supp(\vx)| = m \in [0,n]$. Recall from Lemma~\ref{lem:one-bit-diff} that for any $j \in [n]$
\begin{align*}
    E(\vx^{(j)}; \vtheta) - E(\vx; \vtheta) = y_j(\vx) \langle \vz(\vx), \vtheta_j \rangle,
\end{align*}
where $y_j(\vx) = 1 - 2x_j$, $\vz(\vx) = [\vx,1]$, and $\vtheta_j = [\vw_j,b_j]$. Furthermore,
\begin{align*}
        \langle \vz(\vx), \vtheta_j \rangle
        &=  \vx^T\vw_j + \beta_3\\
        &= \sum_{l=1}^n \1(l \in supp(\vx) \wedge l \neq j) w_{jl} + \beta_3\\
        &= \beta_1 \sum_{l=1}^n \1(l \in supp(\vx) \wedge l \sim j) + \beta_2 \sum_{l=1}^n \1(l \in supp(\vx) \wedge l \nsim j \wedge j \neq l)  + \beta_3.
\end{align*}
Let $c_j(\vx) = \sum_{l=1}^n \1(l \in supp(\vx) \wedge l \sim j)$ denote the number of edges of the graph adjacent to the $j$th edge. Then as
\begin{align*}
m &= \sum_{l = 1}^n \1(l \in supp(\vx) \wedge j \neq l) + \sum_{l = 1}^n \1(l \in supp(\vx) \wedge j = l)\\
&= \sum_{l = 1}^n \1(l \in supp(\vx) \wedge l \sim j \wedge j \neq l) + \sum_{l = 1}^n \1(l \in supp(\vx) \wedge l \nsim j \wedge j \neq l) + \1(j \in supp(\vx))\\
& = \sum_{l = 1}^n \1(l \in supp(\vx) \wedge l \sim j) + \sum_{l = 1}^n \1(l \in supp(\vx) \wedge l \nsim j \wedge j \neq l) + x_j,
\end{align*}
it follows that
\[
\sum_{l = 1}^n \1(l \in supp(\vx) \wedge l \nsim j \wedge j \neq l) = m - c_j(\vx) - x_j.
\]
As a result, the condition that must be satisfied for all $j \in [n]$ is
\[
y_j(\vx) \langle \vz(\vx), \vtheta_j \rangle =
(1 - 2x_j)\left( c_j(\vx)(\beta_1 - \beta_2) + \beta_2(m  - x_j)  + \beta_3\right) \geq 1.
\]
If $\beta_1 = \beta_2$ then the left-hand side simplifies to an expression which depends only on the sparsity of the representation of the graph. Under this assumption, it suffices to find a $\beta_2, \beta_3 \in \R$ such that
\[
(1 - 2x_j)\left(\beta_2(m  - x_j)  + \beta_3\right) \geq 1.
\]
Let $\beta_3 = 1 - \beta_2m$, if $x_j = 0$ then
\[
(1 - 2x_j)\left(\beta_2(m  - x_j)  + \beta_3\right) =  \beta_2 m + \beta_3 = 1
\]
whereas, if $x_j = 1$, then
\[
- (\beta_2(m  - 1)  + \beta_3) = \beta_2 - 1.
\]
Therefore, with $\beta_1 = \beta_2 = 2$ and $\beta_3 = 1 - 2 m$ we have
\[
E(\vx^{(j)}; \vtheta) - E(\vx; \vtheta) \geq 1
\]
for all $j \in [n]$.
\end{proof}

We now make two remarks about the construction in the previous lemma. First, $F(2,2,1-2m)$ satisfies the unit-margin inequalities for $\vx'\in\{0,1\}^n$ if and only if $\|\vx'\|_0=m$. To see this, let $s=\|\vx'\|_0$ and $\delta=s-m$. For an active coordinate $j\in\operatorname{supp}(\vx')$, the energy gap is $1-2\delta$, whereas for an inactive coordinate $j\notin\operatorname{supp}(\vx')$, it is $1+2\delta$. If $0<s<n$, requiring both gaps to be at least one gives $\delta\leq0$ and $\delta\geq0$, hence $\delta=0$. At the boundary $s=0$, only the inactive-coordinate constraints occur, and they require $m=0$; at $s=n$, only the active-coordinate constraints occur, and they require $m=n$. Thus in every case $s=m$.

Second,
\[
\|F(2,2,1-2m)\|^2 = 4n(n-1)+n(1-2m)^2.
\]
Therefore, when $m=\Theta(n)$, the squared norm scales as $\Theta(n^3)$ and the norm as $\Theta(n^{3/2})=\Theta(v^3)$.

\subsection{Constructing an invariant, small norm parameter which memorizes \texorpdfstring{$k$}{k}-cliques}

Our goal in this section is to show that small norm parameters exist which can memorize specific isomorphism classes. In particular, we consider the case of $k$-cliques: recall that a $k$-clique graph has a fully connected subset of $k$ vertices while the remaining $v-k$ vertices are isolated. We denote the set of representations of $k$-cliques on $v$ vertices as $\cC_{v,k}$ and trivially note $|\cC_{v,k}| = \binom{v}{k}$. Towards constructing low-norm invariant parameters that strictly memorize all $k$-cliques, the following lemma derives specific expressions for the energy difference derived in Lemma~\ref{lem:one-bit-diff}. To state this result, for $\vx \in \cC_{v,k}$ let $\textnormal{Clique}(\vx)\subset [v]$ denote the subset of the vertices of the graph which are in the fully connected subset.

\begin{lemma} \label{lem:gii-energy-diff}
    Let $\vbeta \in \R^3$ and suppose $\vtheta = F(\vbeta) =  \mV \vomega$, for some $\vomega \in \R^q$. For $\vx \in \cC_{v,k}$ and any $j \in [n]$, define $r = |\textnormal{Clique}(\vx) \cap \textnormal{Ind}^{-1}(j)| \in \{0,1,2\}$ as the number of vertices in the $j$th vertex pair which are also in the clique of $\vx$. Then for any $j \in [n]$
    \begin{align*}
       \vu_j(\vx)^T \vomega = y_j(\vx)\vz(\vx)^T\vtheta_j =
        \begin{cases}
            \frac{\beta_2}{2} k^2 - \frac{\beta_2}{2} k + \beta_3, &r = 0,\\
            \frac{\beta_2}{2} k^2 + \left( \beta_1 - \frac{3}{2}\beta_2 \right)k + (\beta_2 + \beta_3 - \beta_1), &r = 1,\\
            - \left( \frac{\beta_2}{2} k^2 + \left(2 \beta_1 - \frac{5}{2}\beta_2 \right)k + \left(3 \beta_2 - 4 \beta_1 + \beta_3 \right) \right), & r = 2.
        \end{cases}
    \end{align*}
\end{lemma}

\begin{proof}
    By definition
    \begin{align*}
        \vz(\vx)^T\vtheta_j &= \vw_j^T \vx + b_j\\
        &= \sum_{l =1}^n w_{jl} \1(l \in \textnormal{supp}(\vx)) + \beta_3\\
        &=  \beta_1 \sum_{l =1}^n\1(l \in \textnormal{supp}(\vx) \wedge l \sim j)  +  \beta_2 \sum_{l =1}^n \1(l \in \textnormal{supp}(\vx) \wedge l \nsim j \wedge l \neq j) + \beta_3
    \end{align*}
    Observe each $j \in [n]$ can be placed in one of three distinct categories with respect to $\vx$: in particular, either both, one or neither of the vertices of $j$ are in the $k$-clique of the graph represented by $\vx$. Fixing an arbitrary $j \in [n]$, we denote these events in turn as $\Phi_r$ for $r \in \{0,1,2 \}$, where
    \begin{align*}
        \Phi_r(\vx) = \{ j \in [n] \;: \; |\text{Ind}^{-1}(j)\cap \text{Clique}(\vx)| = r \}.
    \end{align*}
    Note $\Phi_2(\vx) = \textnormal{supp}(\vx)$. If $j \in \Phi_0(\vx)$ then neither of the vertices of $j$ are in the clique of $\vx$, as a result the $j$th edge cannot be adjacent to any edge in the clique. If $j \in \Phi_1(\vx)$ then exactly one vertex of $j$ is in the clique, furthermore there are $k-1$ other vertices in the clique this vertex is connected to via an edge. Finally, if $j \in \Phi_2(\vx)$ then both of its vertices are connected via edges to $k-2$ other vertices in the clique. As a result,
    \begin{align*}
        \sum_{l =1}^n\1(l \in \textnormal{supp}(\vx) \wedge l \sim j) = \begin{cases}
            0,&j \in \Phi_{0}(\vx),\\
            k-1, &j \in \Phi_{1}(\vx),\\
            2(k-2), &j \in \Phi_{2}(\vx).
        \end{cases}
    \end{align*}
    There are $\binom{k}{2}$ edges in the clique. Subtracting the adjacent edges counted above, and excluding edge $j$ itself when $j\in\textnormal{supp}(\vx)=\Phi_2(\vx)$, gives
    \begin{align*}
        \sum_{l =1}^n\1(l \in \textnormal{supp}(\vx) \wedge l \nsim j \wedge l \neq j) = \begin{cases}
            \binom{k}{2},&j \in \Phi_{0}(\vx),\\
            \binom{k}{2} - k+1, &j \in \Phi_{1}(\vx),\\
            \binom{k}{2} - 2k+ 3, &j \in \Phi_{2}(\vx).
        \end{cases}
    \end{align*}
    As a result: if $j \in \Phi_0(\vx)$ then
    \begin{align*}
        \vz(\vx)^T\vtheta_j = \left( \beta_2 \frac{k(k-1)}{2} + \beta_3\right)
        = \frac{\beta_2}{2} k^2 - \frac{\beta_2}{2} k + \beta_3.
    \end{align*}
    If $j \in \Phi_1(\vx)$ then
    \begin{align*}
        \vz(\vx)^T\vtheta_j = \beta_1 (k-1) +  \beta_2 \frac{k^2 - 3k + 2}{2} + \beta_3 = \frac{\beta_2}{2} k^2 + \left( \beta_1 - \frac{3}{2}\beta_2 \right)k + (\beta_2 + \beta_3 - \beta_1).
    \end{align*}
    Finally, if $j \in \Phi_2(\vx)$ then
    \begin{align*}
        \vz(\vx)^T\vtheta_j = \beta_1 (2k - 4) + \beta_2 \frac{k^2 - 5k + 6}{2} + \beta_3 = \frac{\beta_2}{2} k^2 + \left(2 \beta_1 - \frac{5}{2}\beta_2 \right)k + \left(3 \beta_2 - 4 \beta_1 + \beta_3 \right).
    \end{align*}
    To conclude, observe $y_j(\vx) = 1 - 2x_j = -1$ iff $j \in \Phi_2(\vx)$.
\end{proof}

We now derive a simple bound on the norm of parameters which are invariant to edge adjacency preserving permutations.

\begin{lemma} \label{lem:gii-norm}
    Let $\vbeta \in \R^3$ and $\vtheta = F(\vbeta)$. Then
    \[
    \| \vtheta \|^2 \leq \beta_2^2 v^4 + 2 \beta_1^2 v^3 + \beta_3^2 v^2.
    \]
\end{lemma}
\begin{proof}
    For any fixed edge index $r \in [n]$, note as each vertex of this edge is a member of $v-2$ other vertex pairs then
    \begin{align*}
        \sum_{c = 1}^n \1(c \sim r) = 2(v-2)
    \end{align*}
    Moreover, as there are $\binom{v}{2}$ unordered vertex pairs in total
    \begin{align*}
        \sum_{c = 1}^n \1(c \nsim r \wedge c \neq r) = \binom{v}{2} - 2(v-2) - 1 = \frac{v^2 - 5v + 6}{2}.
    \end{align*}
    Therefore, and also noting that $n = \binom{v}{2} \leq v^2$, we have
    \begin{align*}
        \| \vtheta \|^2 &= \| \mW \|_F^2 + \| \vb \|^2\\
        &= \sum_{r=1}^n \left(\beta_1^2 \sum_{c=1}^n \1(c \sim r) + \beta_2^2 \sum_{c=1}^n \1(c \nsim r \wedge c \neq r) + \beta_3^2\right) \\
        &= n\left( \beta_1^2 2(v-2) + \beta_2^2 \left(\frac{v^2 - 5v + 6}{2} \right) + \beta_3^2 \right)\\
        & \leq n (\beta_2^2 v^2 + 2 \beta_1^2 v + \beta_3^2)\\
        & \leq \beta_2^2 v^4 + 2 \beta_1^2 v^3 + \beta_3^2 v^2.
    \end{align*}
\end{proof}

We now present a low-norm construction for memorizing $k$-cliques. In particular, Lemma~\ref{lem:gii-strict-mem-ex} shows that the $k$-clique graph-isomorphism class can be memorized using a parameter whose norm is $O(\sqrt{v})$. This contrasts with the general construction in Lemma~\ref{lemma:express-inv-param}, whose norm is $\Theta(n^{3/2})=\Theta(v^3)$ when $m=\Theta(n)$.

\NormKClique*

\begin{proof}
Using Lemma~\ref{lem:gii-energy-diff}, and substituting $\beta_1=-10/k$, $\beta_2=38/k^2$ and $\beta_3=0$, we obtain
\[
\vu_j(\vx)^T\vomega =
\begin{cases}
19\left(1-\frac{1}{k}\right), & r=0,\\
9-\frac{47}{k}+\frac{38}{k^2}, & r=1,\\
1+\frac{55}{k}-\frac{114}{k^2}, & r=2.
\end{cases}
\]
If $k\geq 5$, then
\[
19 \left(1-\frac{1}{k}\right) \geq 1,
\]
while
\[
9-\frac{47}{k}+\frac{38}{k^2} \geq 9-\frac{47}{5}+\frac{38}{25} > 1,
\]
and
\[
1+\frac{55}{k}-\frac{114}{k^2} \geq 1.
\]
Hence $E(\vx^{(j)};\vtheta)-E(\vx;\vtheta)\geq 1$ for all $\vx\in\cC_{v,k}$ and all $j\in[n]$.

It remains to bound the norm. Since $F(\vbeta)$ assigns weight $\beta_1$ to adjacent edge pairs, weight $\beta_2$ to non-adjacent edge pairs, and has zero bias, we have
\[
    \|\vtheta\|^2
    =
    \|\mW\|_F^2
    \leq
    n 2(v-2)\beta_1^2 + n(n-1)\beta_2^2.
\]
Since $n=\binom{v}{2}$, $\beta_1^2=100/k^2$, and $\beta_2^2=38^2/k^4$, if $k=cv$ then
\[
    n 2(v-2)\beta_1^2 = O(v),
    \qquad
    n(n-1)\beta_2^2 = O(1).
\]
Therefore there exists a constant $C>0$, depending only on $c$, such that $\|\vtheta\|^2\leq Cv$.
\end{proof}


\subsubsection{Invariance of the full orbit HSVM solution}

The following lemma states a well known result that the average orbit action of a parameter is invariant to the action of the underlying group.
\begin{lemma}\label{lem:projection-inv}
    Let $\vtheta = (\mW, \vb) \in \Theta$ and $\Gamma_n$ denote a subgroup of $\cP_n$. Then $\textnormal{Proj}_{\Gamma_n}(\vtheta) = \frac{1}{|\Gamma_n|}\sum_{\mQ \in \Gamma_n} \mQ \vtheta \in \Psi(\Gamma_n)$.
\end{lemma}

\begin{proof}
    For typographical ease let $\vtheta' = \textnormal{Proj}_{\Gamma_n}(\vtheta)$. Then
    \[
    \vtheta' = (\mW', \vb') := \frac{1}{|\Gamma_n|}\sum_{\mQ \in \Gamma_n} \mQ\vtheta = \left(\frac{1}{|\Gamma_n|}\sum_{\mQ \in \Gamma_n} \mQ^T \mW \mQ, \frac{1}{|\Gamma_n|}\sum_{\mQ \in \Gamma_n} \mQ^T \vb \right).
    \]
    Given $\Gamma_n$ is a subgroup, then for any $\mQ' \in \Gamma_n$
    \begin{align*}
        \mQ'\vtheta' &= \left(\mQ'^T\mW' \mQ', \mQ'^T \vb' \right)\\
        &= \left(\frac{1}{|\Gamma_n|}\sum_{\mQ \in \Gamma_n} (\mQ\mQ')^T \mW (\mQ\mQ'), \frac{1}{|\Gamma_n|}\sum_{\mQ \in \Gamma_n} (\mQ\mQ')^T \vb \right)\\
        &= \left(\frac{1}{|\Gamma_n|}\sum_{\mQ \in \Gamma_n} \mQ^T \mW \mQ, \frac{1}{|\Gamma_n|}\sum_{\mQ \in \Gamma_n} \mQ^T \vb \right)\\
        &= \vtheta',
    \end{align*}
    therefore $\vtheta' \in \Psi(\Gamma_n)$.
\end{proof}

Using the previous lemma, we show that the full orbit HSVM solution lies in the invariant set.

\OrbitHSVMInv*

\begin{proof}
    Let $\cO=\operatorname{Orb}(\vx_0,\Gamma_n)$. We first recall that the HSVM may equivalently be written directly in terms of $\vtheta=(\mW,\vb)\in\Theta$. Indeed, $\mV$ gives a one-to-one correspondence between the parameter vector $\vomega$ and $\vtheta$, and
    \[
        \|\vomega\|^2=\|\mW\|_F^2+\|\vb\|^2.
    \]
    Thus $\vtheta^*=\mV\vomega^*$ is the unique solution of
    \begin{equation}\label{eq:theta-orbit-hsvm}
        \min_{\vtheta=(\mW,\vb)\in\Theta}
        \bigl(\|\mW\|_F^2+\|\vb\|^2\bigr)
        \quad\text{subject to}\quad
        E(\vx^{(j)};\vtheta)-E(\vx;\vtheta)\geq1
    \end{equation}
    for every $\vx\in\cO$ and $j\in[n]$. Uniqueness follows because the feasible set is convex and the squared-norm objective is strictly convex.

    Fix an arbitrary $\mQ\in\Gamma_n$, and let $\pi_{\mQ}$ denote the permutation of the coordinates represented by $\mQ$, with the convention that
    \begin{equation}\label{eq:permutation-commutes-with-flip}
        \mQ\vx^{(j)}=(\mQ\vx)^{(\pi_{\mQ}(j))}
        \qquad\text{for every }\vx\in\{0,1\}^n\text{ and }j\in[n].
    \end{equation}
    We claim that the transformed parameter $\mQ\vtheta^*$ is also feasible for~\eqref{eq:theta-orbit-hsvm}. By the intertwining identity~\eqref{eq:intertwining} and~\eqref{eq:permutation-commutes-with-flip}, for every $\vx\in\cO$ and $j\in[n]$,
    \begin{align*}
        E(\vx^{(j)};\mQ\vtheta^*)-E(\vx;\mQ\vtheta^*)
        &=E(\mQ\vx^{(j)};\vtheta^*)-E(\mQ\vx;\vtheta^*)\\
        &=E((\mQ\vx)^{(\pi_{\mQ}(j))};\vtheta^*)-E(\mQ\vx;\vtheta^*)\\
        &\geq1.
    \end{align*}
    The final inequality holds because $\mQ\vx\in\cO$ and $\pi_{\mQ}(j)\in[n]$, so the preceding expression is one of the constraints in the full-orbit problem. This is the step that uses the fact that the HSVM is defined on all of $\cO$, rather than on an arbitrary subset of the orbit.

    The transformation also preserves the objective. Writing $\vtheta^*=(\mW^*,\vb^*)$, we have
    \begin{align*}
        \|\mQ\vtheta^*\|^2
        &=\|\mQ^T\mW^*\mQ\|_F^2+\|\mQ^T\vb^*\|^2\\
        &=\|\mW^*\|_F^2+\|\vb^*\|^2
        =\|\vtheta^*\|^2,
    \end{align*}
    where the second equality follows because $\mQ$ is a permutation matrix. Hence $\mQ\vtheta^*$ is feasible and attains the same objective value as the unique minimizer $\vtheta^*$. It follows that
    \[
        \mQ\vtheta^*=\vtheta^*.
    \]
    Since $\mQ\in\Gamma_n$ was arbitrary, $\vtheta^*$ is fixed by every element of $\Gamma_n$, and therefore $\vtheta^*\in\Psi(\Gamma_n)$.
\end{proof}

\subsection{Approximate invariance and orbit margins}\label{app:invariance2}

Recall that $\vtheta=\mV\vomega$. For a subgroup \(\Gamma_n\leq\cP_n\), define
\[
    \cJ_{\Gamma_n}=\{\vomega\in\R^q:\mV\vomega\in\Psi(\Gamma_n)\},
\]
so $\cJ_{\Gamma_n}$ is the set of $\vomega$ parameters whose corresponding Hopfield parameter $\vtheta=\mV\vomega$ is invariant under $\Gamma_n$. Let \(\operatorname{Proj}_{\cJ_{\Gamma_n}}\) denote orthogonal projection onto this subspace. The following lemma relates proximity to this invariant subspace to the margins attained across an orbit.

\begin{restatable}{lemma}{ProximityInvariant}\label{lemma:proximity-invariant}
    For an $\vx_0 \in \{0,1\}^n$, suppose $\vomega$ satisfies \(\langle\vu_j(\vx_0),\vomega\rangle\geq1\) for every \(j\in[n]\). Let \(\cO=\operatorname{Orb}(\vx_0,\Gamma_n)\) and
    \[
        R=\max_{\vx\in\cO,\,j\in[n]}\|\vu_j(\vx)\|,
        \qquad
        \eta=\|\operatorname{Proj}_{\cJ_{\Gamma_n}}^\perp(\vomega)\|.
    \]
    If \(\vtheta=\mV\vomega\) then
    \[
        E(\vx^{(j)};\vtheta)-E(\vx;\vtheta)\geq1-2R\eta
        \qquad
        \text{for every }\vx\in\cO,\ j\in[n].
    \]
    In particular, \(\eta<1/(2R)\) gives strict memorization of \(\cO\), and \(\eta\leq1/(4R)\) gives margin at least \(1/2\).
\end{restatable}

\begin{proof}
    Write \(\vomega_0=\operatorname{Proj}_{\cJ_{\Gamma_n}}(\vomega)\). At the representative \(\vx_0\), Cauchy--Schwarz gives
    \[
        \langle\vu_j(\vx_0),\vomega_0\rangle
        \geq
        \langle\vu_j(\vx_0),\vomega\rangle
        -\|\vu_j(\vx_0)\|\,\|\vomega-\vomega_0\|
        \geq1-R\eta.
    \]
    Since \(\mV\vomega_0\in\Psi(\Gamma_n)\), Lemma~\ref{lemma:gii-lb} propagates this bound to every \(\vx\in\cO\) and every coordinate \(j\). Applying Cauchy--Schwarz once more, for every $\vx \in \cO$ we have
    \[
        \langle\vu_j(\vx),\vomega\rangle
        \geq\langle\vu_j(\vx),\vomega_0\rangle
             -\|\vu_j(\vx)\|\,\|\vomega-\vomega_0\|
        \geq1-2R\eta.
    \]
    Lemma~\ref{lem:one-bit-diff} identifies this inner product with the claimed energy gap.
\end{proof}

\subsection{Concentration of the exact sample HSVM}\label{app:exact-hsvm-concentration}

Let \(\cO=\operatorname{Orb}(\vx_0,\Gamma_n)\), \(M=|\cO|\), \(\vomega^*=\operatorname{HSVM}(\cO)\), and \(R=\max_{\vx\in\cO,j\in[n]}\|\vu_j(\vx)\|\). Each constraint in the full-orbit HSVM is labelled by a pair
\[
    (\vx,j)\in\cO\times[n]
\]
and has the equivalent forms
\[
    \langle\vu_j(\vx),\vomega\rangle\geq1
    \qquad\Longleftrightarrow\qquad
    E(\vx^{(j)};\vtheta)-E(\vx;\vtheta)\geq1,
    \quad \vtheta=\mV\vomega.
\]
Thus every $\vx\in\cO$ contributes one constraint for each of its $n$ one-bit neighbours. We regard these as labelled constraints, so two different pairs remain distinct labels even if they happen to define the same linear inequality.

For each \(\mQ\in\Gamma_n\), define its action on the constraint labels by
\[
    \mQ\cdot(\vx,j)=(\mQ\vx,\pi_{\mQ}(j)),
\]
where \(\pi_{\mQ}\) is defined by~\eqref{eq:permutation-commutes-with-flip}. This is the appropriate action because
\[
    \mQ\vx^{(j)}=(\mQ\vx)^{(\pi_{\mQ}(j))}.
\]
Since \(\mQ\vx\in\cO\) whenever \(\vx\in\cO\), and since $\pi_{\mQ}$ is a permutation of $[n]$, this action permutes the finite collection \(\cO\times[n]\) of full-orbit constraint labels. This action also describes exactly how the numerical values of the constraints transform. For any parameter $\vtheta$ and any constraint label $(\vx,j)$, the intertwining identity gives
\begin{equation}\label{eq:energy-gap-constraint-action}
\begin{aligned}
    E(\vx^{(j)};\mQ\vtheta)-E(\vx;\mQ\vtheta)
    &=E(\mQ\vx^{(j)};\vtheta)-E(\mQ\vx;\vtheta)\\
    &=E((\mQ\vx)^{(\pi_{\mQ}(j))};\vtheta)-E(\mQ\vx;\vtheta).
\end{aligned}
\end{equation}
Thus the value of constraint $(\vx,j)$ at the transformed parameter $\mQ\vtheta$ is the value of constraint $(\mQ\vx,\pi_{\mQ}(j))$ at the original parameter $\vtheta$.

We also recall the standard KKT representation of an HSVM solution. There are nonnegative coefficients $\alpha_{\vx j}$ such that
\begin{equation}\label{eq:standard-hsvm-dual-certificate}
    \vomega^*=\sum_{\vx\in\cO}\sum_{j=1}^n
    \alpha_{\vx j}\vu_j(\vx),
    \qquad
    \alpha_{\vx j}>0
    \ \Longrightarrow\
    \langle\vu_j(\vx),\vomega^*\rangle=1.
\end{equation}
Here the first identity is the stationarity condition, with the multipliers rescaled to account for the normalization of the squared-norm objective, and the implication is complementary slackness. We call any collection of coefficients satisfying~\eqref{eq:standard-hsvm-dual-certificate} a dual certificate. This representation is standard; the content of the next lemma is that, for the full-orbit problem, a certificate can be chosen to respect the group action and its total coefficient mass is then distributed uniformly over the orbit.

\begin{lemma}[Invariant dual certificate]\label{lemma:invariant-dual-certificate}
    The full-orbit HSVM admits a dual certificate \(\bar\alpha_{\vx j}\) that is invariant under the induced action on \(\cO\times[n]\), meaning
    \[
        \bar\alpha_{\mQ\vx,\pi_{\mQ}(j)}=\bar\alpha_{\vx j}
        \qquad
        \text{for every }\mQ\in\Gamma_n,\ \vx\in\cO,\ j\in[n].
    \]
    Moreover,
    \[
        \sum_{j=1}^n\bar\alpha_{\vx j}=\frac{\|\vomega^*\|^2}{M}
        \qquad\text{for every }\vx\in\cO.
    \]
\end{lemma}

\begin{proof}
    Let $\mathcal I=\cO\times[n]$, write $s=(\vx,j)$ for a constraint label, and abbreviate $\vu_s=\vu_j(\vx)$. Begin with any dual certificate $(\alpha_s)_{s\in\mathcal I}$ supplied by~\eqref{eq:standard-hsvm-dual-certificate}. Fix $\mQ\in\Gamma_n$. Since $\vtheta=\mV\vomega$ is a one-to-one parametrization of $\Theta$, there is a unique linear map $\mT_{\mQ}:\R^q\to\R^q$ satisfying
    \[
        \mV\mT_{\mQ}\vomega=\mQ(\mV\vomega)
        \qquad\text{for every }\vomega\in\R^q.
    \]
    Applying $\mQ$ only permutes the entries of $\mW$ and $\vb$. Consequently it preserves the parameter norm, and hence
    \[
        \|\mT_{\mQ}\vomega\|
        =\|\mV\mT_{\mQ}\vomega\|
        =\|\mQ(\mV\vomega)\|
        =\|\mV\vomega\|
        =\|\vomega\|.
    \]
    Thus $\mT_{\mQ}$ is orthogonal. By the definition of the constraint vectors, the energy gap associated with $s=(\vx,j)$ at a parameter $\mV\vomega$ is $\langle\vu_s,\vomega\rangle$. Applying~\eqref{eq:energy-gap-constraint-action} with $\vtheta=\mV\vomega$, together with the definition of $\mT_{\mQ}$, therefore gives
    \[
        \langle\vu_s,\mT_{\mQ}\vomega\rangle
        =\langle\vu_{\mQ\cdot s},\vomega\rangle.
    \]
    By the definition of the transpose, the left-hand side is
    $\langle\mT_{\mQ}^T\vu_s,\vomega\rangle$. Hence
    \[
        \langle\mT_{\mQ}^T\vu_s-\vu_{\mQ\cdot s},\vomega\rangle=0
        \qquad\text{for every }\vomega\in\R^q.
    \]
    Taking $\vomega=\mT_{\mQ}^T\vu_s-\vu_{\mQ\cdot s}$ shows that this difference has squared norm zero. Therefore
    \begin{equation}\label{eq:constraint-vector-action}
        \vu_{\mQ\cdot s}=\mT_{\mQ}^T\vu_s.
    \end{equation}
    Lemma~\ref{lemma:orbit-hsvm-inv} gives $\mQ\vtheta^*=\vtheta^*$, so injectivity of $\mV$ also gives
    \begin{equation}\label{eq:omega-fixed-by-action}
        \mT_{\mQ}\vomega^*=\vomega^*.
    \end{equation}

    Now relabel the original certificate by defining
    \[
        \alpha^{\mQ}_s=\alpha_{\mQ^{-1}\cdot s}
        \qquad\text{for every }s\in\mathcal I,
    \]
    so the coefficient assigned to $s$ after relabelling is the coefficient originally assigned to its preimage under $\mQ$. These coefficients remain nonnegative. Using~\eqref{eq:constraint-vector-action} and changing the summation index from $s$ to $\mQ\cdot s$, their stationarity sum is
    \begin{align*}
        \sum_{s\in\mathcal I}\alpha^{\mQ}_s\vu_s
        &=\sum_{s\in\mathcal I}\alpha_{\mQ^{-1}\cdot s}\vu_s\\
        &=\sum_{s\in\mathcal I}\alpha_s\vu_{\mQ\cdot s}\\
        &=\mT_{\mQ}^T\sum_{s\in\mathcal I}\alpha_s\vu_s
        =\mT_{\mQ}^T\vomega^*
        =\vomega^*.
    \end{align*}
    The last equality follows from~\eqref{eq:omega-fixed-by-action} because $\mT_{\mQ}$ is orthogonal. Complementary slackness is also preserved. Indeed, if $\alpha^{\mQ}_s>0$, set $r=\mQ^{-1}\cdot s$. Then $\alpha_r>0$, $s=\mQ\cdot r$, and
    \[
        \langle\vu_s,\vomega^*\rangle
        =\langle\vu_{\mQ\cdot r},\vomega^*\rangle
        =\langle\mT_{\mQ}^T\vu_r,\vomega^*\rangle
        =\langle\vu_r,\mT_{\mQ}\vomega^*\rangle
        =\langle\vu_r,\vomega^*\rangle
        =1.
    \]
    Therefore $(\alpha^{\mQ}_s)_{s\in\mathcal I}$ is another dual certificate for every $\mQ\in\Gamma_n$.

    Average these certificates by setting
    \[
        \bar\alpha_s=\frac{1}{|\Gamma_n|}
        \sum_{\mQ\in\Gamma_n}\alpha^{\mQ}_s.
    \]
    Nonnegativity, stationarity, and complementary slackness are preserved by this average, so $(\bar\alpha_s)_{s\in\mathcal I}$ is a dual certificate. For any $\mR\in\Gamma_n$,
    \begin{align*}
        \bar\alpha_{\mR\cdot s}
        &=\frac{1}{|\Gamma_n|}
          \sum_{\mQ\in\Gamma_n}
          \alpha_{\mQ^{-1}\cdot(\mR\cdot s)}\\
        &=\frac{1}{|\Gamma_n|}
          \sum_{\mP\in\Gamma_n}
          \alpha_{\mP^{-1}\cdot s}
        =\bar\alpha_s,
    \end{align*}
    where the second equality uses the bijective change of index $\mP=\mR^{-1}\mQ$. Thus the averaged certificate is invariant under the induced action. Taking its inner product with \(\vomega^*\) and using complementary slackness yields
    \[
        \sum_{\vx,j}\bar\alpha_{\vx j}
        =\left\langle\sum_{\vx,j}\bar\alpha_{\vx j}\vu_j(\vx),\vomega^*\right\rangle
        =\|\vomega^*\|^2.
    \]
    Finally, define $A(\vx)=\sum_{j=1}^n\bar\alpha_{\vx j}$. If $\vx'=\mQ\vx$, invariance of the certificate and the fact that $\pi_{\mQ}$ permutes $[n]$ give
    \[
        A(\vx')
        =\sum_{j=1}^n\bar\alpha_{\mQ\vx,\pi_{\mQ}(j)}
        =\sum_{j=1}^n\bar\alpha_{\vx j}
        =A(\vx).
    \]
    The action is transitive on $\cO$, so $A(\vx)$ is constant over the $M$ elements of $\cO$. Since $\sum_{\vx\in\cO}A(\vx)=\|\vomega^*\|^2$, this constant is $\|\vomega^*\|^2/M$, proving the final identity.
\end{proof}

Lemma~\ref{lemma:invariant-dual-certificate} gives a particularly simple population dual certificate. Every orbit element carries the same collection of $n$ coefficients, up to the coordinate relabelling induced by the group action, and contributes the same total dual mass. The next lemma packages these coefficients into a contribution $\vg(\vx)$ whose uniform average is the population HSVM solution. It then reduces comparison of the sample and population HSVM solutions to concentration of the empirical average of these contributions.

\begin{lemma}[Empirical certificate comparison]\label{lemma:empirical-certificate-comparison}
    Let $(\bar\alpha_{\vx j})$ be the invariant dual certificate from Lemma~\ref{lemma:invariant-dual-certificate}, and define
    \[
        \vg(\vx)=M\sum_{j=1}^n
        \bar\alpha_{\vx j}\vu_j(\vx).
    \]
    The function $\vg$ satisfies
    \[
        \E_{\vx\sim U(\cO)}[\vg(\vx)]=\vomega^*.
    \]
    Furthermore, for every $\vx\in\cO$,
    \begin{equation}\label{eq:empirical-certificate-radius}
        \|\vg(\vx)-\vomega^*\|
        \leq \|\vomega^*\|^2R+\|\vomega^*\|
        \leq2\|\vomega^*\|^2R.
    \end{equation}
    Moreover, for any sample $\cS_N=(\vx_i)_{i\in[N]}$, if
    \[
        \vg_N=\frac1N\sum_{i=1}^N\vg(\vx_i)
        \qquad\text{and}\qquad
        \widehat{\vomega}_N=\operatorname{HSVM}(\cS_N),
    \]
    then
    \begin{equation}\label{eq:empirical-certificate-comparison}
        \|\widehat{\vomega}_N-\vomega^*\|
        \leq2\|\vg_N-\vomega^*\|.
    \end{equation}
\end{lemma}

\begin{proof}
    Lemma~\ref{lemma:invariant-dual-certificate} gives
    \[
    \begin{aligned}
        \E_{\vx\sim U(\cO)}[\vg(\vx)]
        &=\frac1M\sum_{\vx\in\cO}\sum_{j=1}^n
          M\bar\alpha_{\vx j}\vu_j(\vx)
        =\vomega^*.
    \end{aligned}
    \]
    Since the coefficients $\bar\alpha_{\vx j}$ are nonnegative, Lemma~\ref{lemma:invariant-dual-certificate} and $\|\vu_j(\vx)\|\leq R$ give
    \[
        \|\vg(\vx)\|
        \leq M\sum_{j=1}^n
        \bar\alpha_{\vx j}\|\vu_j(\vx)\|
        \leq \|\vomega^*\|^2R.
    \]
    Feasibility of $\vomega^*$ gives $1\leq\langle\vu_j(\vx),\vomega^*\rangle\leq R\|\vomega^*\|$ for every constraint. Consequently,
    \[
        \|\vg(\vx)-\vomega^*\|
        \leq \|\vomega^*\|^2R+\|\vomega^*\|
        \leq2\|\vomega^*\|^2R.
    \]

    Let \(\vd=\widehat{\vomega}_N-\vomega^*\). The population solution $\vomega^*$ satisfies every constraint in $\cO$, so in particular it is feasible for the sampled HSVM problem. Since $\widehat{\vomega}_N$ is the minimum-norm solution of that problem,
    \[
        \|\widehat{\vomega}_N\|^2\leq\|\vomega^*\|^2.
    \]
    Substituting $\widehat{\vomega}_N=\vomega^*+\vd$ and expanding the squared norm gives
    \[
        \|\vomega^*\|^2
        +2\langle\vomega^*,\vd\rangle
        +\|\vd\|^2
        \leq\|\vomega^*\|^2.
    \]
    Cancelling $\|\vomega^*\|^2$ and rearranging yields
    \begin{equation}\label{eq:sample-population-norm-comparison}
        \langle\vomega^*,\vd\rangle
        \leq-\frac12\|\vd\|^2.
    \end{equation}
    Next, using the definition of $\vg_N$ and $\vd$, we have
    \[
        \langle\vg_N,\vd\rangle
        =\frac{M}{N}\sum_{i,j}\bar\alpha_{\vx_i j}
          \bigl(\langle\vu_j(\vx_i),\widehat{\vomega}_N\rangle
                -\langle\vu_j(\vx_i),\vomega^*\rangle\bigr).
    \]
    Every summand is nonnegative. If $\bar\alpha_{\vx_i j}=0$, the summand is zero. If $\bar\alpha_{\vx_i j}>0$, complementary slackness for the population certificate gives
    \[
        \langle\vu_j(\vx_i),\vomega^*\rangle=1,
    \]
    while feasibility of $\widehat{\vomega}_N$ for every sampled constraint gives
    \[
        \langle\vu_j(\vx_i),\widehat{\vomega}_N\rangle\geq1.
    \]
    Since $M/N>0$, it follows that
    \begin{equation}\label{eq:empirical-certificate-positive-inner-product}
        \langle\vg_N,\vd\rangle\geq0.
    \end{equation}
    Combining~\eqref{eq:sample-population-norm-comparison} and~\eqref{eq:empirical-certificate-positive-inner-product},
    \begin{align*}
        \langle\vg_N-\vomega^*,\vd\rangle =\langle\vg_N,\vd\rangle
          -\langle\vomega^*,\vd\rangle \geq-\langle\vomega^*,\vd\rangle \geq\frac12\|\vd\|^2.
    \end{align*}
    On the other hand, Cauchy--Schwarz gives
    \[
        \langle\vg_N-\vomega^*,\vd\rangle
        \leq\|\vg_N-\vomega^*\|\,\|\vd\|.
    \]
    Combining the lower and upper bounds gives
    \[
        \frac12\|\vd\|^2
        \leq\|\vg_N-\vomega^*\|\,\|\vd\|.
    \]
    If $\vd\neq0$, dividing both sides by $\|\vd\|/2$ yields
    \[
        \|\vd\|\leq2\|\vg_N-\vomega^*\|.
    \]
    Recalling that $\vd=\widehat{\vomega}_N-\vomega^*$ gives~\eqref{eq:empirical-certificate-comparison}. If $\vd=0$, that inequality holds immediately.
\end{proof}

We are now in a position to prove the key result of this section: a concentration bound on the distance between the sample and population HSVM solutions.

\ExactHSVMConcentration*

\begin{proof}
    By Lemma~\ref{lemma:empirical-certificate-comparison}, the centered random vectors \(\vg(\vx_i)-\vomega^*\) are independent and have mean zero. If \(\|\widehat{\vomega}_N-\vomega^*\|>t\), inequality~\eqref{eq:empirical-certificate-comparison} implies \(\|\vg_N-\vomega^*\|>t/2\). Applying Lemma~\ref{lemma:vectorHoeffding} therefore gives the certificate-sensitive bound
    \begin{equation}\label{eq:certificate-sensitive-concentration}
        \P\!\left(\|\widehat{\vomega}_N-\vomega^*\|>t\right)
        \leq2\exp\!\left(
            -\frac{Nt^2}
            {8\bigl(\max_{\vx\in\cO}\|\vg(\vx)-\vomega^*\|\bigr)^2}
        \right).
    \end{equation}
    Using~\eqref{eq:empirical-certificate-radius} gives
    \[
        \P\!\left(\|\widehat{\vomega}_N-\vomega^*\|>t\right)
        \leq2\exp\!\left(-\frac{Nt^2}{32\|\vomega^*\|^4R^2}\right).
    \]
\end{proof}

\UniformOrbitMemorization*

\begin{proof}
    On the event \(\|\widehat{\vomega}_N-\vomega^*\|\leq1/(2R)\), every orbit constraint satisfies
    \[
        \langle\vu_j(\vx),\widehat{\vomega}_N\rangle
        \geq\langle\vu_j(\vx),\vomega^*\rangle
             -R\|\widehat{\vomega}_N-\vomega^*\|
        \geq\frac12.
    \]
    Theorem~\ref{thm:exact-hsvm-concentration} with \(t=1/(2R)\) bounds the complementary event by
    \[
        2\exp\!\left(-\frac{N}{128\|\vomega^*\|^4R^4}\right)\leq\delta
    \]
    under the stated sample-size condition. The conclusion follows from Lemma~\ref{lem:one-bit-diff}.
\end{proof}

The bound is polynomial in the ambient problem size without any assumption on the orbit cardinality. Indeed, if every orbit element has Hamming weight $m$, the universal invariant construction of Lemma~\ref{lemma:express-inv-param} has
\[
    \|\vomega^*\|^2\leq4n(n-1)+n(1-2m)^2,
    \qquad
    R^2\leq m/2+1.
\]
Since the population HSVM has no larger norm than this feasible construction, Corollary~\ref{cor:uniform-orbit-memorization} gives an explicit polynomial sufficient sample size that is independent of $|\cO|$.

\subsection{Clique specialization}

\begin{lemma}[Clique certificate deviation]\label{lemma:clique-certificate-deviation}
    Assume $k=cv\geq5$ for a constant $c\in(0,1)$, let $M=|\cC_{v,k}|$, and let $\vomega^*=\operatorname{HSVM}(\cC_{v,k})$. Let $\vg$ be the orbit contribution constructed in Lemma~\ref{lemma:empirical-certificate-comparison}, and set
    \[
        m=\binom{k}{2},
        \qquad
        s_{\min}=\min\left(
        \left\{
        \binom{k}{2},\ k(v-k),\ \binom{v-k}{2}
        \right\}\setminus\{0\}\right).
    \]
    Then
    \begin{equation}\label{eq:clique-certificate-deviation}
        \max_{\vx\in\cC_{v,k}}\|\vg(\vx)-\vomega^*\|
        \leq\|\vomega^*\|^2\sqrt{\frac{m+1}{s_{\min}}}
             +\|\vomega^*\|
        =O_c(v).
    \end{equation}
\end{lemma}

\begin{proof}
    Fix a clique \(\vx\in\cC_{v,k}\), and let $S=\textnormal{Clique}(\vx)\subseteq[v]$ be its set of $k$ fully connected vertices. Using the notation of Lemma~\ref{lem:gii-energy-diff}, define
    \[
        \Phi_r(\vx)
        =\left\{j\in[n]:
        |\textnormal{Ind}^{-1}(j)\cap S|=r\right\},
        \qquad r\in\{0,1,2\}.
    \]
    Thus $\Phi_2(\vx)=\operatorname{supp}(\vx)$ indexes edges inside $S$, $\Phi_1(\vx)$ indexes edges with one endpoint in $S$, and $\Phi_0(\vx)$ indexes edges outside $S$. Their respective cardinalities are
    \[
        |\Phi_2(\vx)|=\binom{k}{2},\qquad
        |\Phi_1(\vx)|=k(v-k),\qquad
        |\Phi_0(\vx)|=\binom{v-k}{2}.
    \]

    Consider a vertex permutation $\phi:[v]\to[v]$ satisfying $\phi(S)=S$; equivalently, $\phi$ permutes the vertices in $S$ among themselves and the vertices outside $S$ among themselves. Let $\pi_\phi$ be the edge-coordinate permutation induced by $\phi$ as in Definition~\ref{def:vertex-induced-edge-perm}, and let $\mQ_\phi\in\Phi_n$ be its permutation matrix. Because $\phi$ preserves membership in $S$, the induced permutation maps each set $\Phi_r(\vx)$ onto itself. In particular, it maps the support $\Phi_2(\vx)$ onto itself, and hence
    \[
        \mQ_\phi\vx=\vx.
    \]

    Now fix $r\in\{0,1,2\}$ and let $j,j'\in\Phi_r(\vx)$. We can choose a vertex permutation $\phi$ with $\phi(S)=S$ whose induced edge permutation satisfies $\pi_\phi(j)=j'$. Applying the certificate invariance identity from Lemma~\ref{lemma:invariant-dual-certificate} then gives
    \[
        \bar\alpha_{\vx j'}
        =\bar\alpha_{\mQ_\phi\vx,\pi_\phi(j)}
        =\bar\alpha_{\vx j}.
    \]
    Thus the coefficients $\bar\alpha_{\vx j}$ are constant on each set $\Phi_r(\vx)$. They are nonnegative and satisfy
    \[
        M\sum_{j=1}^n\bar\alpha_{\vx j}=\|\vomega^*\|^2.
    \]
    For every nonempty class $\Phi_r(\vx)$, let
    \[
        \mu_r=M\sum_{j\in\Phi_r(\vx)}\bar\alpha_{\vx j}
    \]
    be its total scaled coefficient mass. All sums over $r$ below range over the nonempty classes. Constancy within the class implies
    \[
        M\bar\alpha_{\vx j}
        =\frac{\mu_r}{|\Phi_r(\vx)|}
        \qquad\text{for every }j\in\Phi_r(\vx).
    \]
    Therefore
    \begin{align*}
        M^2\sum_{j=1}^n\bar\alpha_{\vx j}^2
        &=\sum_{r:\,\Phi_r(\vx)\neq\varnothing}
          |\Phi_r(\vx)|
          \left(\frac{\mu_r}{|\Phi_r(\vx)|}\right)^2\\
        &=\sum_{r:\,\Phi_r(\vx)\neq\varnothing}
          \frac{\mu_r^2}{|\Phi_r(\vx)|}\\
        &\leq\frac{1}{s_{\min}}\sum_r\mu_r^2
        \leq\frac{1}{s_{\min}}\left(\sum_r\mu_r\right)^2
        =\frac{\|\vomega^*\|^4}{s_{\min}}.
    \end{align*}
    Here the penultimate inequality uses $\mu_r\geq0$, and the final equality uses $\sum_r\mu_r=M\sum_j\bar\alpha_{\vx j}=\|\vomega^*\|^2$.
    We next bound $\|\vg(\vx)\|$ using the coordinates of the $\vomega$ parametrization. The bias coordinate indexed by $a$ in
    \(
        \vg(\vx)=M\sum_j\bar\alpha_{\vx j}\vu_j(\vx)
    \)
    is
    \[
        M\bar\alpha_{\vx a}y_a(\vx).
    \]
    For $a<b$, the coordinate corresponding to the shared symmetric weight between network coordinates $a$ and $b$ is
    \[
        \frac{M}{\sqrt{2}}
        \left(
            \bar\alpha_{\vx a}y_a(\vx)x_b
            +\bar\alpha_{\vx b}y_b(\vx)x_a
        \right).
    \]
    Therefore
    \begin{align*}
        \|\vg(\vx)\|^2
        &=M^2\sum_{a=1}^n\bar\alpha_{\vx a}^2
          +\frac{M^2}{2}\sum_{1\leq a<b\leq n}
          \left(
              \bar\alpha_{\vx a}y_a(\vx)x_b
              +\bar\alpha_{\vx b}y_b(\vx)x_a
          \right)^2\\
        &\leq M^2\sum_{a=1}^n\bar\alpha_{\vx a}^2
          +M^2\sum_{1\leq a<b\leq n}
          \left(
              \bar\alpha_{\vx a}^2x_b
              +\bar\alpha_{\vx b}^2x_a
          \right)\\
        &=M^2\sum_{a=1}^n\bar\alpha_{\vx a}^2
          \left(1+\sum_{b\neq a}x_b\right)\\
        &=M^2\sum_{a=1}^n\bar\alpha_{\vx a}^2(1+m-x_a)\\
        &\leq(m+1)M^2\sum_{a=1}^n\bar\alpha_{\vx a}^2.
    \end{align*}
    Here the inequality uses $(u+v)^2/2\leq u^2+v^2$, together with $x_a^2=x_a$ and $y_a(\vx)^2=1$, and the penultimate equality uses $\sum_ax_a=m$. Combining this with the preceding bound on $M^2\sum_a\bar\alpha_{\vx a}^2$ and applying the triangle inequality gives
    \[
        \|\vg(\vx)-\vomega^*\|
        \leq\|\vomega^*\|^2\sqrt{\frac{m+1}{s_{\min}}}
             +\|\vomega^*\|.
    \]
    This argument applies to every clique. Moreover, Lemma~\ref{lem:gii-strict-mem-ex} gives \(\|\vomega^*\|^2=O_c(v)\), while $m=\Theta_c(v^2)$ and $s_{\min}=\Theta_c(v^2)$. The right-hand side is therefore $O_c(v)$.
\end{proof}

\CliqueUniformMemorization*

\begin{proof}
    Lemma~\ref{lemma:orbit-hsvm-inv}, applied with the clique orbit $\cC_{v,k}$ and the vertex-induced edge action $\Phi_n$, gives $\vtheta^*=\mV\vomega^*\in\Psi(\Phi_n)$.

    By Lemma~\ref{lemma:clique-certificate-deviation}, there is a constant $A_c>0$, depending only on $c$, such that
    \[
        \max_{\vx\in\cC_{v,k}}
        \|\vg(\vx)-\vomega^*\|\leq A_cv.
    \]
    Substituting this estimate into the certificate-sensitive concentration bound~\eqref{eq:certificate-sensitive-concentration} gives, for every $t>0$,
    \[
        \P\!\left(
            \|\widehat{\vomega}_N-\vomega^*\|>t
        \right)
        \leq2\exp\!\left(-\frac{Nt^2}{8A_c^2v^2}\right).
    \]
    Taking $t=\epsilon$, the right-hand side is at most $\delta$ whenever
    \[
        N\geq8A_c^2v^2\epsilon^{-2}\log(2/\delta).
    \]
    Since $\mV$ is an isometry,
    \[
        \|\widehat{\vtheta}_N-\vtheta^*\|
        =\|\mV(\widehat{\vomega}_N-\vomega^*)\|
        =\|\widehat{\vomega}_N-\vomega^*\|.
    \]
    Thus the first assertion holds with $C_1=8A_c^2$.

    For the second assertion, Lemma~\ref{lemma:bounded-representations} and $m=\binom{k}{2}$ give
    \[
        R^2
        \leq\frac{m}{2}+1
        =\frac12\binom{k}{2}+1
        =O_c(v^2).
    \]
    Hence there is a constant $D_c>0$, depending only on $c$, such that $R\leq D_cv$. Taking $t=1/(2R)$ in the preceding concentration bound gives
    \begin{align*}
        \P\!\left(
            \|\widehat{\vomega}_N-\vomega^*\|>\frac{1}{2R}
        \right)
        &\leq2\exp\!\left(-\frac{N}{32A_c^2v^2R^2}\right)\\
        &\leq2\exp\!\left(-\frac{N}{32A_c^2D_c^2v^4}\right).
    \end{align*}
    This probability is at most $\delta$ whenever
    \[
        N\geq32A_c^2D_c^2v^4\log(2/\delta).
    \]
    On the complementary event, for every $\vx\in\cC_{v,k}$ and $j\in[n]$,
    \begin{align*}
        \langle\vu_j(\vx),\widehat{\vomega}_N\rangle
        &\geq\langle\vu_j(\vx),\vomega^*\rangle
             -\|\vu_j(\vx)\|
              \|\widehat{\vomega}_N-\vomega^*\|\\
        &\geq1-R\frac{1}{2R}=\frac12.
    \end{align*}
    Lemma~\ref{lem:one-bit-diff} identifies this inner product with the corresponding energy gap. Thus $\widehat{\vtheta}_N$ memorizes every $k$-clique with energy gap at least $1/2$, proving the second assertion with $C_2=32A_c^2D_c^2$.
\end{proof}

\section{Additional Experiments and Further Preliminary Results}\label{app:add-exp}

\subsection{Further training and test accuracy curves}
\begin{figure}[!htb]
    \centering
    \setlength{\tabcolsep}{3pt}
    \begin{tabular}{ccc}
        \includegraphics[width=.32\textwidth]{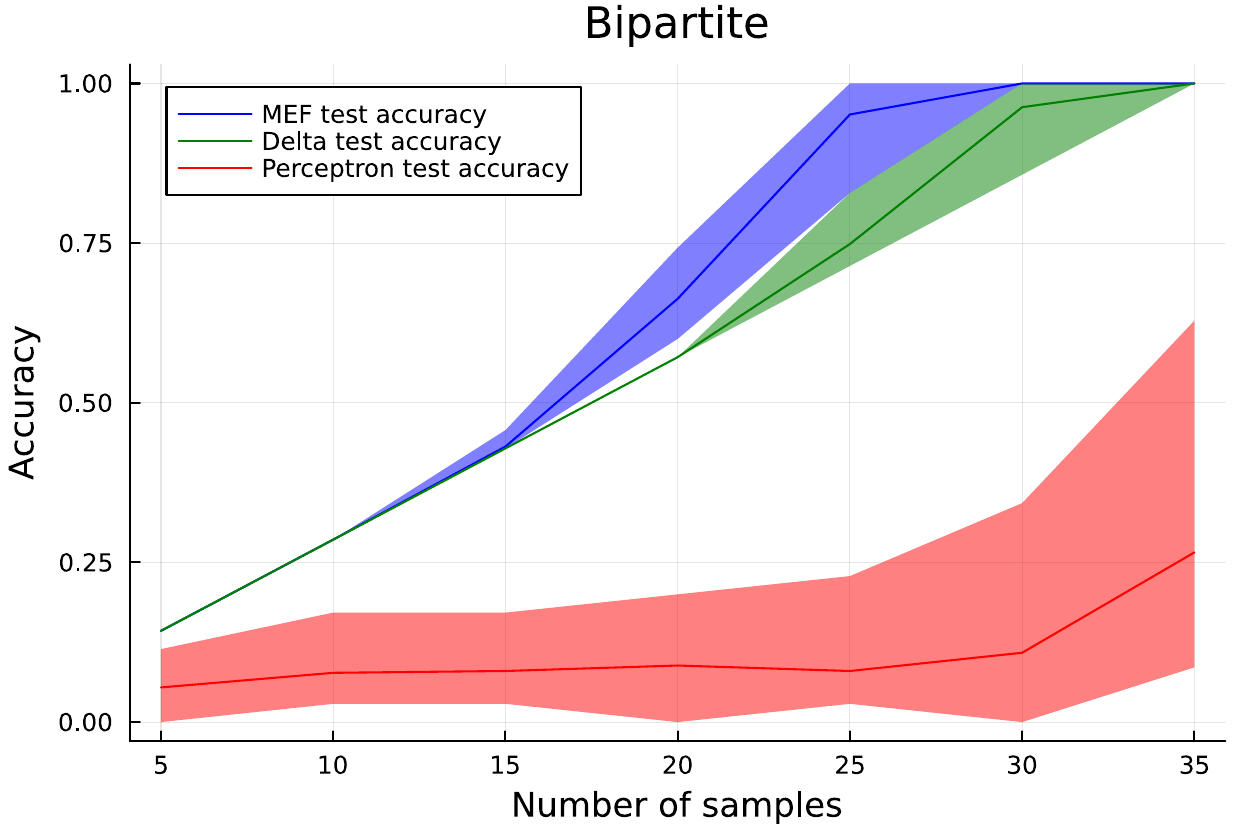} &
        \includegraphics[width=.32\textwidth]{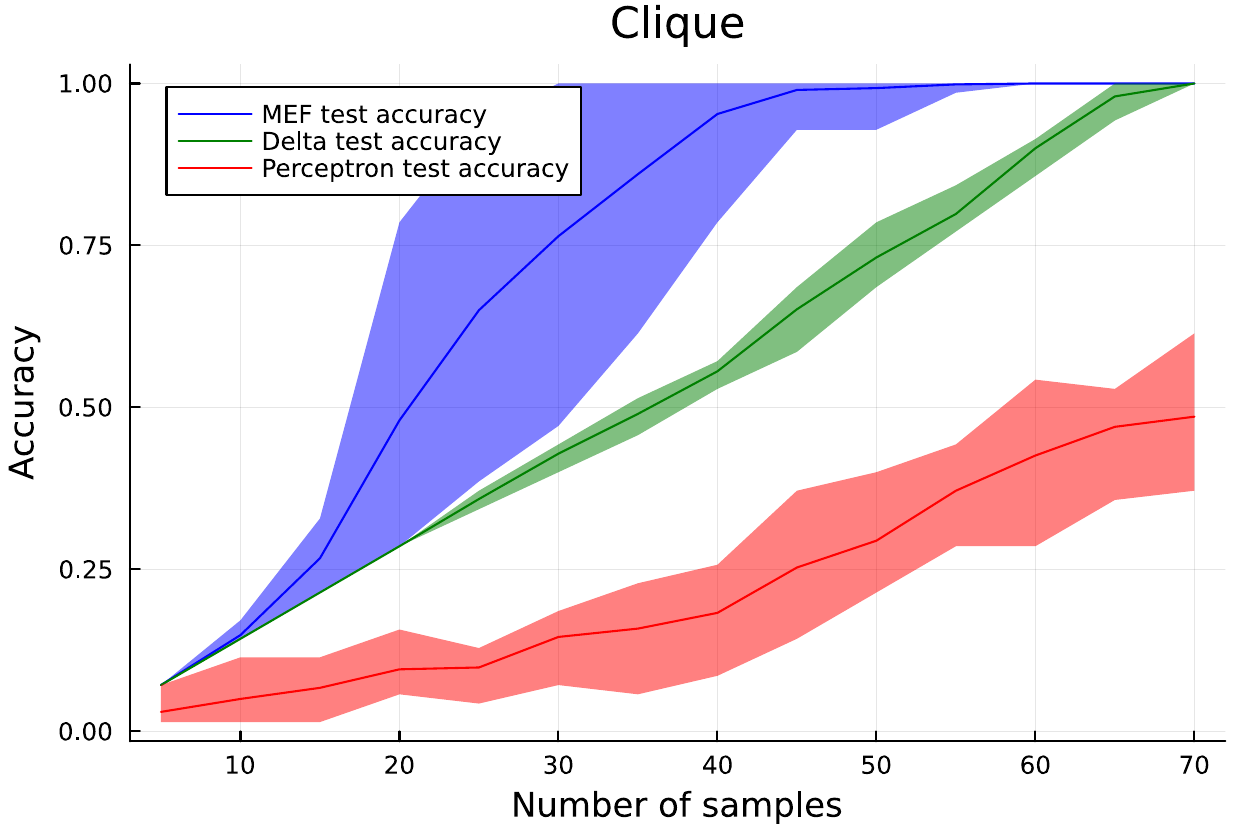} &
        \includegraphics[width=.32\textwidth]{figs/Small/paley_graph.pdf} \\
        \includegraphics[width=.32\textwidth]{figs/Big/random_bipartite.pdf} &
        \includegraphics[width=.32\textwidth]{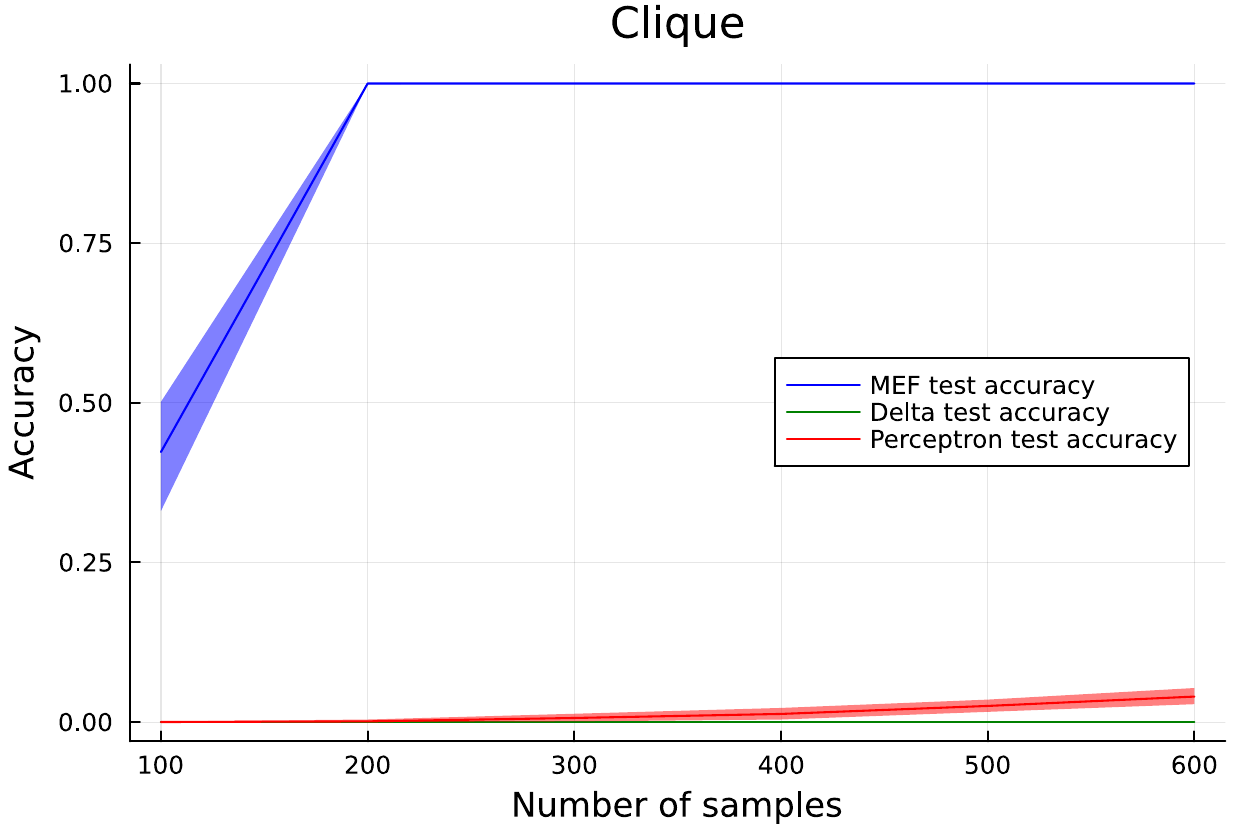} &
        \includegraphics[width=.32\textwidth]{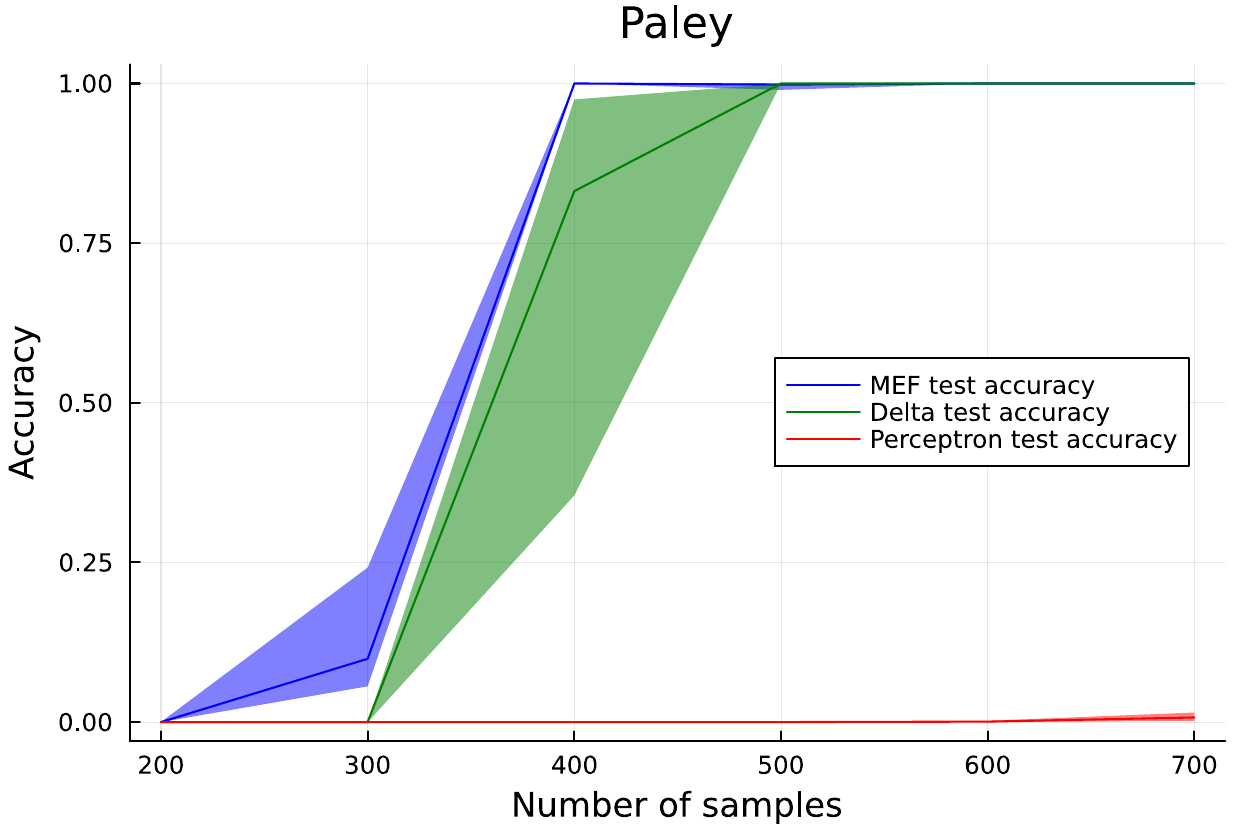} \\
    \end{tabular}
    \caption{\textbf{Test accuracy vs.\ training sample size for isomorphism classes at two scales.} Top row: $v=8$ (class sizes: bipartite $35$ and quadratic-residue $2{,}520$). Bottom row: $v=20$ (the bipartite class has size $92{,}378$). The quadratic-residue graphs are labelled ``Paley'' in the plots. Curves show the mean and minimum--maximum range over $10$ trials. Networks are trained with the Perceptron, Delta (MSE), and MEF learning rules.}
    \label{fig:app:accuracy_combined}
\end{figure}

Fig.~\ref{fig:app:accuracy_combined} shows test accuracy versus training sample size, with the mean and minimum--maximum range over $10$ trials, for Hopfield networks trained by MEF, Perceptron, and Delta (the latter two are used only as baselines; see Appendix~\ref{app:learning-rules}). For small graphs ($v=8$), we enumerate the full isomorphism class and report the true accuracy, i.e., the fraction of the class memorized. For larger graphs ($v=20$), accuracy is estimated on an independent random sample of $1{,}000$ graphs. Within our hyperparameter range, the Delta rule using Adam failed on the $k$-clique class, whereas MEF learned all classes and was insensitive to optimizer choice.

\subsection{Further parameter heatmaps}\label{app:more-heatmaps}
\begin{figure}[!htb]
    \centering
    \includegraphics[scale=0.25]{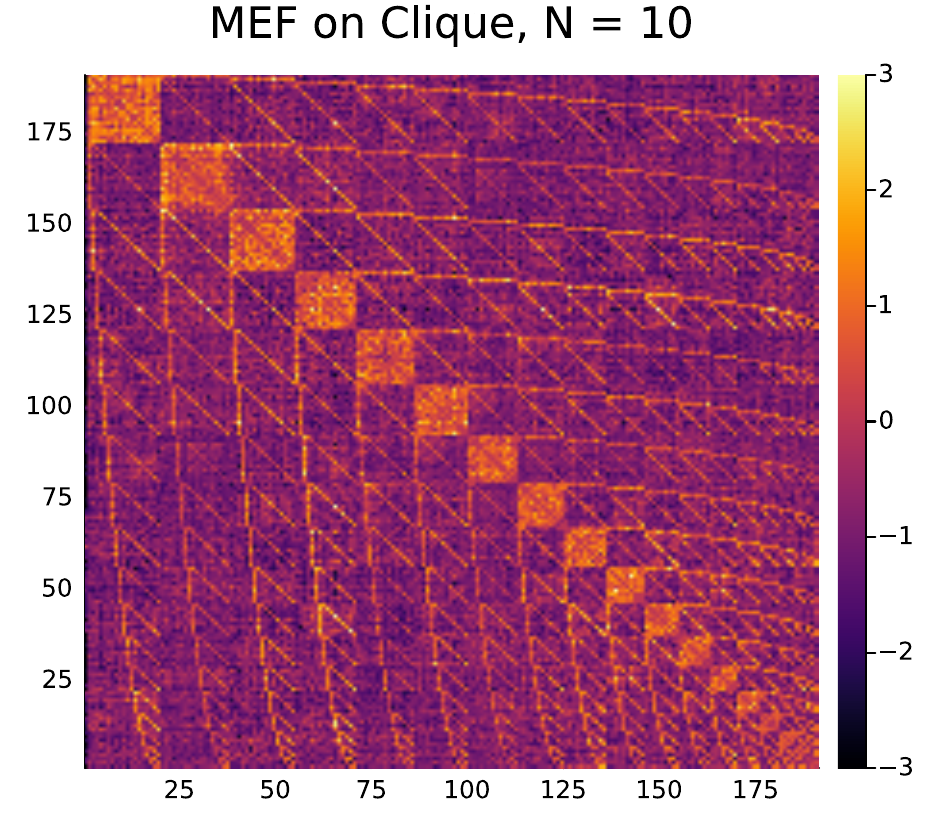}
    \includegraphics[scale=0.25]{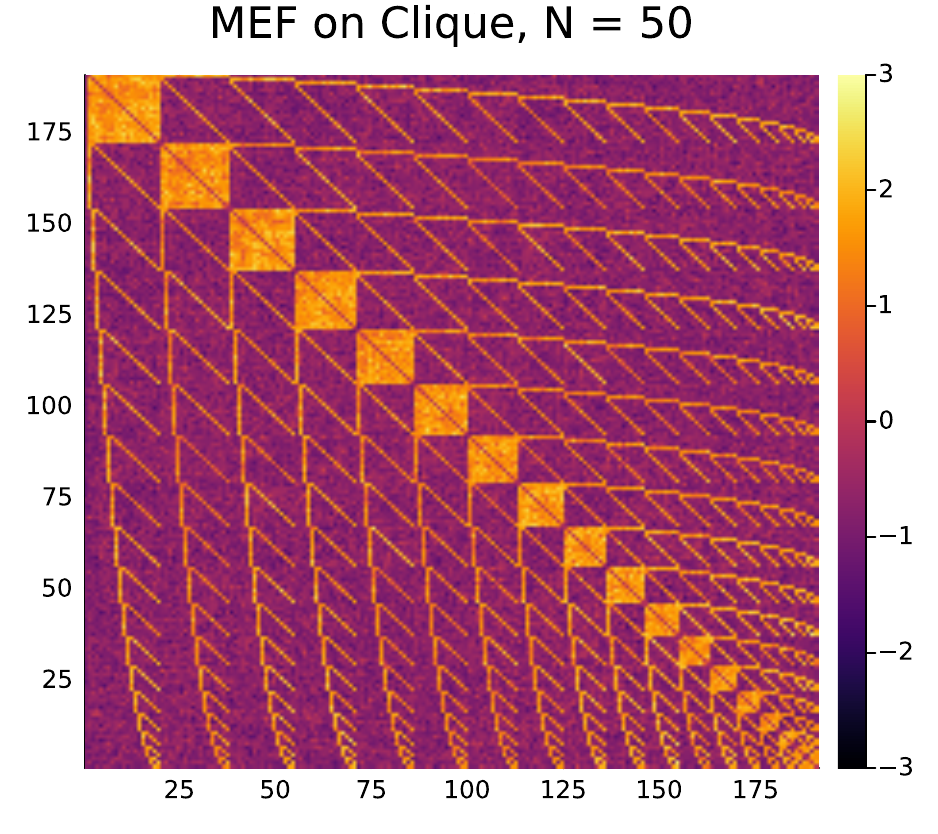}
    \includegraphics[scale=0.25]{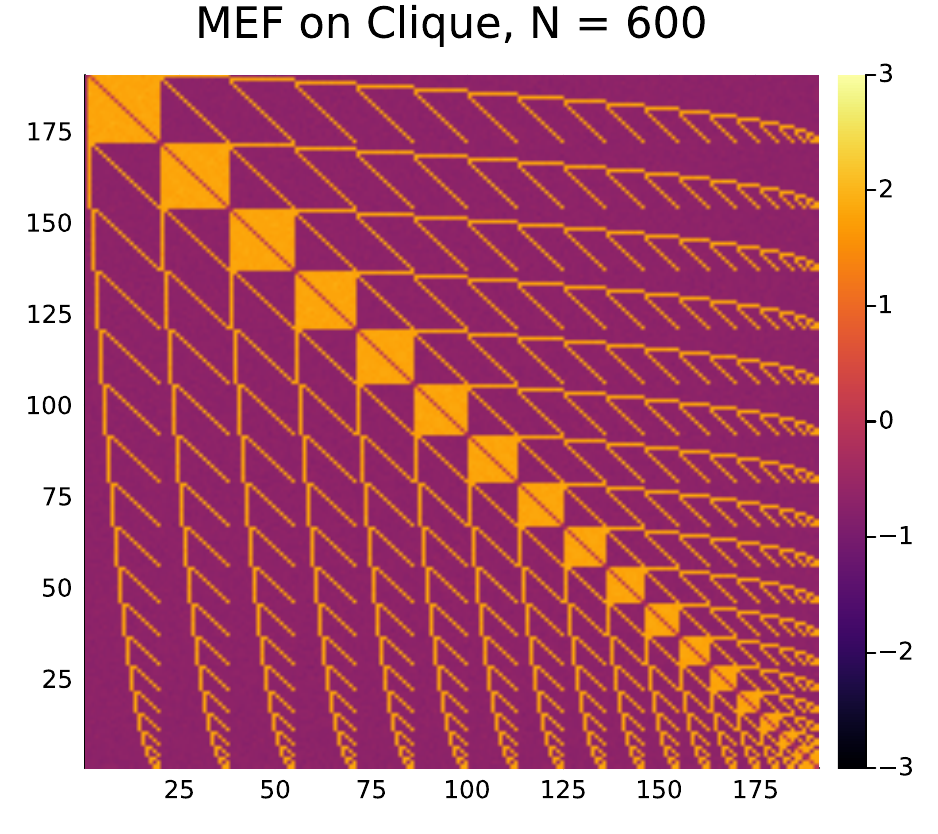}
    \includegraphics[scale=0.25]{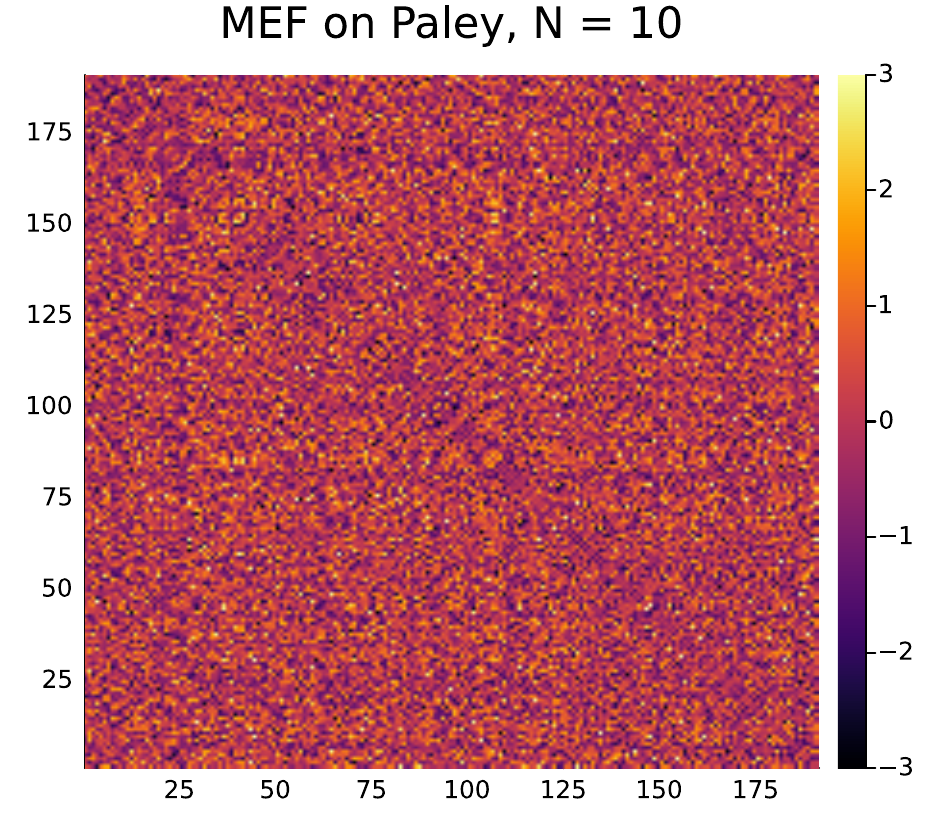}
    \includegraphics[scale=0.25]{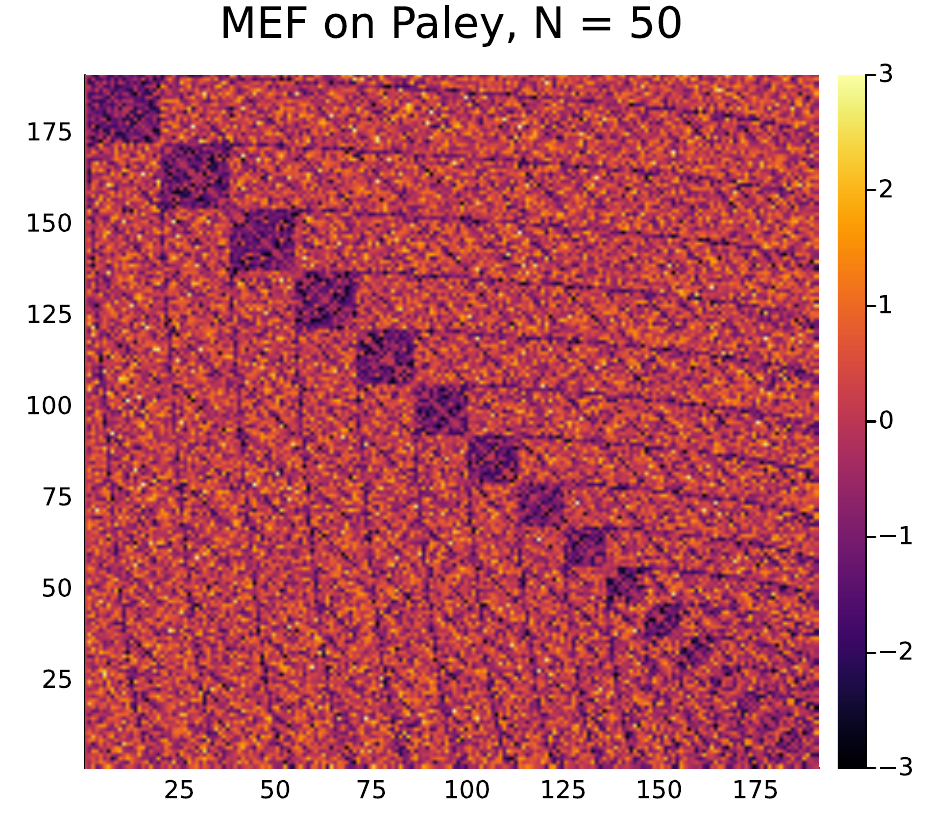}
    \includegraphics[scale=0.25]{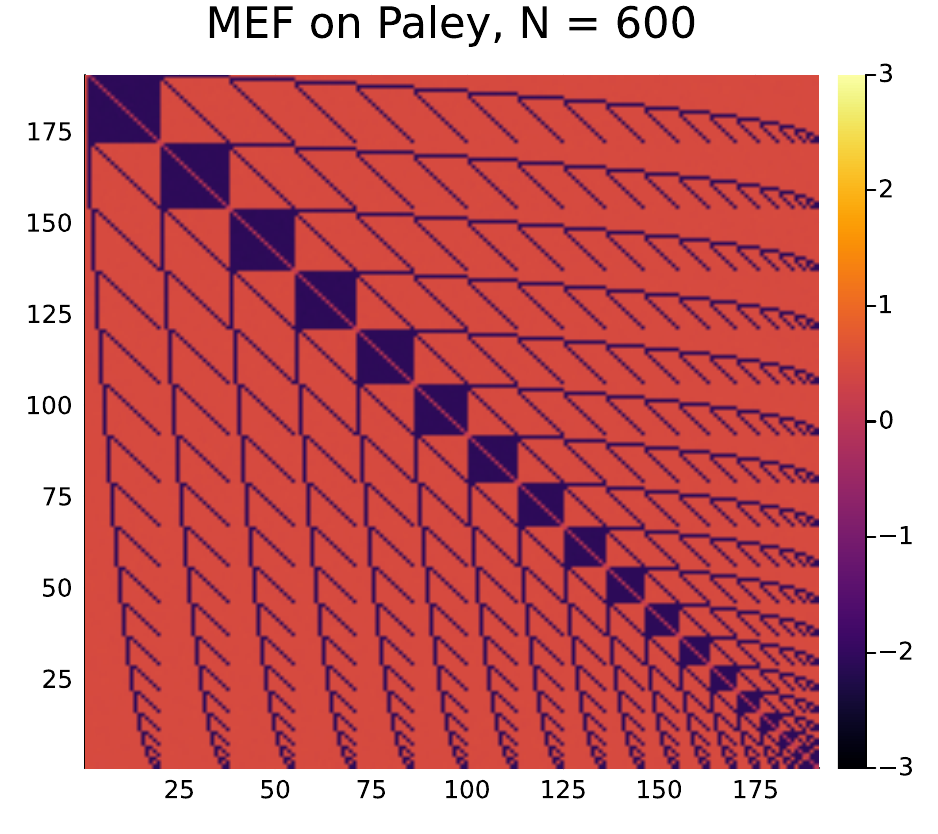}
    \includegraphics[scale=0.25]{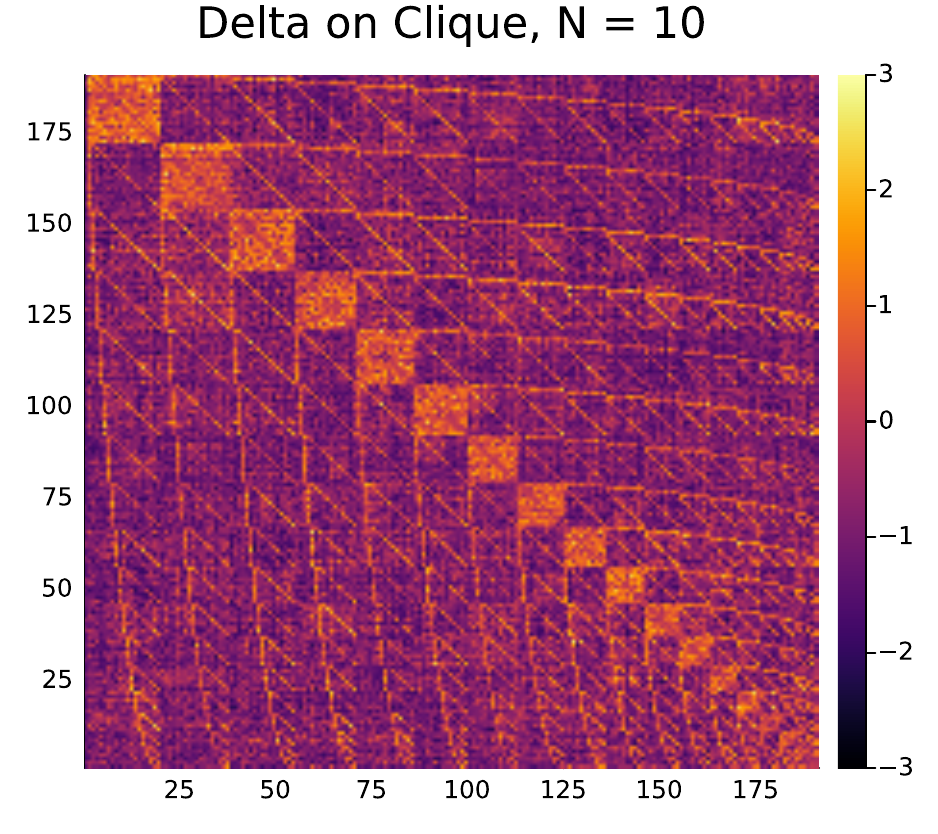}
    \includegraphics[scale=0.25]{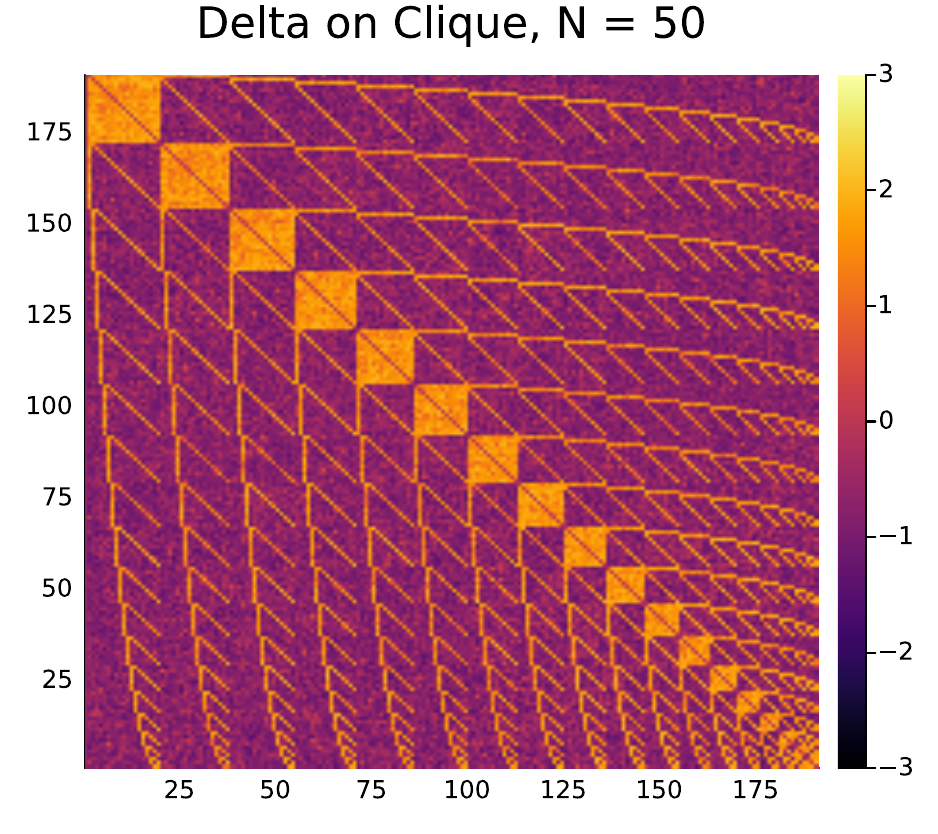}
    \includegraphics[scale=0.25]{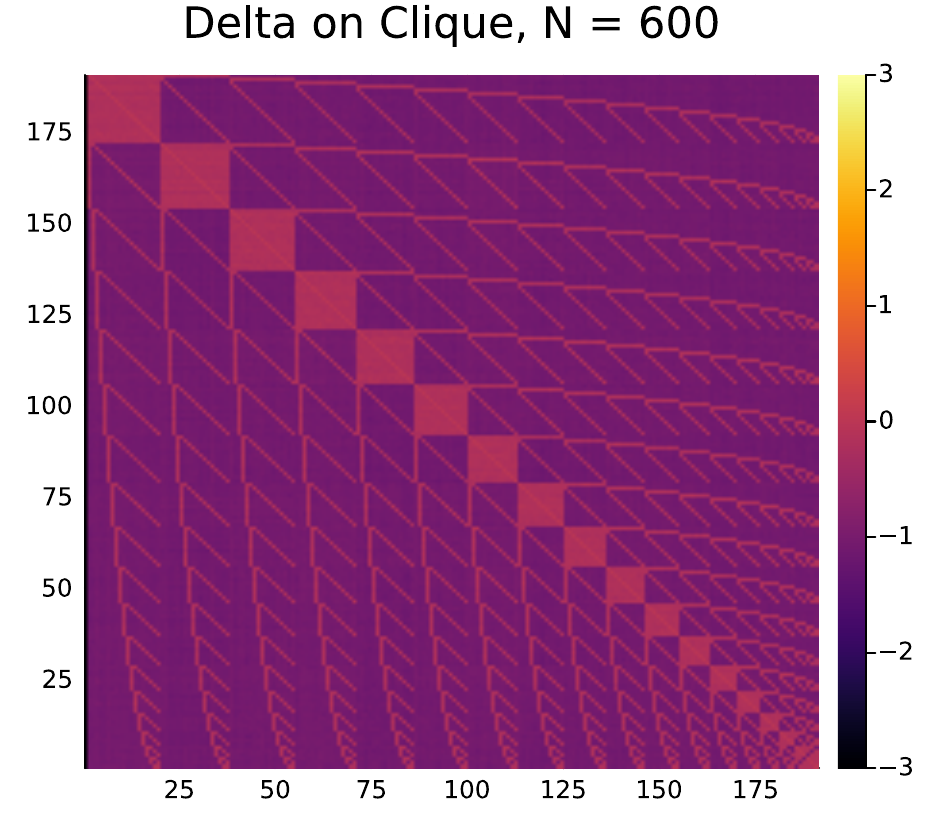}
    \includegraphics[scale=0.25]{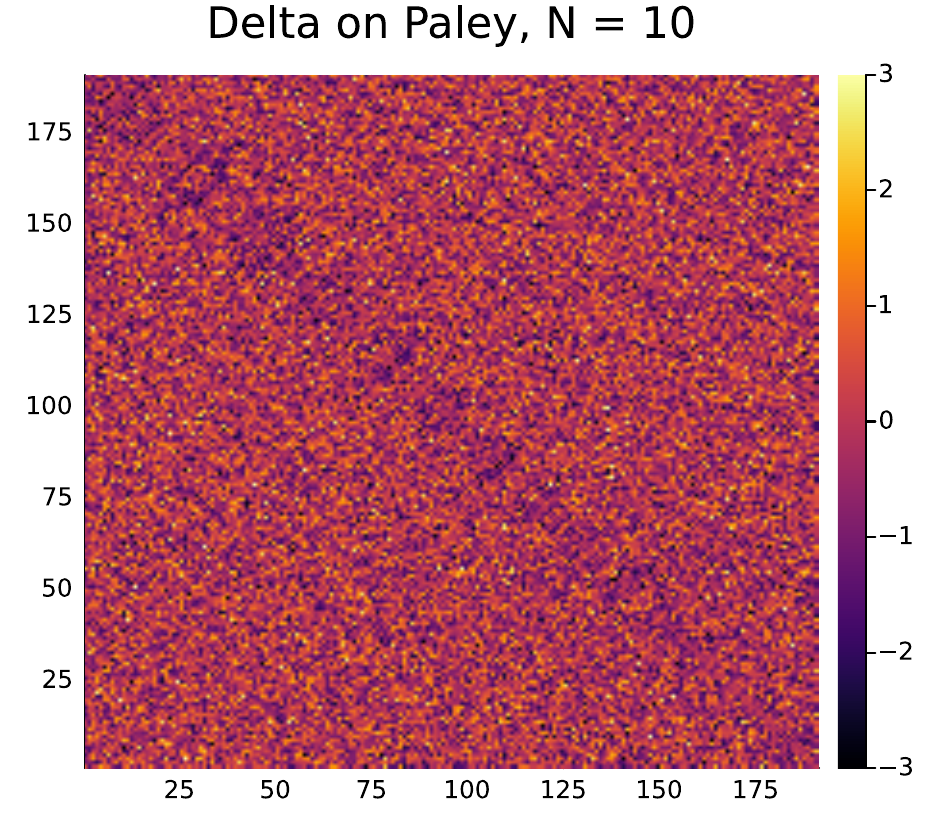}
    \includegraphics[scale=0.25]{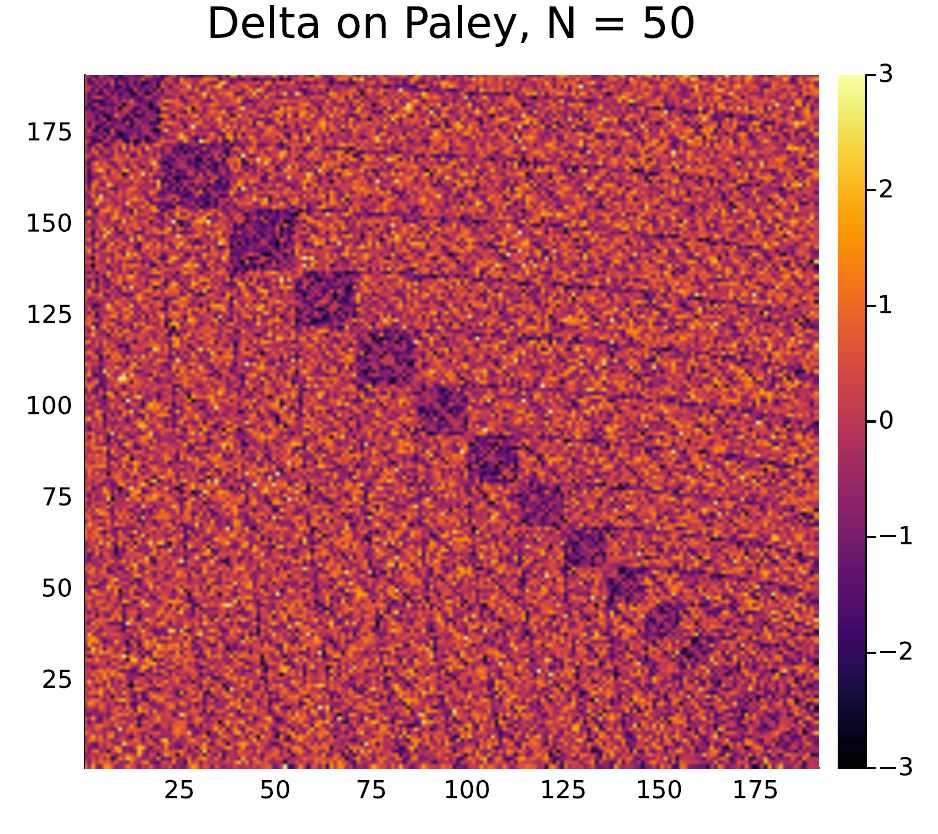}
    \includegraphics[scale=0.25]{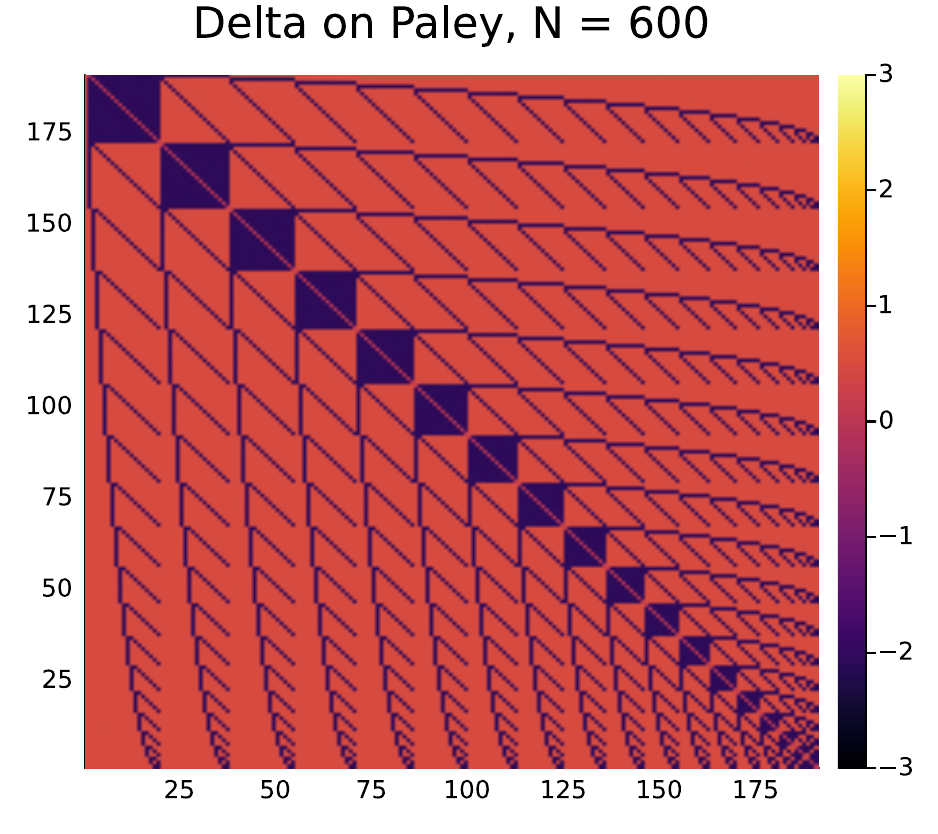}
    \caption{\textbf{Weights found by MEF and Delta while varying $N$.} Networks were trained on samples from the isomorphism classes of $10$-cliques and quadratic-residue graphs on $v=20$ vertices, with sample sizes ranging from $N=10$ to $600$. The quadratic-residue graphs are labelled ``Paley'' in the plots.}
    \label{fig:weight_heatmaps2}
\end{figure}

Fig.~\ref{fig:weight_heatmaps2} shows the weight matrices for MEF and Delta on clique and quadratic-residue graph data. The plots retain the label ``Paley.'' The Delta rule also appears to return solutions that approach the invariant space as the sample size $N$ increases.

\subsection{Sample Complexity for Robust Exponential Memory}\label{app:HCP}

\begin{figure}
    \centering
    a)\includegraphics[scale=0.58]{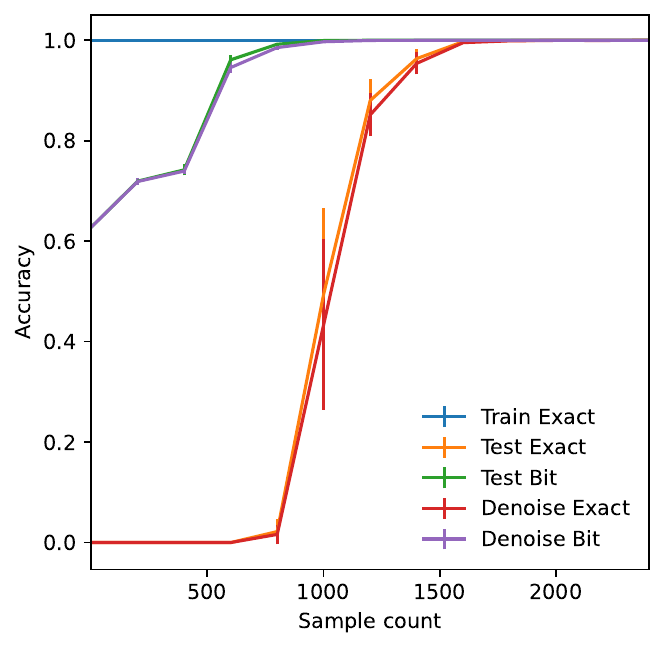}
    b)\includegraphics[scale=0.58]{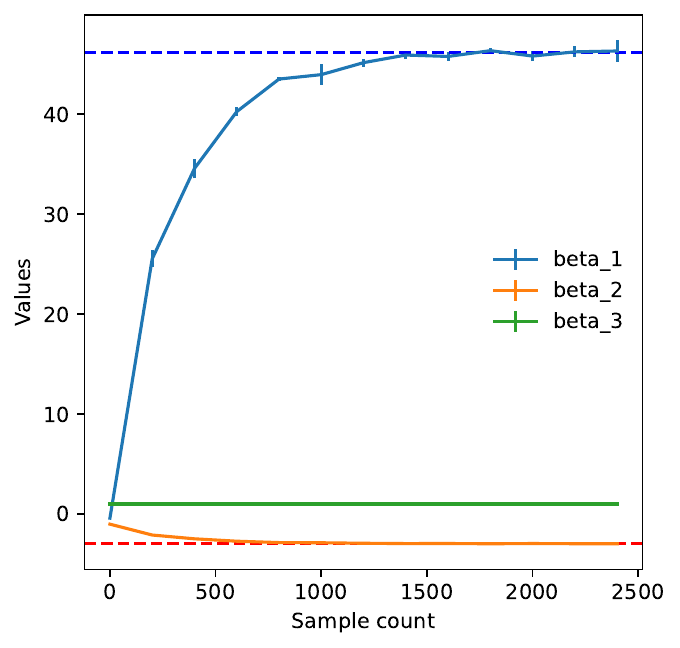}
    \caption{\textbf{Generalization on the Hidden Clique Problem.} a) Generalization and denoising accuracy (exact and average-bit accuracy) as functions of the number of $50$-clique samples for MEF-trained HNs on $100$-vertex graphs ($n=4{,}950$ neurons). Generalization accuracy was computed on a test set of $10{,}000$ novel graphs. Denoising accuracy was computed by corrupting $5\%$ of the bits in these graphs and evaluating the attractor reached from each noisy pattern ($5$ trials; error bars show standard deviations). b) For each learned parameter type (weights indexed by adjacent or nonadjacent edges, and thresholds), we plot its average against the number of training samples, normalized so that the thresholds have mean absolute value one. Dotted horizontal lines show the parameter-type averages obtained using the largest training sample.}
    \label{fig:HCP}
\end{figure}

Robust exponential memory in Hopfield networks can be viewed through the related lenses of error-correcting codes and recovery of latent combinatorial structure under noise. Prior work has shown, specifically for $k$-cliques, that Hopfield networks can store entire isomorphism classes with large basins of attraction and asymptotically optimal noise tolerance. Corollary~\ref{cor:clique-uniform-memorization} now shows that a polynomial number of uniform clique samples suffices for the exact sample HSVM to memorize the entire clique orbit with high probability, while Theorem~\ref{lem:gen-svm-bounds} separately controls fresh-sample memorization for MEF. Transferring the uniform-orbit guarantee to a prescribed finite number of MEF iterations, and proving comparable basin guarantees for the learned solution, remain natural future questions. Fig.~\ref{fig:HCP} illustrates the empirical phenomenon by plotting generalization and denoising accuracy against sample count for large MEF-trained HNs ($v=100$; $n=4{,}950$ neurons).

\subsection{Clique generalization in DAMs}\label{app:dams}

\begin{figure}
    \centering
    a) \setlength{\unitlength}{1pt}%
    \begin{picture}(183.7,139.9)
        \put(0,0){\includegraphics[scale=0.45]{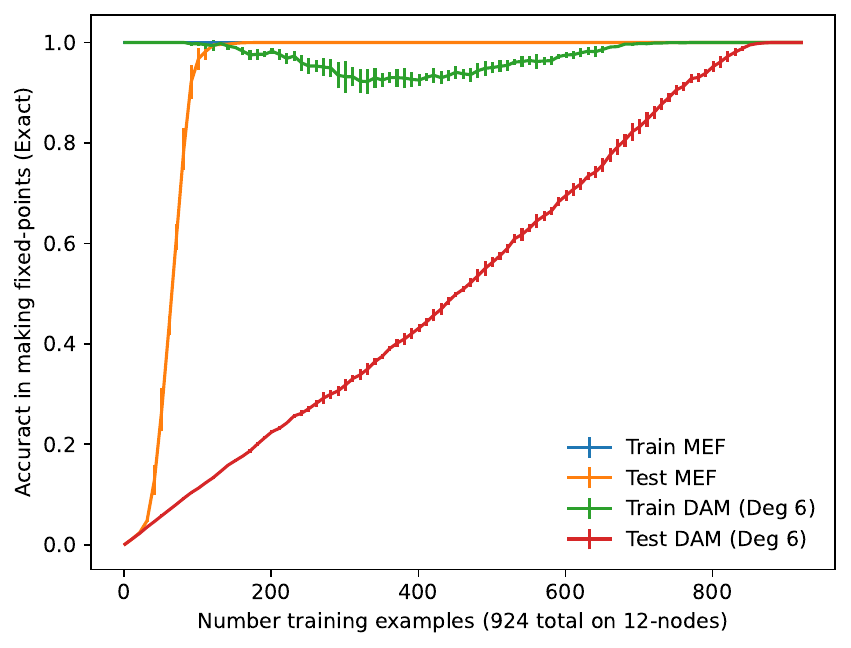}}
        \put(0,22){\color{white}\rule{9pt}{100pt}}
        \put(3,70){\rotatebox[origin=c]{90}{\tiny Accuracy in making fixed points (Exact)}}
    \end{picture}
    b) \includegraphics[scale=0.45]{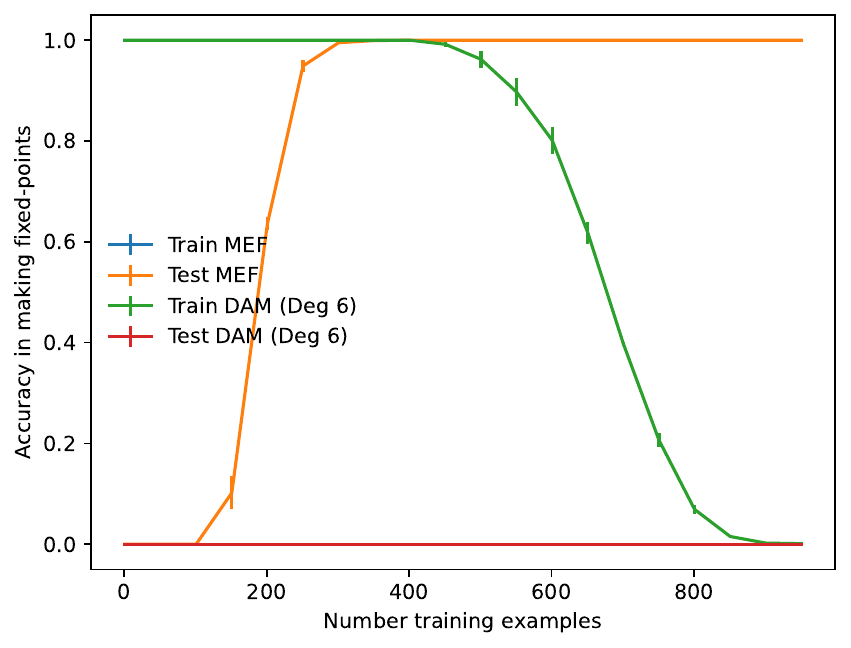}
    \caption{\textbf{DAM models trained on cliques.} Generalization performance of DAMs and MEF-trained HNs for a) the $6$-clique problem on graphs with $v=12$ vertices and b) the $16$-clique problem on graphs with $v=32$ vertices ($496$-neuron networks). Training accuracy is the percentage of clique training samples that are attractors. In a), test accuracy is the percentage of all $6$-cliques that are fixed points; in b), it is evaluated on a random set of $10{,}000$ $16$-cliques. Scores are averaged over four trials; error bars show standard deviations.}
    \label{fig:gen_dams}
\end{figure}

We conducted experiments comparing the generalization performance of DAMs with that of MEF when presented with increasing numbers of cliques as training data.  We used the architecture and algorithm described in \citet{krotov2016dense} for polynomial degree $6$ in the energy function.  The results in Fig.~\ref{fig:gen_dams} show very different generalization behaviors.  These preliminary investigations suggest that DAMs do not generalize well in the setting of cliques, but much more work is needed to understand their behavior.

\subsection{Graph Families}\label{app:classes}

\begin{figure}
    \centering
    a)\includegraphics[scale=0.44]{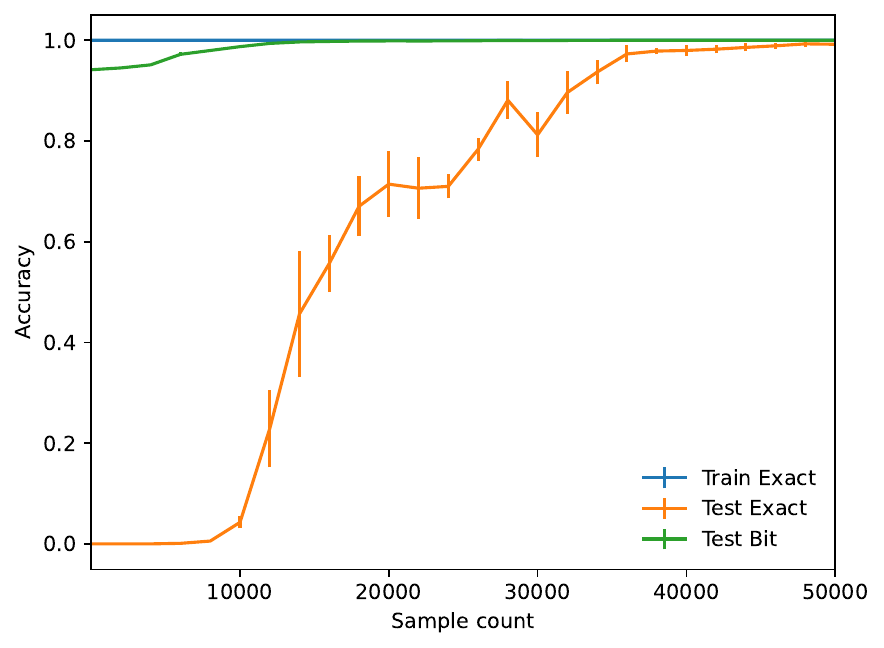}
    b)\includegraphics[scale=0.44]{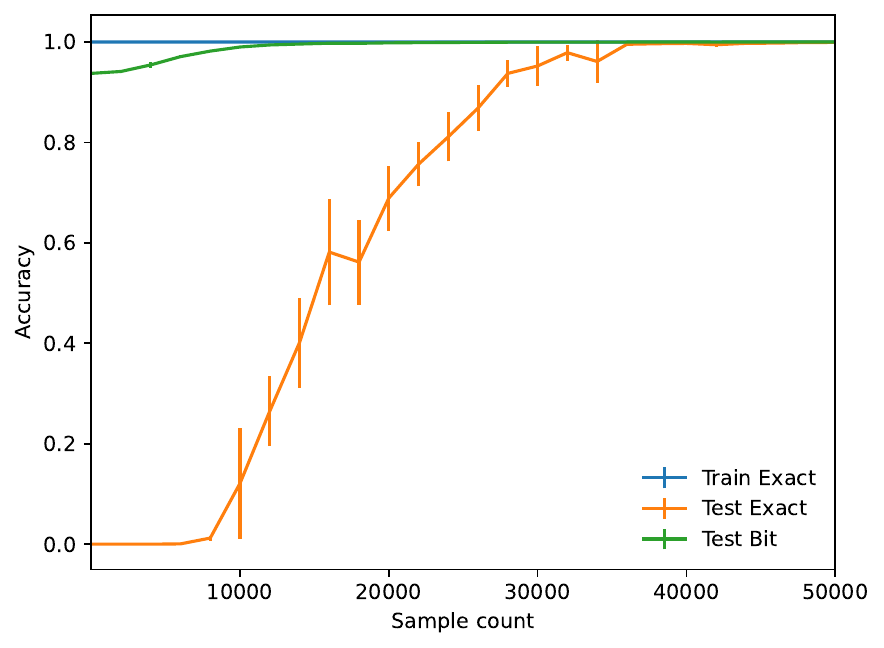}
    \\
    c)\includegraphics[scale=0.44]{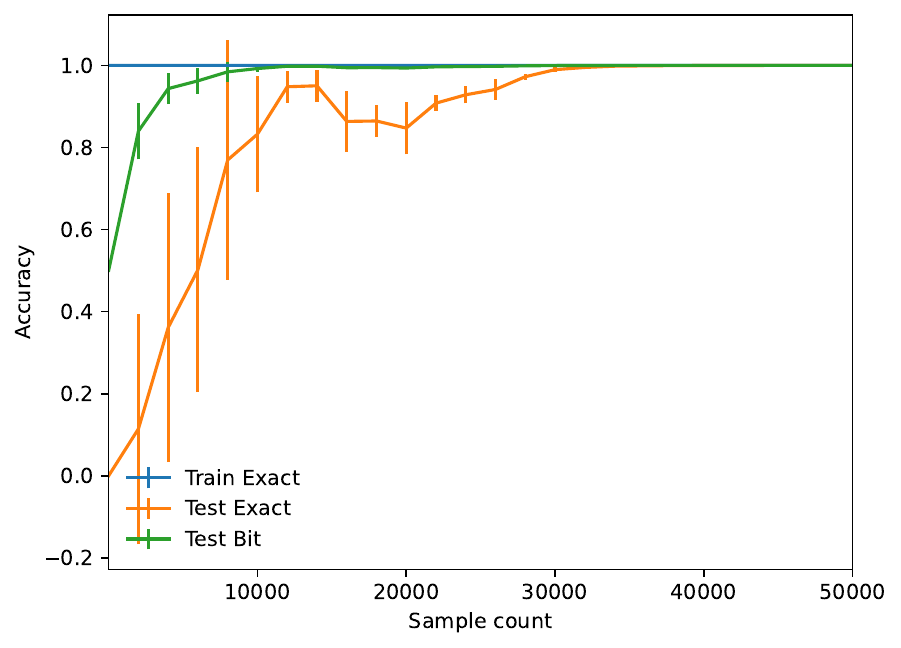}
    d)\includegraphics[scale=0.44]{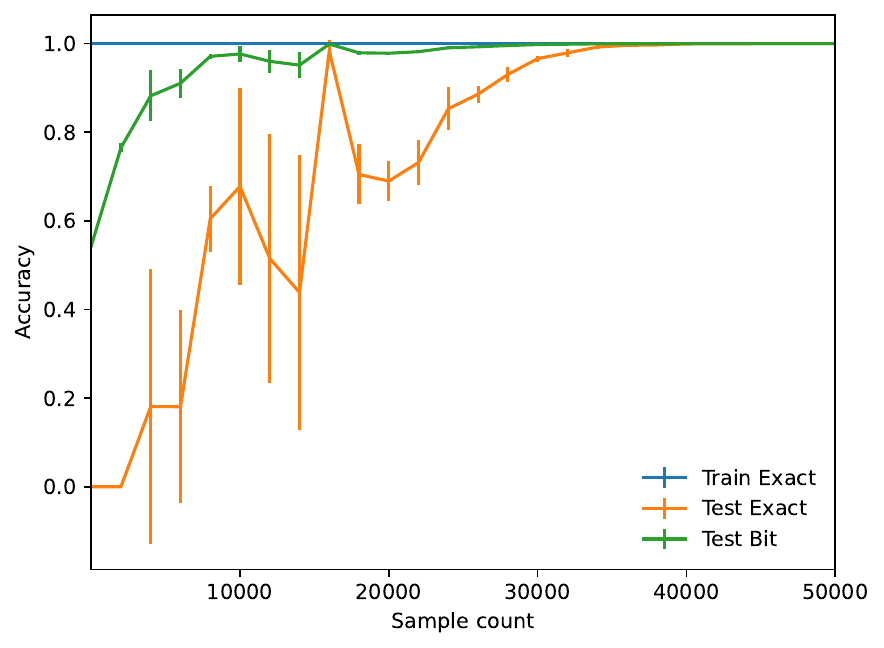}
    \caption{\textbf{Generalization for other graph classes.} Accuracy of MEF-trained HNs as a function of the number of training samples for a) a chain graph with $v=32$, $k=16$ ($496$ neurons); b) a cycle graph with $v=32$, $k=16$ ($496$ neurons); c) a bipartite graph with $v=32$, $k=16$ ($496$ neurons); and d) the Johnson graph $J(7,3)$ on $v=35$ vertices ($595$ neurons). Results are averaged over $10$ trials; error bars show standard deviations. Test sets contain $10{,}000$ novel graphs sampled uniformly from the isomorphism class.}
    \label{fig:other_families}
\end{figure}

\begin{figure}[!htb]
    \centering
    a)\includegraphics[scale=0.49]{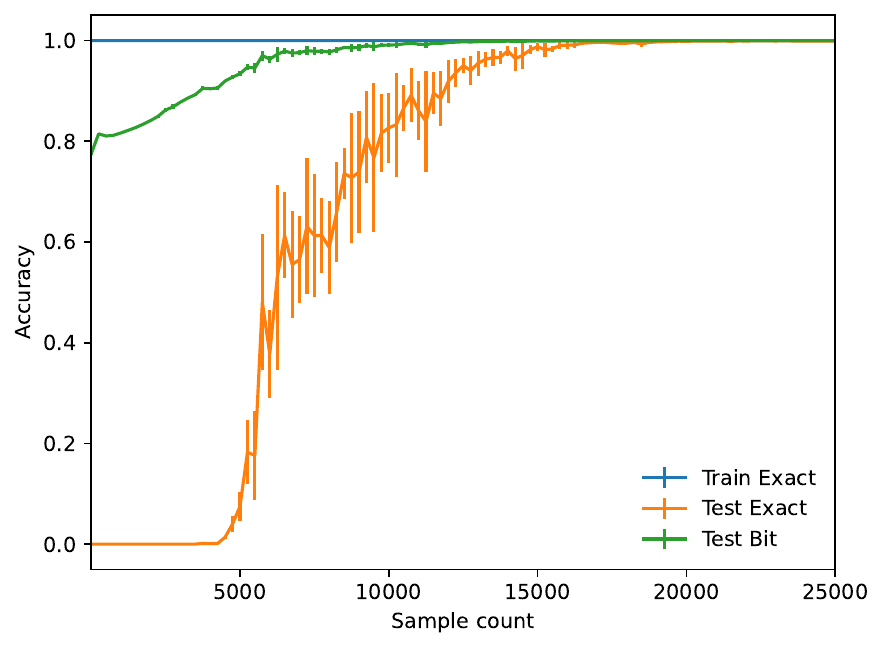}
    b)\includegraphics[scale=0.49]{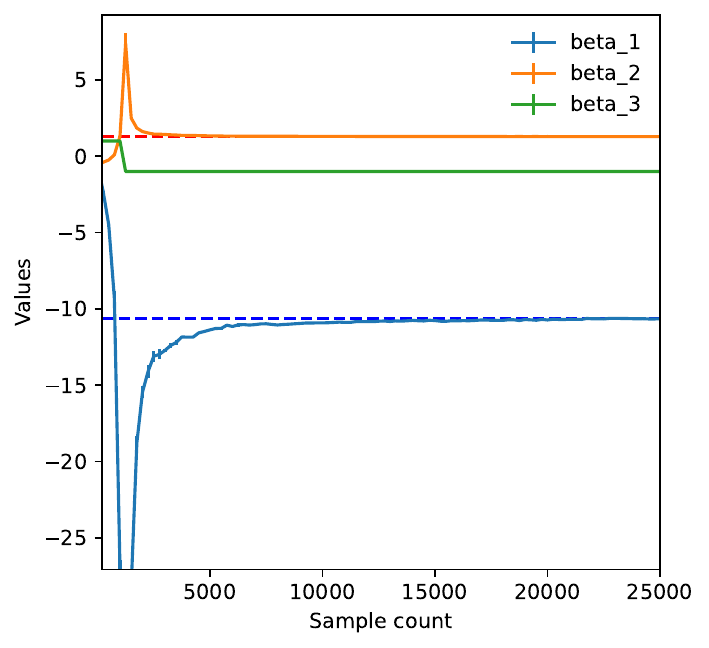}
    \\
    c)\includegraphics[scale=0.49]{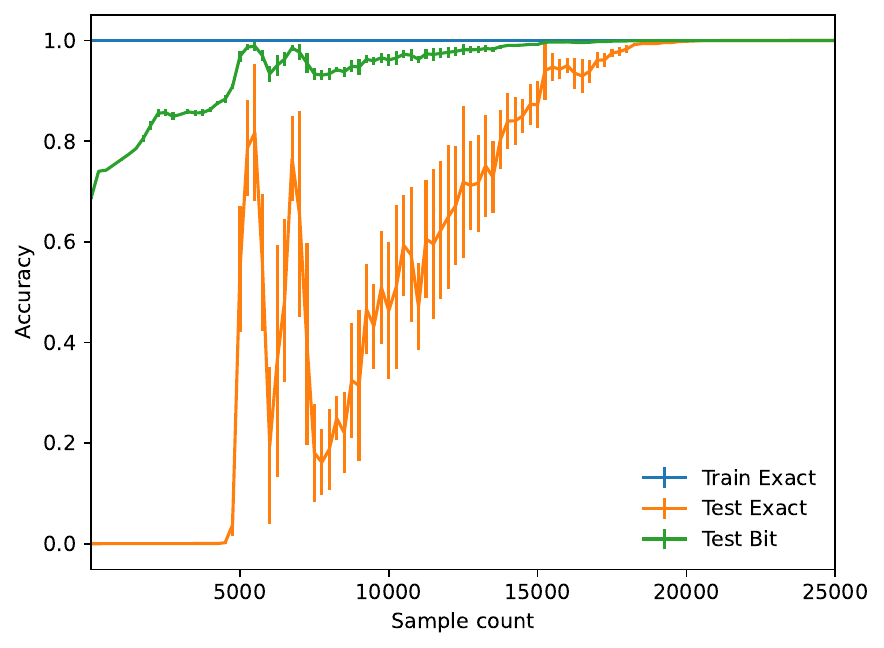}
    d)\includegraphics[scale=0.49]{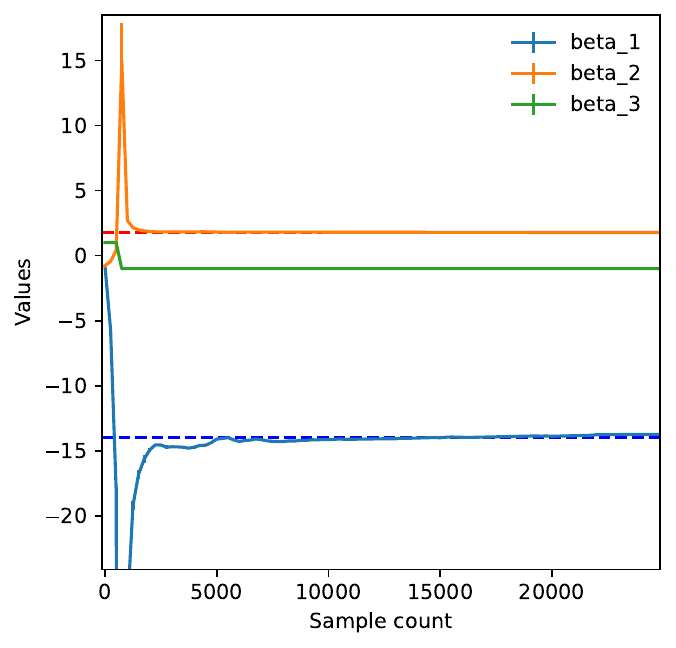}
    \caption{\textbf{Generalization for circulant graphs.} a) Accuracy of MEF-trained HNs as a function of the number of training samples for a circulant graph with $v=32$ and jump sequence $[2,4]$. b) Averages of each learned parameter type for the data in a), normalized so that the thresholds have mean absolute value one; dotted horizontal lines show the averages obtained using the largest training sample. c)--d) Corresponding results for a circulant graph with $v=32$ and jump sequence $[2,3,5]$. Results are averaged over $10$ trials; error bars show standard deviations. Test sets contain $10{,}000$ novel graphs sampled uniformly from the isomorphism class.}
    \label{fig:other_families2}
\end{figure}

We also observe generalization for several other graph families. Fig.~\ref{fig:other_families} shows training and test accuracy, both bitwise and exact, for chain, cycle, bipartite, and Johnson graphs (see also Fig.~\ref{fig:weight_hists}). In the first three cases, $k$ denotes the number of vertices in the corresponding chain, cycle, or bipartite subset. Accuracy generally increases with the number of training samples, although some settings exhibit nonmonotone, double-descent-like behavior; see Fig.~\ref{fig:other_families}d for the Johnson family and Section~\ref{DDP} for cliques. Fig.~\ref{fig:other_families2} reports additional experiments with circulant graphs.

\subsection{Double Descent Phenomenon}\label{DDP}

\begin{figure}[!htb]
    \centering
    a)\includegraphics[scale=0.43]{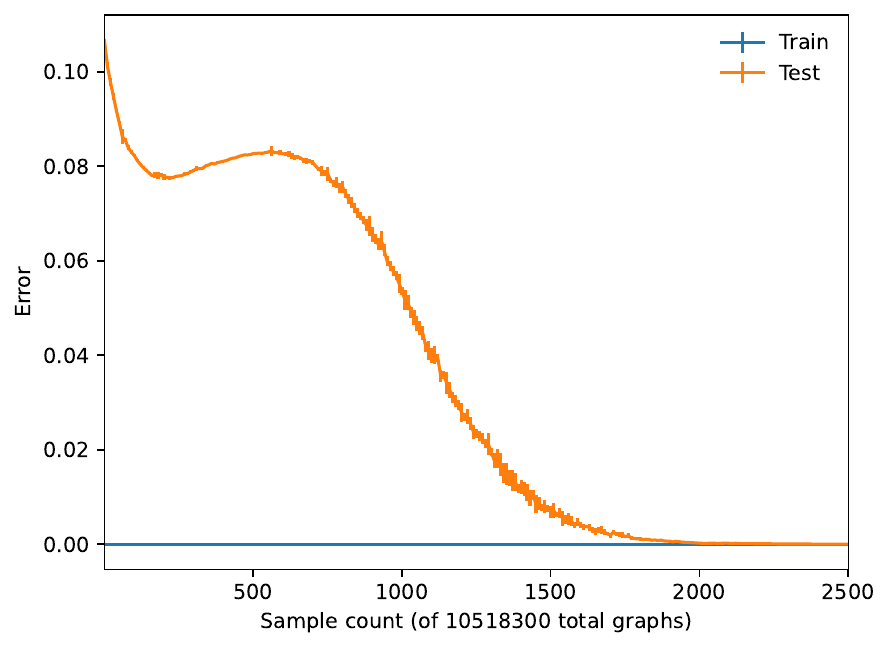}
    b)\includegraphics[scale=0.43]{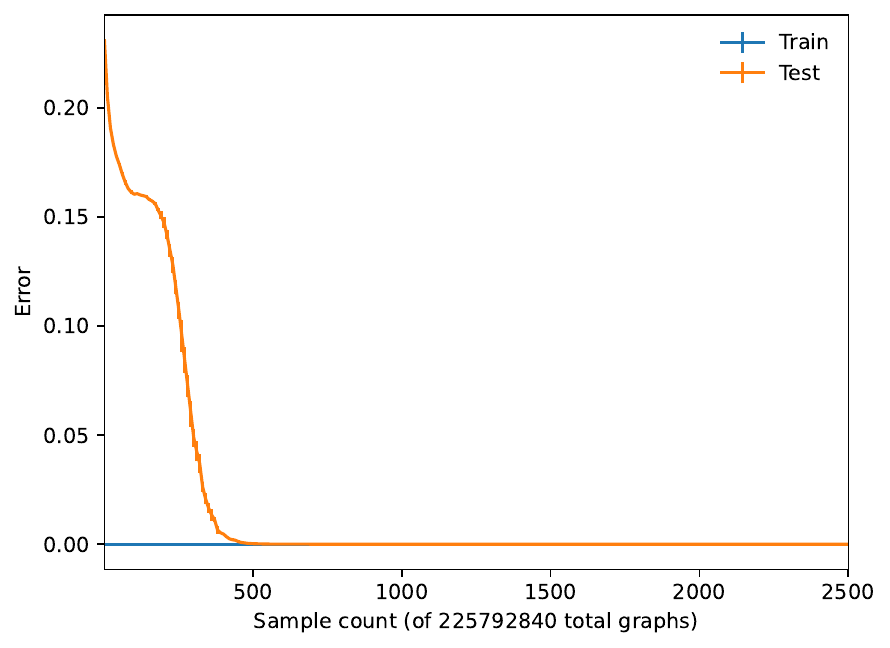}
    \caption{\textbf{Double-descent phenomenon.} Average test bit error on $10{,}000$ random cliques as a function of the number of MEF training samples for a) $v=32$, $k=8$, and b) $v=32$, $k=12$. The nonmonotone behavior is less pronounced for the larger value of $k$. Results are averaged over $5$ independently sampled training sets; error bars show standard deviations.}
    \label{fig:ddescent}
\end{figure}

In some experiments, test error decreased with the number of training samples, then increased temporarily before decreasing again to zero. Fig.~\ref{fig:ddescent} gives two examples for the bit error between attractors and samples in the clique setting. This behavior appears more pronounced when $k$ is small relative to $v$. These experiments also suggest that more samples are required to learn the full isomorphism class for small $k$, with the easiest cases occurring when $k$ is near $v/2$.

Further nonmonotone behavior in exact accuracy (the percentage of samples that are fixed points of the dynamics) can be found in Fig.~\ref{fig:other_families}(c,d) for bipartite and Johnson graphs, and Fig.~\ref{fig:other_families2}(c) for circulant graphs. We postpone theoretical analysis of these findings for future work.

\clearpage
\subsection{The Hopfield Network Negative Graph-Isomorphism Certificate (HNNGIC)}
Lemma~\ref{lemma:express-inv-param} implies that any graph-isomorphism class can be stored in a Hopfield network. This observation suggests a one-sided certificate of graph non-isomorphism, which we call the \textit{Hopfield Network Negative Graph-Isomorphism Certificate} (HNNGIC). Algorithm~\ref{alg:hngic} returns True only when it certifies that the two input graphs are non-isomorphic; otherwise it returns Unknown.

\begin{algorithm}[H]
\SetAlgoLined
\KwIn{edge representations $\vx_1,\vx_2\in\{0,1\}^n$ of two graphs, and computational budget $B$}
\KwOut{True or Unknown}
\BlankLine
\textbf{Step 1:} minimize $L(F(\vbeta);\{\vx_1\})$ within computational budget $B$, returning $\vbeta^*\in\R^3$\;

\textbf{Step 2:} \If{$\min_{j\in[n]}\bigl(E(\vx_1^{(j)};F(\vbeta^*))-E(\vx_1;F(\vbeta^*))\bigr)\leq0$}{
    \textbf{return} Unknown\;
}

\textbf{Step 3:} \If{$H(\vx_2;F(\vbeta^*))\neq\vx_2$}{
    \textbf{return} True\;
}

\textbf{return} Unknown\;
\caption{Hopfield Network Negative Graph-Isomorphism Certificate (HNNGIC)}
\label{alg:hngic}
\end{algorithm}

The algorithm takes $\vx_1$ as the reference graph and minimizes the single-example energy-flow loss while restricting the parameters to the edge-adjacency-invariant subspace $\Psi(\cQ_n)$. Step~2 explicitly verifies strict memorization of $\vx_1$. By Lemma~\ref{lemma:gii-lb}, strict memorization then propagates to every graph in its isomorphism orbit. Consequently, if $\vx_2$ is not a fixed point, it cannot be isomorphic to $\vx_1$, and returning True is sound. If $\vx_2$ is a fixed point, the algorithm returns Unknown because it may be a spurious state unrelated to the orbit of $\vx_1$. Lemmas~\ref{lemma:express-inv-param} and~\ref{lemma:gii-lb} show that a three-parameter invariant network strictly memorizing $\vx_1$ exists, although the budget-limited optimization in Step~1 need not find it.